# NATURAL LANGUAGE PROCESSING
# FOR  WORD SENSE DISAMBIGUATION AND
# INFORMATION EXTRACTION

**A**
**Thesis**
**Submitted to the**
**Jai Narain Vyas University, Jodhpur**
**for the award of the degree of**

## DOCTOR   OF   PHILOSOPHY

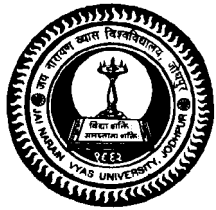

**Supervisor**
**Dr. V.S. Bansal, Ph.D.(U.K.),**
**M.S.(U.S.A.)**

**Researcher**
**K.R. Chowdhary**
Associate Professor,
Department of Computer
Science & Engineering

## Department of  Electrical Engineering
Faculty of Engineering, M.B.M. Engineering College,
Jai Narain Vyas University, Jodhpur-342011.
February 2004.

# NATURAL LANGUAGE PROCESSING FOR  WORD SENSE DISAMBIGUATION AND INFORMATION EXTRACTION

A
Thesis
Submitted to the
Jai Narain Vyas University, Jodhpur
for the award of the degree of

DOCTOR   OF   PHILOSOPHY

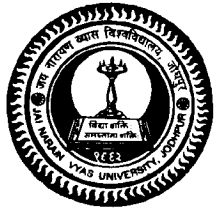

**Supervisor**
**Dr. V.S. Bansal, Ph.D.(U.K.),**
**M.S.(U.S.A.)**

**Researcher**
**K.R. Chowdhary**
Associate Professor,
Department of Computer
Science & Engineering

## Department of  Electrical Engineering
Faculty of Engineering, M.B.M. Engineering College,
Jai Narain Vyas University, Jodhpur-342011.
February 2004.

**Dr. V.S. Bansal,**

Ph.D.( U.K.), M.S.( U.S.A.)

# CERTIFICATE

It gives me great pleasure to certify that Shri K.R. Chowdhary, Associate Professor, Department of Computer Science & Engineering, has carried out the research work, reported in this thesis, under my supervision. The thesis, entitled "Natural Language Processing for Word Sense Disambiguation and Information Extraction" is an original piece of work of the researcher and has not been submitted else where or to any University in India or abroad.

I wish him all  success.

(Dr. V.S. Bansal)

Dedicated

To

my parents,

&

all  my teachers

# Acknowledgements

First and foremost, I would like to thank my supervisor, Dr. V.S. Bansal (UGC/AICTE Professor Emeritus, and former Head, Department of Electrical Engineering), for his guidance, support, and encouragement through out the period of my research work. He always gave useful advice and direction while still allowing me to go on my own way, often to realize that he was right all along. I am incredibly fortunate to have had him as an advisor all the time. I would also like to thank Professor Sushil Bhandari (former Dean Faculty of Engineering), Professor (late) D.S. Bhandari (former Head, Department of Electrical Engineering), Shri N.K. Patel (Head, Department of Electrical Engineering) and Professor V. P. Gupta (Dean Faculty of Engineering), for encouragement, support, and motivation for perusing this research work.

I am, also thankful to my children (Pratibha, Keerti, and Prabhat) who waited patiently and missed all kinds of outings for three years, and to my wife (Suman) for taking over all house hold responsibilities and playing the role of father of the children most of the times during the period of  my confinement to this research work.

(K.R. Chowdhary)

# Contents

# List of Abbreviations

| | |
|---|---|
| ACM | Association for Computing Machinery |
| *Bel* | Belief |
| *bpa* | Basic Probability Assignment |
| CFG | Context Free Grammar |
| DAG | Directed Acyclic Graph |
| D-S | Dempster-Shafer |
| DR | Document Retrieval |
| FDR | Fuzzy Document Retrieval |
| GATE | General Architecture of Text Engineering |
| *idf* | Inverse Document Frequency |
| IE | Information Extraction |
| IEEE | Institute of Electrical and Electronics Engineers, Inc. |
| IIES | Intelligent Information Extraction System |
| IR | Information Retrieval |
| KB | Knowledge Base |
| LE | Language Engineering |
| MRD | Machine Readable Dictionary |
| NE | Named Entity |
| NL | Natural Language |
| NLP | Natural Language Processing |
| POS | Parts-of-Speech |
| SDL | Structured Description Language |
| QAS | Question Answering System |
| SIGIR | Special Interest Group on Information Retrieval |
| *Synset* | Synonym's Set |
| *tf* | Term Frequency |
| WSD | Word Sense Disambiguation |
| WWW | World Wide Web |

# List of Figures

# List of Tables

# List of Algorithms

# PREFACE

With ever increasing electronically available information, developing machine tools for the extraction of information of interest has assumed considerable importance. No human can read, understand and synthesize megabytes of text on everyday basis [42]. The information and knowledge is often in the diffused form of Natural Language (NL) texts. Large varieties of information processing applications have to deal with NL texts. Intelligent Information Access Systems, Automatic Document Classifications, Machine Translation, NL Interface with machine and Digital Libraries greatly benefit by having access to the machine interpretation of texts under consideration [52], [85].

This research work deals with Natural Language Processing (NLP) and extraction of essential information in an explicit form. The most common among the information management strategies is Document Retrieval (DR) and Information Filtering [78]. DR systems may work as combine harvesters, which bring back useful material from the vast fields of raw material. With large amount of potentially useful information in hand, an Information Extraction (IE) system can then transform the raw material by refining and reducing it to a germ of original text. Consider a situation in which financial analysts are investigating properties of semiconductors. They may like to know several things, namely - the chemicals required to be deposited to produce insulating layers, the thickness of the layers, the temperature at which the layers are formed, etc. Such information is frequently available in large gamut of journals. A Document Retrieval system collects the relevant documents carrying the required information, from the repository of texts. An IE system then transforms them into information that is more readily digested and analyzed. It isolates relevant text fragments, extracts relevant information from the fragments, and then arranges together the targeted information in a coherent framework.

The goal of IE research is to establish theories and build systems that find and link the relevant information while ignoring the extraneous and irrelevant information. IE has vast potential for real-life applications. Information available in unstructured texts can be translated into traditional databases. The users can then probe through various queries.

Suppose it is required to track the profits of companies under two different groups. The relevant information in this case may include Company name, company group and its profit. An IE system that tracks news releases in this area, can update a database as new information becomes available, and can help detect trends as soon as public announcements are made. Other IE systems may process news events, including meetings of important people, formation of new companies and announcement of new products.

This can be done by:

i)      developing a language/logic, for expressing textual knowledge in a computer. It is expected that many different programs will use this knowledge, hence this representation language needs declarative semantics,

ii)      building the knowledge base by encoding the knowledge in the form of logic developed in (i) above, so that machine can use it for reasoning purpose,

iii)      developing suitable procedures/algorithms for implementing (i) and (ii) described above.

The main objective of the research work presented in this thesis is to develop a theory to explicitly and declaratively represent the body of knowledge, in a way that renders it usable by a number of programs. The thrust of the presented research work is to devise and explore various techniques of NL Texts processing so that the relevant information is extracted speedily and unambiguously.

The thesis presents a new approach for Word Sense Disambiguation using thesaurus. An illustrative example supports the effectiveness of this approach for speedy and effective disambiguation. A Document Retrieval method, based on Fuzzy Logic has been described and its application is illustrated. A question – answering system describes the operation of information extraction from the retrieved text documents. The process of information extraction for answering a query is considerably simplified by using a Structured Description Language (SDL) which is based on cardinals of queries in the form of "who", "what", "when", "where" and "why".

The modern world of information is highly complex. As the complexity of a system increases, it gets plagued by a gamut of uncertainties, viz. imprecision, incompleteness, vagueness and ambiguity. The salient features of Bayesian inference networks have been explained keeping in view their suitability for Document Retrieval and Information Extraction. The task of DR and IE in the presence of probabilistic uncertainty has been successfully carried out in this thesis using this approach.

The thesis concludes with the presentation of a novel strategy based on Dempster-Shafer theory of evidential reasoning, for document retrieval and information extraction. This strategy permits relaxation of many limitations, which are inherent in Bayesian probabilistic approach. Dempster-Shafer theory permits handling of tangled hierarchies of documents. When the data is incomplete, then probabilities are represented by Belief functions and ignorance interval. With increasing number of evidences, the ignorance level progressively decreases which makes the task of DR and IE more sharply focused.

# Chapter 1

# Literature Survey and Research Objectives

---

## 1.1 Introduction

Information processing is concerned with the representation, storage, organization of, and access to information items [86]. The information should be represented and organized in such a way as to provide the user with easy access to the information in which he/she is interested [3], [71], [79]. Unfortunately, specifying the *user's information need* is not a simple problem. Consider, for example, the following typical information need of a user:

> Find the documents containing information about college cricket teams, which: (i) were nominated by the university for National games, and (ii) have played in the Asian games, and (iii) have participated in World Olympics at least once. The information extracted from these documents must also include the names of players, their home/contact addresses, and phone/email addresses.

Clearly, this full description of the user information need cannot be used directly, to request information, from the currently available interface to Web search engines. Instead, the user must first translate this information need into a *query* representation, which may be processed by a search engine. In the most common form, this translation yields a set of *keywords* (or *index terms*), which summarizes the description of user's information need. Given a query, the goal of an information processing system is to retrieve the documents, which might be useful or relevant to the user's need.

To answer a query, the emphasis of information processing system is on retrieval of *information* as opposed to the retrieval of *data* in traditional database systems. The user of such a system is concerned more with retrieving information about a subject, where as in retrieving data, the objective is to retrieve data, which satisfies a given query. In other words, the first is more of a psychological requirement where as the second is more numerical. A data retrieval language aims at retrieving all objects, which satisfy clearly defined conditions such as those in a regular expression or in a relational algebra expression. Thus, for a data retrieval system, a single erroneous object among thousand retrieved objects means a total failure. For a system retrieving the information, however, the retrieved objects might be inaccurate, and small errors are likely to be unnoticed. The main reason for this difference is that – Information Retrieval (IR) usually deals with natural language text, which is not always well structured and could be at the same time ambiguous. On the other hand, data retrieval system such as relational database deals with data, which have a well-defined structure and semantics.

Data retrieval, while providing a solution to the user of a database system, does not solve the problem of retrieving information about a subject or topic. To be effective in its attempt to satisfy the user's information need, the information retrieval system must some how interpret the contents of the information carrying documents and rank them according to a degree of relevance to the user query. This interpretation of document content involves extracting syntactic and semantic information from the document text and using it to match the user information need [2]. The difficulty is not only how to extract this information but also how to use it to decide the relevance. In fact, the primary goal of an information retrieval system is to retrieve all the documents from a given collection, which are relevant to a user's query, through a process called Document Retrieval (DR), and at the same time retrieving as few non-relevant documents. A detailed survey of the earliest works in DR appeared in [14], [32].

Having retrieved the documents containing the relevant information, it is followed by locating the precisely required information in the retrieved documents and extracting it, through the process called Information Extraction (IE) [21]. The IE systems generally work by detecting patterns in the text that help identify significant information and fill up some templates of information. The kind of data for IE falls into so-called *semi-structured text* category, and usually contains information for most of the expected fields in some order. In the semi-structured text, the positions of the information fields are fixed and values are limited to pre-defined set. A bibliography is an example of semi-structured text. At the opposite end lies the task of extracting information from *free text*, which though unstructured, is assumed grammatically correct. Here IE systems rely more on *syntactic*, *semantic* and *discourse* knowledge in order to assemble relevant information, which is scattered all over the document.

IE algorithms face different challenges depending on the extraction targets and kind of text embedded in. The targets may be simple, carrying the required information in single slot, or they may be distributed in the document in multiple slots.

It seems reasonable to believe that one can produce the accurate extraction of information from documents if one can design programs, which can understand the document. This is only possible by in depth Natural Language Processing [4], [28], [36], [41]. However, In-depth Natural Language Processing is a complex endeavor that can unduly strain computational resources [30].

Many of the limitations imposed by word-based techniques can be overcome with IE techniques. In particular, one can consistently achieve high precision because the IE approach is sensitive to context. Linguistic phrases and context surrounding the phrases are recognized easily and handled naturally. Thus, the portions of text, which are not relevant, can be effectively ignored. This considerably simplifies the job of a Natural Language Processing (NLP) system. This approach of information extraction is computationally less expensive compared to a full-blown NLP System as the phrases and sentences, which are non-relevant to the domain, are ignored. In addition, since the system is only concerned with domain specific portions of the text, difficult problems in NLP are often simplified. IE is, therefore, a practical and feasible technology that has wide spread applications.

## 1.2 Past Research in Document Retrieval and Information Extraction

### 1.2.1 Overview of Technology Trends

Several events have recently occurred which have had a major effect on research in this area [20]. First, computer hardware is more capable of running sophisticated search algorithms making use of massive amounts of data, with acceptable response time. Second, Internet access, such as World-Wide-Web (WWW), brings new search requirements from untrained users who demand user-friendly, effective text searching systems [47], [61]. These two events have contributed to creating the interest in accelerating research to produce more effective search methodologies, which make use of natural language processing techniques. This has resulted in accelerated research in the area of information retrieval, dominated by statistical methods to automatically match natural language user queries against the text database records [13]. Yet, to date NLP techniques have not significantly improved the performance to retrieve the Information, although much effort has been put in this direction [49]. The motivation and drive for NLP techniques for information retrieval is mostly intuitive. Users decide on the relevance of documents by reading and analyzing them, and if one can automate document analysis, this should help in the process of deciding on the document relevance.

Most researchers in the information retrieval community believe that retrieval effectiveness is easily improved by means of statistical methods rather than by pure NLP-based approaches, like context-free grammars, parsing and parts-of-speech tagging. The fact that only a fraction of information retrieval research is based on extensive natural language based process techniques indicates that NLP techniques do not dominate the current thrust of information retrieval research. Yet, NLP resources used in extracting information from text, namely thesauri, lexicons, dictionaries, and proper name databases, are used regularly in document retrieval research. It suggests, therefore, that NLP resources have more impact on DR effectiveness than the NLP techniques. One reason for this is that the current NLP techniques are not designed to handle large amounts of texts from many different domains at the same time. However, DR systems are required to work on broad domains in order to be useful. Thus, there is an inherent granularity mismatch between the statistical techniques used in DR and the linguistic techniques used in NLP. The statistical techniques attempt to match the *rough* statistical approximation of a text record to a query. Further, the refinement of this process using *fine-grained* NLP techniques often adds only noise to the matching process, or fails because of the vagueness of the natural language [20].

What is needed is document retrieval by taking advantage of NLP resources. Among these resources are online Machine Readable Dictionaries (MRDs) and thesaurus. MRDs are special texts whose subject matter is a language, or a pair of languages in the case of a bilingual dictionary. The purpose of dictionary is to provide a wide range of information about words - etymology, pronunciation, stress, morphology and syntax - to give definitions of senses of words. This process supplies knowledge not just about the language, but also about the world itself.

## 1.2.2 Different Methodologies for DR and IE

Moreover, even if the user does have questions in mind, the aim is to retrieve overall information so these questions, as well as other questions prompted by reading the documents, are answered. This means that Document Retrieval must find relationships between the information needs of users and information in the documents - both considered in general sense and neither directly available to the system [19], [39]. Retrieval depends on *indexing*, that is on some means of indicating what documents are. Indexing requires an *indexing language* with a *term* vocabulary and a method for constructing requests and document *descriptions*. Indexing is the basis for retrieving documents relevant to the user's need. It is supported by a search apparatus specifying conditions for a match between request and document descriptions, as well as modulation methods, to alter these descriptions if no match is made immediately. These tests have shown that indexing documents by individual terms corresponding to words or word *stems* produces results at least as good as those produced when indexing by controlled vocabularies.

In contrast, the statistical DR methods, which simplify the representations based on single terms, have provided significant improvement over such alternative approaches as Boolean querying. Statistical DR methods rank documents based on their similarity to the query or an estimate of the probability of their relevance to the query. Both query and document are treated as collections of numerically weighted terms. The query can be an arbitrator textual statement of the user's information need or might even be a sample document. Statistical DR methods assign higher numeric weights to terms showing evidence of being good content indicators, causing them to have greater influence on the ranking of the documents. The number of occurrences of a term in a document, in the query and in the set of documents as a whole, may all be taken into account when computing the influence the term should have on document's score. In addition, if the user indicates that certain retrieved documents are relevant, this information can be used to re-weight and alter the query term through a process called *relevance feedback* [10].

Recent IE research emphasizes the importance of many neglected areas of NLP while demonstrating that simple methods are adequate for many tasks involved in analyzing the text. Most IE systems depend on off-line components to produce the data and rules that direct on-line processing. Automation and speed-up of these components are critical for rapidly porting systems to new domains.

Two aspects of preprocessing seem to be common to many IE systems:

- Parts-of-speech tagging programs to allow preliminary recognition of phrasal units in sentences.
- Special-purpose rules to recognize semantic classes of phrasal units, including company names, places, people's names, currencies, equipment names, etc.

The real text is rich in proper names, and expressions for dates, values and measurements. These phrasal units are used productively, usually posing no problems to human readers. However, when looking at the *Tipster* rules for a company name, we find ambiguity when a company reported in a form such as: "Tata Steel of India announced today …", where it is difficult to decide whether, "of India" is part of the name or not? The nature of these units require that they be not held in phrasal lexicon, so that an automatic system can recognize them either through a context or through a patterns of sub-units in the text. For some applications, a database of texts and lexicon lists taken from these texts provide adequate coverage. The same recognition capabilities can also support IR needs [21].

An increasing number of research efforts have recently started under the name of *Language Engineering* (LE) rather than with the standard names like *Natural Language Processing* or *Computational Linguistics* [24]. A typical implementation of this is *General Architecture for Text Engineering* (GATE) [25]. It provides a software infrastructure for NLP, on top of which heterogeneous NLP processing modules may be evaluated and refined individually, or may be combined into a larger application system. GATE aims to support both researchers and developers working on component technologies like, parsing, tagging, and morphological analysis, as well as those working on end user applications like, information extraction, text summarization and second language learning.

### 1.2.3 Current Techniques for Information Extraction

The texts used for information extraction fall in two categories – structured texts and free texts [60]. In structured texts, the information positions remain fixed and the values of this information are limited to pre-defined set. Consequently, the IE systems focus on specifying delimiters and the order of these fields. The free-text is unstructured, however grammatically correct. The IE systems rely more on syntactic, semantic and worlds knowledge in order to assemble the relevant information [37], [93]. The syntactic processing mostly involves identification of nouns and verbs. Following sections discuss the syntactic and semantic features of Natural Language (NL) text and their roles for extraction of information.

**a. Object Identification**

The initial stage is to identify objects that are fundamental in the relevant documents. For the system to be generic to any domain, a technique is necessary to automatically identify the relevant objects in the context of the corpus. The most difficult scenario is to identify/classify all type of entities automatically, called *named entity identification.* Different examples can be – names of persons, organizations, locations, times, dates and currencies. To identify the multitude of different entity classes occurring in the vast range of alternative domains, it is necessary to utilize large amount of world knowledge, which can provide substantial collections of related words. Having identified the named entities, the next task is to find out the entities that are fundamental objects within the required task. The entities, which are less important, will represent the features associated with the objects or their relationships.

The WordNet is a popular lexical resource, where the knowledge is hierarchically structured in the style of thesaurus [55]. The objective of WordNet is to provide an aid for searching dictionaries conceptually. It consists of a large number of word forms organized into word meanings or set of synonyms, called synsets. This resource identifies useful relationship between words for IE; in addition, a naive form of sense disambiguation is automatically carried out by identifying the node in the hierarchy where majority of nodes intersect in a semantic network representation.

**b. Role of Nouns and Verbs in IE**

The documents can be classified based on the role of verbs and nouns present in the documents [45]. The listing below shows the ontological categories that express the fundamental conceptual components of propositions.

[THING]     [DIRECTION]     [ACTION]

[PLACE]     [EVENT]     [AMOUNT]

Each category permits the formation of a *What* – question. For example, for [THING], "What did you buy?" can be answered by the noun "a book". Many Natural language analysis systems focus on noun phrases in order to identify information on *who, what,* and *where*. In text summarization, the multiword noun phrases are the target of attention [50]. The named entity task, which identifies nouns and noun phrases, has generated number of projects [51]. However, the named entity task provides no information about *what happened* in the document, i.e., about the *event* or *action*. The *what* question for [ACTION] and [EVENT] can only be answered by verbal constructions like, in the question "What did you do?", where the response must be a verb, e.g., jog, write, fall, etc.

The distinction in the ontological categories of nouns and verbs is reflected in IE systems. Given the noun phrases *fares* and *Indian Airlines* that occur with in a particular article, the reader will know what the story is about. However, the reader will not know the [EVENT], i.e., what happened to the *fares* or to *Indian Airlines*. Did the air prices *rise*, *fall* or *stabilize*? These verbs are most typically applicable to prices and the event. Little progress has been made so far on ways to efficiently utilize the verbal information in the source text [45].

## 1.2.4 Word Sense Disambiguation

Many words/phrases in a text of natural language have more than one meaning. When a person understands a sentence with an ambiguous word in it, that understanding is built based on just one of the meanings. Thus, as part of the human language understanding process, an appropriate meaning should be chosen from the range of possibilities. The concept of disambiguation, however, is obscure. Humans often cannot agree about which of a given collection of senses is being used in a particular sentence. Lexicographers themselves do not agree about the number of senses for a given word, nor about the way, a word's use should be divided into senses.

NLP and information retrieval need automatic methods of disambiguation [1], [88]. In machine translation, one must disambiguate the input text in order to yield a correct translation to another language. Information retrieval systems might be more effective if they were able to disambiguate the words in a query and in stored documents. In addition, the corpus analysis and lexicography could become more automated if words in the corpora were disambiguated. Senses are often chosen in an ad hoc way, usually with fewer and broader senses than would be found in most standard dictionaries.

All the following have been implemented as the basis of experimental Word Sense Disambiguation (WSD) at various times: parts-of-speech tagging, semantic preferences, collating items or classes, dictionary definitions and synonym list [72]. Most recent research on WSD has adopted a corpus-based learning approach [34]. In addition, many different learning approaches have been used, which includes neural networks, probabilistic algorithms, decisions lists and exemplar-based learning algorithms. The heart of exemplar-based learning is a measure of the similarity or distance between two examples [31].

## 1.3 Research Objectives

One of the principle objectives of the proposed work is to develop a theory to explicitly and declaratively represent information in a way that renders it usable by number of programs. This is in principle no different from the way in which a book "structures" information in tables and charts on the assumption that a wide audience will be able to use that data in innumerable ways. Since programs do not read and understand natural language texts, the structuring presented must be rather different, in particular, much less ambiguous.

It is proposed that, to process natural language text one needs three things: -

   i)      a language or logic, for expressing information in a computer. Since, it is proposed that many different programs will use this information, the representation language needs declarative semantics

   ii)     a structuring for the information in the language developed in (i) above, so that machine or man can apply to it the reasoning mechanism

iii)     a set of procedures or algorithms for using the information represented in (ii),
         for the purpose of IE

One of the thrusts of the presented research work is to devise and explore various information structuring techniques with a view to process the NL texts unambiguously for the purpose of Information Extraction. This is followed by restructuring of NL texts for easy and fast extraction of information to find out the answer of the queries cardinals *Who, What, When, Where* and *Why*. The input text can be research papers, media reports, advertisements, technical reports or any other information in a restricted domain. For example, in a large digital library, different queries can be submitted to extract the information, like, "*Who* are the authors in a specified field (domain)", "*What* are their main contributions", "*When* did they contributed the work", "*What* methodology they have used" and "*What* are their findings"? Similar applications are possible in media reports, advertisements and technical reports or other similar domains.

The decision making for the selection of relevant documents, followed by the extraction of information to answer the queries precisely is a complex task. To accomplish this, following topics have been presented in this thesis:

- Investigation of new disambiguation techniques with their merits and demerits
- Fuzzy Logic based Document Retrieval and Information Extraction
- Structured Description Language for Question-Answering
- Document Retrieval and Information Extraction using Bayesian probabilistic approach
- Document and Information Retrieval using Dempster-Shafer Theory of Evidence

The out come of this research is expected to contribute the following:-

i)      Processing of the NL texts with speed and without ambiguity

ii)     Extraction of vital Information from a set of documents for answering queries

iii)    Developing useful tools to identify and understand the focal points in NL texts

## 1.4 Outlines of the Chapters

**Chapter 1:** This chapter presents basic concepts of Document Retrieval and Information Extraction. A comprehensive review of the research work published in the area of DR and IE is given. This chapter identifies the following broad topics of research:

- New disambiguation techniques
- Fuzzy logic based DR and IE
- Structured Description Language for Question Answering
- Document Retrieval and Information Extraction using Bayesian and Dempster-Shafer theories of probability

**Chapter 2:** A general model of Document Retrieval and Information Extraction has been presented. Various challenges in these two areas are outlined. Methods for text representation and text processing including the concepts of index term selection and use of thesaurus are explained. Finally, a measure for the evaluation of Document Retrieval and Information Extraction is presented.

**Chapter 3:** This chapter deals with disambiguation, which is an important aspect of Natural Language processing. A WordNet based model for disambiguation is presented. An illustrative example for Word Sense Disambiguation brings into focus the effectiveness of WordNet based approach for disambiguation.

**Chapter 4:** Fuzzy Logic based document retrieval is described in this chapter. Basic concepts of fuzzy sets, fuzzy logic and fuzzy relations are presented. An illustrative example is given, which describes an approach of document retrieval using fuzzy logic.

**Chapter 5:** This chapter describes Information Extraction using Structured Description Language. The sentences in a text are represented in the form of transition diagrams us-ing basic *wh-terms who, what, when, where* and *why*. A question answering system is illustrated for answering such queries, which may need multiple documents.

**Chapter 6:** This chapter presents Bayesian probability theory for DR and IE with a view to answer user queries. This theory accounts for various uncertainties, which are invariably present in the natural language texts. A mathematical model for Bayesian probability based DR and IE has been developed and implemented with illustrative examples.

**Chapter 7:** This chapter presents a new approach for Document Retrieval and Information Extraction using Dempster-Shafer theory of evidential reasoning. Basic concepts of Dempster-Shafer theory including the combination rules are described. Illustrative examples for the retrieval of documents and information extraction are given.

**Chapter 8:** This chapter summarizes the important results in the following area of research: -

- Word Sense Disambiguation
- Fuzzy logic based DR
- Structured Description Language for question answering
- Document retrieval and Information Extraction using Bayesian theory of probability
- Document Retrieval and Information Extraction using evidential reasoning

Suggestions for future research work are outlined. A model for Intelligent Information Extraction System (IIES) using feedback concepts is introduced for future research work.

# Chapter  2

# Basic Concepts and Challenges of Document Retrieval and Information Extraction

---

## 2.1 Introduction

The techniques used for Information Extraction (IE), Text categorization, text routing and text filtering systems fall largely on a common platform [73]. However, the information needed from an IE system is of short term, because a user puts the query to the system and on extracting the information, the job of IE is over [14]. Most of the techniques discussed in this chapter are meant for Document Retrieval (DR). However, these may be applied to all the four cases stated above. Text categorization labels the text automatically based on a set of predefined categories and DR retrieves text documents as per the ad hoc user queries. Text routing systems accept a set of profiles or categories of interest, and automatically route texts that satisfy a profile, to the corresponding user.  Text filtering systems allow only certain texts to pass through the filter.  The filter specifies topics that interest the user and only such topics are passed to the user [80].

## 2.2    General Model of DR and IE

Document Retrieval delivers to the user those documents, which enable him to satisfy his information *needs*.  Figure 2.1 shows the model of a DR and IE system [5]. It indicates the basic entities and processes in a DR-IE system.  In this model, a person with some goals and intentions related to the task finds that these goals cannot be attained because the person's knowledge is inadequate. A characteristic of such a "problematic situation" is information need, which prompts the person to engage in active information-seeking behavior, such as submitting a *query* to the DR-IE system. The query, which is represented in a language by the system, is representation of information need. Due to inherent imprecision in the representation of information need, the query in a DR system is always *approximate* and *imperfect*.

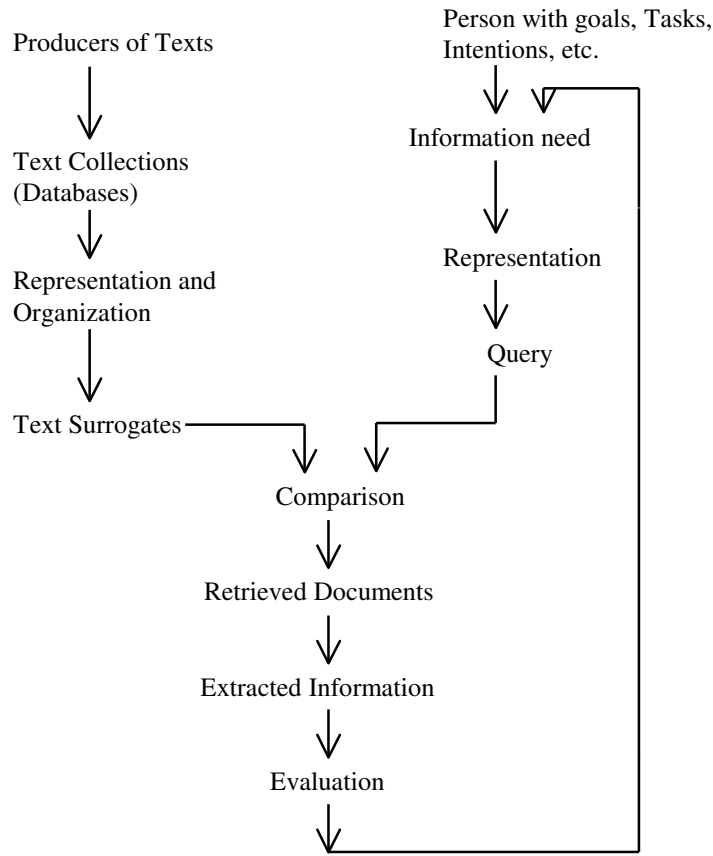

Figure 2.1:General Model of DR-IE.

The DR-IE model also shows the information resources that the user of this system will eventually have access. Here the model considers the *producers* of texts, the *groupings* of texts into *collections* (also called *text databases*), *representations* of texts, and *organization* of these representations into databases of *text surrogates*.  The process of representing the meaning of texts in a form more amenable to processing by computer is of central importance. The typical surrogate consists of a set of index *terms* or *keywords*.

The comparison of a query and surrogates leads to the selection of possibly relevant texts. These texts are retrieved and then used for the extraction of precisely needed information. The extracted information is *evaluated*. The evaluation may also lead to *modification* of query, i.e., the information need. The process of query modification through the user evaluation is called *relevance feedback* in Information Retrieval (IR) [12], [29], [38].

The general model of DR-IE can be further categorized into three different types. These are *Boolean*, *Vector space* and *Probabilistic* retrieval models. The first of these is based on what is called the "exact match" principle, the other two on the concept of "best match". The Boolean model retrieves the text based on the concept of an exact match of a query specification with one or more text surrogates. The term Boolean is used because the query specifications are expressed as words or phrases, which are combined using the standard operators of Boolean Logic. In this model, all surrogates, or more generally texts, containing the combination of words or phrases, specified in the query, are retrieved, and no distinction is made amongst the retrieved documents. Thus, the result of the comparison operation in the Boolean retrieval is a *partition* of the text database into a set of relevant documents, and a set of non-relevant documents. The Boolean exact-match retrieval model is the standard model for current large-scale operational DR systems, including the Internet search. The major problem with this model is that it does not allow for any form of relevance ranking of retrieved document set.

The vector space model treats the texts and queries as vectors in a multi-dimensional space, the dimensions of which are the words used to represent the texts. Queries and text are compared based on their vectors. The assumption is that the more similar a vector representing a text to a query vector, the more likely that the text is relevant to that query. The terms or dimensions of a query, or text representation, can be weighted, to take account of their importance. These weights are computed based on a statistical distribution of the terms in the text databases.

Probabilistic DR systems rank the text documents in the order of their probability of relevance to the query, given all the evidence available. This principle considers that rep-resentation of both the information need and the text are uncertain; and that the relevance relationship between them is imprecise. The probabilistic retrieval model suggests about the variety of sources of evidence, which could be used to estimate the probability of relevance of text document to a query. The most typical source of such evidence is the statistical distribution of terms in the texts.

## 2.3 Challenges of DR and IE

Information retrieval is a complex task. Query-based DR systems must be able to accept a query about any topic and find relevant texts that contain the specified information about the query. Often texts are very large, and some times IR systems are required to operate in real-time, which demands that DR and IE should be fast. In addition, the search is conducted on the Natural Language (NL) text, which inherently has many ambiguities and imprecision. The following part discusses some of the challenges faced for DR and IE.

Synonymy occurs in the text when different words or phrases mean essentially the same thing. For example, a user is looking for articles on corporate investments. One simple strategy is to look for word "invest". Nevertheless, this strategy will fail to retrieve many texts, which concerns corporate investment, because there is lot of different ways to refer to investments. The words "finance", "fund", "support", "capitalize" etc., also refer to investments. Thus, the concerning text may not be retrieved by the above query.

NL texts contain many words and phrases that have similar meanings. It is often difficult for users to provide all the words in the query. To address this problem, DR systems expand the query to include all the synonym words for each query term using thesaurus.

Another challenge is *Polysemy*. It occurs when a single word has more than one meaning. For example, the word "shot" may have different meanings when used in different contexts, as in the following cases:

A shooting, in  - He shot at tiger.
An attempt, in - I took a shot at playing the lottery.

A photograph, in - He took a nice shot of Red Fort.

Polysemy has the effect of leading a system to return documents irrelevant to user's information need. Word sense disambiguation techniques are used to resolve the issue of polysemy.

Yet, another challenge lies in identifying *semantics and Role Relationships*. Some information can be identified only through semantics. Suppose a user is interested in finding out the names of teachers who are teaching the courses in the area of "Artificial Intelligence". First, the system must know, what are the courses related to Artificial Intelligence. Second, the system must be able to identify the role-relationship that depends on the context, i.e., the person must be Lecturer, Reader or Professor. Recognizing the role-relationship requires natural language processing to analyze the text conceptually and identify the appropriate relationships [16].

Texts come in all shapes and sizes. There are two parameters which are important to DR systems: (i) *Length of the text* and (ii) *Cohesiveness*. Some texts are short (e.g., memos) and some texts are very long (e.g., books). Short documents usually discuss only a single topic. Abstracts are good examples of short texts, which are cohesive also. Longer documents, however, usually discuss many different topics in separate sections. Even medium length documents, such as newspapers and magazine articles, typically discuss several varying topics in different paragraphs. Until recently, however, most of the standard DR collections consist of short, cohesive documents (often technical abstracts). The current sets of information are fairly large and heterogeneous. As a result, the problem of subtopic identification is receiving increasing attention [9].

## 2.4 Text Representation

The most obvious option for searching text for a basic query is scanning of text sequentially to find the occurrences of text patterns in the document. Based on this, the relevancy of the document can be judged. However, the text collections for information retrieval and extraction are invariably large, and the search may be in real-time. Thus a sequential search, even using the most efficient methods and fast computing machines, is inadequate. Therefore, the text documents are represented in such forms, which reduce the effort for the search.

One of the frequently used text representation technique is *Inverted File* format. It is a data structure for efficiently indexing texts using the words in the texts. One can view an inverted file as a structure of list of words where each word is followed by an identifier of text that contains the word. In addition, the number of occurrences of each word in the text is stored in this structure. Figure 2.2 illustrates that word 'AI' appears five times in doc-12. Similarly, the words 'reasoning' and 'inferring' appear in the document names mentioned against each along with the frequency of their occurrences. The inverted file helps a DR system to quickly determine which documents contain a given set of words, and how often each word appears in the document. Other information can also be stored in the inverted file such as location of each word in the text using the pointers, to directly reach to those locations for further text analysis.

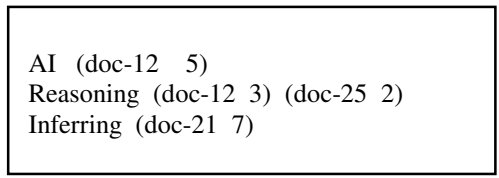

AI  (doc-12  5)
Reasoning  (doc-12  3)  (doc-25  2)
Inferring  (doc-21  7)

Figure 2.2.   Inverted File Representation.

The inverted file can be implemented using a data structure such as sorted array, hash table, b-tree, as well as using the combination of these structures.

Word based retrieval systems are some times augmented with phrase-based indexing in order to represent explicitly the multi-word phrases in a query [23]. Phrases may have a meaning different from their constituent words. For example, if a user wants to retrieve documents about "Artificial Intelligence", the entire phrase should be used to represent the concept. If the words "Artificial" and "Intelligence" are taken independently, many irrelevant texts may be retrieved.

Recognizing phrases often requires parts-of-speech tagging and phrase bracketing, which are not supported by most DR systems. One way to achieve phrase-based indexing is to use proximity measures to specify the acceptable distance between two related words.  In this, two or more words are considered a phrase if they are separated by no more then N words *distance* in the text.  If N is zero, the words must be adjacent.  If N is one then they are separated by one word. The phrases "class teacher" and "hide and seek" represents phrases where N is zero and one respectively. The advantage of this approach is that the inverted file can be easily adapted for phrase recognition by simply including the location of the words in the text using a pointer.  However, the disadvantage is that the size of the inverted file will increase if too many phrase-based index terms are present.

## 2.5   Text Processing

To arrive to the representation, the text database is preprocessed and stored in a structure, which helps in fast searching. This preprocessed form is called *text representation*, because this is the form precisely seen by the DR system (figure 2.3).

The traditional approaches for DR are word-based [69], [82]. These treat each text as a collection of words and retrieve documents by searching for relevant words or phrases. *Inverted files, stop word lists* and *stemming algorithms* are frequently used in word-based DR systems.

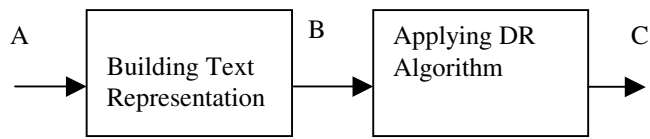

A -  Input Text
B -  Text in inverted File Format
C -  Retrieved Text

Figure 2.3.  DR System

### 2.5.1 Stopwords and Stemming

Words that are too frequent among the documents in the collection are less informative, thus they are not good discriminators for documents. For example, a word that occurs in 80% of the documents in a collection is not of much use for the purpose of retrieval. Such words are referred to as *stop words*. These are to be *filtered* out before the text is processed for DR. This also reduces the size of the document index. The major stop words are - articles, prepositions, and conjunctions.

Word-based DR techniques retrieve documents by searching for texts that contain relevant words. However, looking for an exact match often does not solve the problem, due to different morphological variants of each word. Inflectional variants are different syntactic forms for a single part-of-speech, such as verb tenses (look, looked, looking, looks), and singular and plural nouns (book, books) [33]. Stemming algorithms are often used to reduce words to a root form called *stem*. One of the most commonly used stemming algorithms is *Porter's* algorithm [62]. The *stems*, instead of original words are used to construct the inverted tree. The two main advantages of stemming algorithms are space efficiency and retrieval generality. The size of the inverted file can be reduced dramatically because many different words that are indexed under the same stem require only a single entry in the inverted file. In addition, the generality is enhanced, since query terms no longer have to match the text exactly. The stemming algorithm is also applied to the query so that only the stems are matched.

### 2.5.2 Index Terms Selection

If, after the elimination of stop words and stemming, all the distinct words in the text are used for creating an index, the index will not be meaningful and at the same time, it will be too large. Instead, an abstract view of the text is adopted to create the index. Thus, only the selected terms are used for indexing a text. In bibliographic sciences, this selection is done by a specialist. However, looking to the size of documents used for DR the selection of index terms is automated.

Different approaches are used for selection of index terms. A good approach is the identification of noun groups. A sentence in NL text is usually composed of nouns, pronouns, articles, verbs, adjectives, adverbs, and connectives. While the words in each grammatical class are used with a particular purpose, it is argued that most of the semantics is carried out due to the noun words. Thus, index is built using the noun words. This is done through systematic elimination of articles, verbs, adjectives, adverbs, pronouns and connectives. Since it is common to combine two or three nouns in a single component (e.g., *Information Technology*), it makes sense to cluster nouns, which appear adjacent in the text, into a single indexing component (or concept). Thus, instead of using nouns as index terms, noun groups are used as index terms. A noun group is a set of nouns whose syntactic distance in the text, measured in terms of number of words between two words, does not exceed a particular threshold. This may be either one or two word. Next, out of number of nouns in a text document, those terms, which have highest frequency and are not common among the other text documents are taken as indexing terms. These nouns are potential candidates for text document representation.

Assigning weights to terms in a document and those in a query vector has enormous impact on the effectiveness of a retrieval system. The two factors that are critical in deriving the effective term weights are: Term Frequency (*tf*) in a single document, and the distribution of terms across a collection. The terms, which occur more frequently with in a document, are considered to have higher weights, and the corresponding document has higher relevance to the information need. Second factor is distribution of terms across the whole collection. The terms, which are limited to few documents, are good for discrimination of these documents from rest of the collection, compared to the terms, which are distributed uniformly across the entire collection. A measure, which favors the terms that occur in the fewer documents, is useful. The fraction $N/n_t$, where $N$ is total number of documents in the collection, and $n_t$ is the number of documents in which term $t$ occurs, provides this measure. Fewer the documents in which this term occurs, higher is the weight of the term. Due to large number of documents in many collections, it is used with a logarithmic function, called Inverse Document Frequency (*idf*) term weight, and expressed by equation (2.1) [76].

$$idf_t \;\; = \;\; \log\!\left(\frac{N}{n_t}\right) \qquad\qquad \ldots \qquad (2.1)$$

Combining the *tf* factor with *idf* factor results in a scheme known as *tf*\**idf* weighing. Considering the $i^{th}$ term in the query vector as $t_i$, its weight in the document $d_j$ in a collection of N documents is given by:

$$w_{i,j} \;\; = \;\; tf_{i,j} \;\; * \;\; idf_i \qquad\qquad \ldots \qquad (2.2)$$

In the above, $tf_{i,j}$ is term frequency of term $t_i$ in the document $d_j$, and $idf_i$ is Inverse Document Frequency of $t_i$ in the collection of documents.

In many ad hoc retrieval systems, such as web search engines, user queries are not very much like documents at all, because their length is just 2 to 3 key words. In such a situation, the raw term frequency (*tf*) is not a very useful factor; instead, the following formula is preferred for weighing the query terms [71]

$$w_{i,j} \;\; = \;\; \left(0.5 \;\; + \;\; \frac{0.5 tf_{i,j}}{\max_l tf_{l,j}}\right) \;\; * \;\; idf_i \qquad\qquad \ldots \qquad (2.3)$$

where $0.5 tf_{i,j}/\max_l tf_{l,j}$ is *normalized* term frequency of query term $t_i$ in document $d_j$ and $\max_l tf_{l,j}$ denotes the frequency of the most frequent term $t_l$ in document $d_j$.

## 2.5.3 Thesaurus

In its simplest form the thesaurus consists of - (i) a precompiled list of words in a given domain of knowledge, and (ii) a set of related words for each word in this list. Related words are derived from synonimity relationship. For example, an entry in the *Webster's* thesaurus looks like this:

> *Complexion (SYN.) paint, pigment, hue, color, dye, stain, tincture,*
> *tinge, tint, shade.*

The thesaurus is meant to provide standard vocabulary for indexing and searching, and assist users in locating terms for proper query formulation. It also provides classified hierarchies that allow the broadening and narrowing of the query request according to the needs of the user. The motivation behind building a thesaurus is based on the idea of using a controlled vocabulary for indexing and searching [59]. A controlled vocabulary presents important advantages, like – normalization of indexing concepts, reduction of noise and identification of indexing terms with a clear semantic meaning [64], [84].

## 2.6 Evaluation of DR and IE

DR and IE systems focus on "aboutness" or "appropriateness" to the requirement of information need. The effectiveness of DR is a measure of the ability of the system to satisfy the user in terms of the relevance of the document retrieved. The effectiveness of any IR system is measured by its characteristics, namely - *precision, recall,* and *fallout* [7], [43], [66], [87]. The recall is the fraction of the relevant documents, which have been retrieved, and precision is measured as the fraction of the relevant documents to retrieved documents. Fallout is measure of systems ability to ignore spurious information in the text. It is defined as non-relevant documents retrieved as a fraction of total non-relevant documents.

Let the relevant documents be represented by set *A* and retrieved documents by the set *B*. Then $\overline{A}$ and $\overline{B}$ are respectively the sets of non-relevant and non-retrieved documents. Therefore, it is possible to construct four sets using the various combinations of above, as shown in Table 2.1.

Table 2.1: Measuring effectiveness of retrieval system.

|  | Relevant Document set ( $A$ ) | Non-relevant Document set ( $\overline{A}$ ) |
|---|---|---|
| Retrieved Document set ( $B$ ) | $A \ \cap \ B$ | $\overline{A} \ \cap \ B$ |
| Non-retrieved Document set ( $\overline{B}$ ) | $A \ \cap \ \overline{B}$ | $\overline{A} \ \cap \ \overline{B}$ |

The measures of effectiveness of a retrieval system can be defined as follows:

$$\text{Re}\,call \ = \ \frac{|A \cap B|}{|A|} \qquad\qquad \ldots \qquad (2.4)$$

$$\text{Pr}\,ecision \ = \ \frac{|A \cap B|}{|B|} \qquad\qquad \ldots \qquad (2.5)$$

$$Fallout \ = \ \frac{|\overline{A} \cap B|}{|\overline{A}|} \qquad\qquad \ldots \qquad (2.6)$$

The above definition of precision and recall indicate that they are antagonistic to one another. A conservative system that strives for perfection in terms of precision will invariably lower its recall score. Similarly, a system that strives for coverage will get more things wrong, thus lowering the precision score.

The information extraction system's performance measures are defined on similar lines to that for IR. *Recall is* a measure of how much relevant information the system has extracted from the text, and *precision* is a measure of how much of the information that the system returned is actually correct. If $A$ is set of correct answers in the collection, $B$ is set of answers returned by the system, and $\overline{A}$, $\overline{B}$ have their conventional meanings, then the definitions of (2.4) to (2.6) also apply to precision, recall, and fallout measurements for IE.

A single measure which combines precision and recall, called *harmonic mean* of precision and recall [75], can be expressed as follows:

$$H(i) \;=\; \frac{2}{\frac{1}{r(i)} + \frac{1}{p(i)}} \qquad\qquad \dots \qquad (2.7)$$

In above, $r(i)$ is the recall for the document $d_i$ in the ranking, $p(i)$ is the precision for the document $d_i$ in the ranking and $H(i)$ is harmonic mean of $r(i)$ and $p(i)$. The function H assumes the values in the interval [0, 1]. It is zero when no relevant document has been retrieved. When all the retrieved documents are relevant, and all the relevant documents have been retrieved, its value is 1. The harmonic mean assumes high value when both the precision and recall are high. Therefore, any attempt to achieve the high value for H amounts to best possible performance with respect to precision and recall.

## 2.7  Discussion

This chapter presented a survey of a gamut of emerging techniques for information representation, document retrieval and information extraction. The previous work done in each area has also been highlighted. It may be pointed out that the choice of a technique is highly problem specific. Often a hybrid approach, judiciously blending apparently different techniques, provides improved results in the form of faster speed and better comprehension.

# Chapter 3

# Disambiguation

---

## 3.1 Introduction

The word-sense ambiguity due to *polysemy* is a major barrier for many systems that accept Natural Language (NL) inputs [55]. For example, an English language word may carry different senses in different contexts. Therefore, the systems for machine translation must be able to assess the sense, which the author possibly has in mind. In Document Retrieval (DR) and Information Extraction (IE), a query intended to elicit material relevant to one sense of a polysemous word may elicit unwanted material relevant to other senses of that word. Choosing among the alternative senses of a polysemous word is a matter of distinguishing between different sets of linguistic contexts in which the word form can be used to express the word sense. Humans are quite skillful in making such distinctions. For example, the following statements are ambiguous:

    *(i)*     *Bat in his hands flies high*
    *(ii)*    *Crane is in the field*

In the first case, it is not clear whether *bat* stands for an instrument for playing games, or a special kind of bird. Similarly, in the second it is not clear whether *crane* stands for lifting machine or a bird with long neck.

Following are the basic approaches for disambiguation [53]:

1. Word Sense Disambiguation (WSD) based synonyms information provided by the Machine Readable Dictionaries (MRDs) and thesaurus [56].

2. WSD program learns the necessary disambiguation knowledge from a large sense-tagged corpus, in which word occurrences have been tagged manually with senses from some wide coverage dictionary, such as Longman's Dictionary of Contemporary English (LDOCE) or WordNet. After learning on sense tagged corpus, in which all the occurrences of a word has been correctly tagged, the WSD program assigns the correct sense to the word.

Following are the works already carried out in Word Sense Disambiguation. *Sanderson and Rijsbergen* use artificially ambiguous words called pseudowords [72]. *Krovetz and Croft* attempt to resolve the lexical ambiguity using LDOCE [48]. *Rilof* and *Lehnert* have used training corpus for disambiguation in the application of automatic text classification [67]. *Voorhees has* used WordNet for disambiguation in the text retrieval applications making use of stem vectors [83]. In addition, *Roth has* used statistics based machine learning approaches for disambiguation [68]. Further, Crovetz *and Croft* in [48] reports that "…little quantitative information is available about the extent of the problem or about the impact that the disambiguation has on information retrieval systems".

An algorithm for word sense identification must distinguish between the sets of linguistic contexts, raising the question of how much context is required. There are a number of ways to define linguistic contexts. In this research, sentential context has been used, i.e., two words co-occur in the same sentence if their contexts are same. Therefore, sense identification is a matter of disambiguating among the sets of sentential contexts. A model is also presented for resolving the semantic disambiguity caused due to polysemic nature of keywords in the query. The valid sense of an ambiguous word is resolved based on the closeness of its sense to the context of the sentence. Successful disambiguation of the words prior to Information Retrieval (IR), like *crane* and *bat* in the above example, would resolve the problem of retrieving the non-relevant documents.

The work carried out in this research uses an approach of MRDs and thesaurus. Following is the hypothesis used for this: Representation of word senses in Network form shows the association, and thus the relations between different words. This helps in fast and easy navigation through the senses, which in turn helps in finding the relations, like *transitive* and *asymmetric* between the words, thus resolving the ambiguity between the words. A *Semantic Network* is one such representation technique for word relations [86].

## 3.2 Word Senses and Networks

The relations among the objects in a semantic network are specified with the help of operators: *subset*, *member* and *properties*. In a typical case, following relations may exist between different objects.

$$Mammals \subset Animals \qquad Mammals\ Has\_4\_Legs$$
$$Birds \subset Animals \qquad Birds\ Have\_Property\ Fly$$
$$Cat\ \subset Mammals \qquad Cheetah \in Cat$$
$$Bat \subset Mammals \qquad Pat \in Bat$$
$$Penguins \subset Birds \qquad Opus \in Penguin$$

The above relations can be represented in the form of a semantic *network*, shown in figure 3.1. The network form of representations is helpful to deduce the relationship from one object to another. For example, though it is explicitly not specified in the Knowledge-Base (KB) used for construction of semantic network in figure 3.1, still it can be deduced that "a cat has 4 legs". In such a network, if there is any path leading from one object $w_i$ to $w_j$ via any number of object nodes, then $w_i$ and $w_j$ are related, otherwise not. The nodes in a continuous link form a common context, thus helping in resolving the ambiguity.

Semantic Networks allow inheritance. Due to this, information can be stored once only at the highest level and all the objects at the lower levels inherit the properties of higher-level objects. This reduces the redundancy as well as the size of total KB required to build a semantic network.

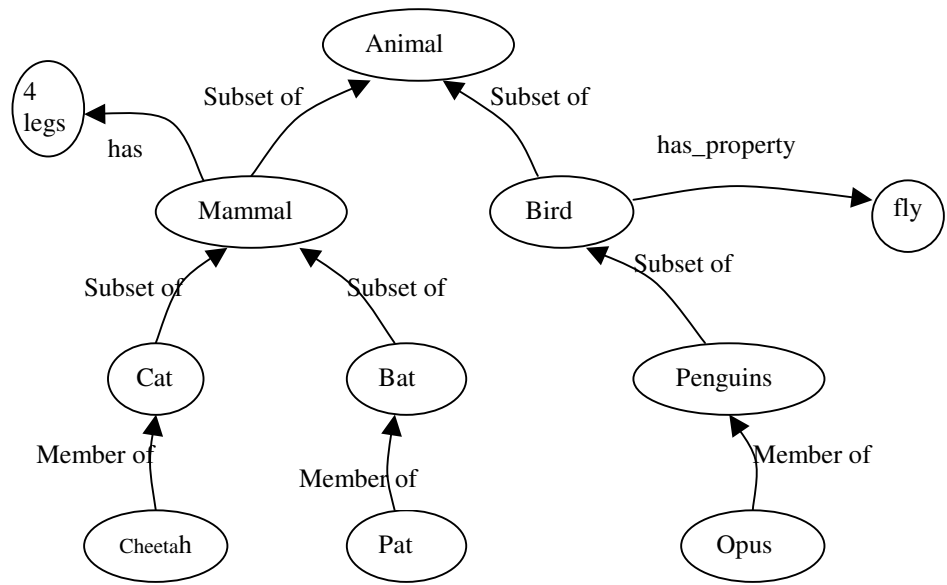

Figure 3.1: A Semantic Network.

Due to the relational and inheritance properties of semantic networks, the lexical and semantics knowledge of words in a dictionary can be represented using these properties. The word forms represent nodes in the network, and various relations between them, like – Synonimity, category, sub-category, and components of an object, can be represented through direct links between objects using pointers.

## 3.3 WordNet

WordNet is a manually constructed online lexical database in the form of public domain dictionary, where lexical objects have been semantically organized in the form of a large network [54]. The lexical units are synonym sets, called *synsets*. These are related by *symmetric relation*. There are four databases, separate for nouns, verbs, adjectives and adverbs. Each of these databases consists of a set of lexical entries corresponding to unique orthographic forms, accompanied by set of senses associated with each form, in the form of synsets. The database is accessible manually through the built-in browser as well as through a set of programs. Figure 3.2 shows the senses for the word entry *board*.

The noun board has 9 senses (first nine from tagged texts)
1.board - - (a committee having supervisory powers; "the board has seven members")
2. board- -(the flat piece of material designed for a special purpose; "he nailed the boards across the windows")
3. board, plank - - (a stout length of sawn timber; made in a wide variety of sizes and used for many purposes)
4. display panel, display board, board - - (a board on which information can be displayed to public view)
5. board, gameboard - - (a flat portable surface (usually rectangular) designed for board games; "he got out of the board and set up the pieces")
6. board, table - - (food or meals in general; "she sets a fine table"; "room and board")
7. control panel, instrument panel, control board, board, panel - - (an insulated panel containing switches and dials and meters for controlling electrical devices; "he checked the instrument panel"; "suddenly the board lit up like a Christmas tree")
8. circuit board, circuit card, board, card - - (a printed circuit that can be inserted into expansion slots in a computer to increase the computer's capabilities)
9. dining table, board - - (a table at which meals are served; "he helped her clean the dining table"; "a feast was spread upon the board")

The verb board has 4 senses (first 2 from tagged texts)

1.board, get on - - (get on board of (trains, buses, aircraft, ships, etc.))
….

Figure 3.2: Part of the entry for *board* in WordNet.

If a word has more than one sense, it will appear in more than one synsets. For example, the word *board* has nine senses, four of these are: {*board, committee*}, {*board, plank*}, {*board, control panel, instrument panel, panel*}, {*board, circuit board, circuit card*}. To disambiguate a word, its valid sense in the current context needs to be determined.

Each concept subsumes other concepts, which are more specific than earlier, called *hyponyms*. Thus, the synsets are organized in a hierarchy via super-class/sub-class relationships in the form of hypernym/*hyponym*. Figure 3.3 and 3.4 show the use of hypernym and hyponym, respectively, for the word form *board*. The hypernyms are generic words, where as the hyponyms are specific or subcategories.

```
9 senses of board

sense 1
board - - (a committee having supervisory powers; "the board has seven members")
 => committee, commission - -  (a special group delegated to consider some matter)
  => administrative  unit - -  (a unit with administrative responsibility)
   =>  unit, social unit - - (an organization regarded as part of a larger social group; "the
coach said the offensive unit did a good job"; "after the battle the soldier had trouble re-
joining his unit")
     =>organization, organisation - - (a group of people who work together)
       =>  social group - - (people sharing some social relation)
         => group, grouping - - (any number of entities (members) considered as a unit)

sense 2
board - - (a flat piece of material designated for a special purpose; "he nailed boards
across the windows")

…..
```

Figure 3.3: Hypernym (*board is a kind of …*) from WordNet.

The Hyponymy is a *transitive* and *asymmetrical* relation, where a single superordinate generates a hierarchical structure in which a hyponym is said to be below its superordinate. This forms a chain of relations in which a hyponym inherits all the features of the more generic concept and adds at least one feature that distinguishes it from its superordinates. For example, the concept *Cat* has a hypernym *mammal*, and one of its hyponym is *Cheetah*. Thus, a lexical tree can be constructed by following trails of superordinate terms, like: *cheetah @→ cat @→ mammals*.  Here '@→' is *transitive* and *asymmetric* semantic relation that can be read '*is-a*' or '*a kind of* (*ako*)'. By convention '@→ ' is said to point upward. This design creates a sequence of levels or hierarchy, going from many specific terms at the lower level to a few generic terms at the top. Hierarchies also provide a conceptual skeleton for nouns.  Whenever it is the case that a noun *u* @→ noun *v,* there is always an inverse relation, *v~→ u.* The inverse semantic relation goes from generic to specific, so it is a specialization. Thus, due to the existence of subordinate/superordinate relation the WordNet can be searched upwards as well as downward with equal ease.

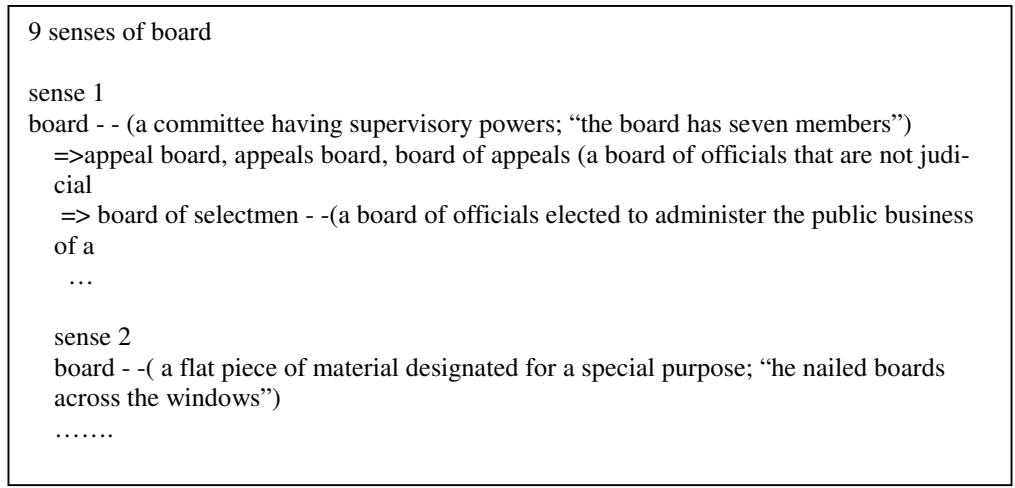

Figure 3.4: Hyponym (…*is a kind of board*) from WordNet.

## 3.4 Word Sense Disambiguation Model

The disambiguation principle used in this thesis is based on the simple logic that the meaning of a sentence is the result of the combined effect of semantics of words used in that sentence. Thus, there is a kind of dependency relationship between the senses of words in a sentence. Further, the sense of each word is affected by the sense carried by other words in the sentence. For example, the words in the phrases – "sales tax", "class teacher", and "flood control scheme"; the word "tax" indicates that it is the one which is due to the "sales", the word "teacher" stands for the one who is teaching in the "class", and "control" stands for the one used for "flood". This dependency of meanings among the words in a phrase or sentence can be explored to eliminate the ambiguity in the meaning of words in that sentence. Due to this interdependency of semantics of words in a sentence, there should be some overlapping word(s) between one of the sense definitions of the word to be disambiguated and the sense definitions of rest of the context words in the sentence. For each sense of a word, the corresponding hyponyms as well as hypernyms are semantically related to that sense of the word. Thus, in addition to above there should be overlapping words between the hyponyms/hypernyms of valid sense of the word and the definitions of rest of the context words in a sentence.

To establish the valid sense of an ambiguous word *w* in a given sentence, it needs to find out the overlapping words between the two sets, $\mathcal{U}$ and $\mathcal{C}$. Where, set $\mathcal{C}$ is sense definitions of the context words in the sentence. Set $\mathcal{U}$ is variable, constructed as union of - sense definition, hyponyms and hypernyms of the word *w*, by taking its senses one by one in order. The value of variable set $\mathcal{U}$ having maximum overlap with the set $\mathcal{C}$ is correct sense of *w*. In case no overlap is found, the most frequently used sense of *w*, i.e., first sense in the list of senses for *w* in the WordNet database, is taken as a valid sense.

Let there be a query sentence *q*. It is required to disambiguate the word $w \in q$. Following are the terms used in the disambiguation algorithm:

> *w = the word to be disambiguated*
> *q′ = q - {w} are context words in the sentence q*
> $\mathcal{S} = \{s_1, s_2, …, s_m\}$ *are sense definitions of w,*
> *where each $s_i$ is a synset*
> $\mathcal{O}_i$ *is set of hyponyms (subordinates) for $s_i$*
> $\mathcal{E}_i$ *is set of hypernyms (superordinates) for $s_i$*
> $t_1, t_2, t_3, T_1 …T_m$ = *temporary storage*

The algorithm in Figure 3.5 finds the valid sense of an ambiguous word *w* in the user query *q*, and Figure 3.6 illustrates the process of word sense disambiguation for this word.

The problem of word sense ambiguity is relatively more serious for small size queries. In the case of larger queries, single ambiguous word play a small role in determining the sense of the query, because a large number of remaining context words in the query sentence determine the sense.

---

*Algorithm - Disambiguate: (input: query sentence q, w)*
1. *Parse the query q and find its noun, verb, adj, and adv*
2. *Eliminate stopwords in q*
3. $q' = q - \{w\}$
4. $\mathcal{C} = NUL,$ // *set of sense definitions of context words in q'*
5. *for each context word* $v \in q'$ *do*

    a. $\mathcal{C} = \mathcal{C} \cup$ *synset of v*

6. *for each* $s_i \in \mathcal{S}$ *do* //*sense definitions of w*

    a. $t_1 = |s_i \cap \mathcal{C}|$ //*context term overlapping with sense def.*

    b. $t_2 = |\mathcal{O}_i \cap \mathcal{C}|$ // ... *overlapping with hyponym*

    c. $t_3 = |\mathcal{E}_i \cap \mathcal{C}|$ // ... *overlapping with hypernym*

    d. $T_i = t_1 + t_2 + t_3$ //*total overlap*

7. *find the largest of* $T_1 ... T_m$, *let this be* $T_j$
8. *if* $T_j \# 0$

    *output – "$s_j$ is closest sense of w"*

    *else*

    *output sense No.1*

9. *end*

---

Figure 3.5: Word Sense Disambiguation Algorithm.

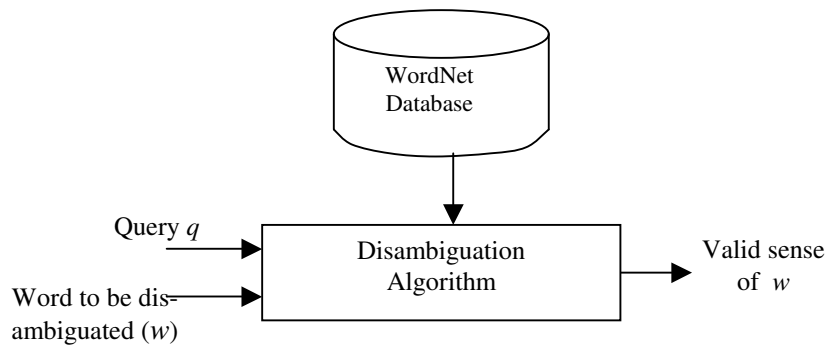

Figure 3.6: Word Sense Disambiguation Process.

## 3.5 Word Sense Disambiguation – Illustrative Example

Given two query phrases, it is required to disambiguate the word *v=board,* in the following queries, using WordNet and context of the query phrase.

1. *Selection (n) board (n)*

2. *Domestic (a) wiring (n) board (n).*

In the above '*n*' stands for noun form of the word, and '*a*' for adjective form of the word.

The different possible senses for *board*, taken from WordNet, are shown in figure 3.2. Following are the sense definitions from WordNet for the context words in the above query phrases.

*Board* Senses: There are nine senses for noun *board* (details in figure 3.2).

The Noun *selection* has five senses:

1. choice, selection, pick - -(the act of choosing, "your choice colours was unfortunate"; "you can take your pick")

2. selection - -(an assortment of things from which a choice can be made; 'the store carried a large selection of shoes")

3. choice, pick, selection - -(the person or thing chosen or selected; "he was my pick for Mayer")

4. survival, survival of the fittest, natural selection, selection - - (a natural process resulting in the evolution of organism best adapted to the environment)

5. excerpt, exact, selection - -(a passage selected from a larger work; "he presented excerpts from William James' philosophical writings")

The adjective *domestic* has 5 senses:

1. domestic - -(of concern to or concerning the internal affairs of a nation; "domestic issues such as tax rates and highway construction")

2. domestic - -(of or relating to the home; "domestic servant"; "domestic science")

3. domestic - - (of or involving the home or family' "domestic worries"; "domestic happi-ness"; "they share the domestic chores"; "everything sounded very peaceful and domes-tic"; "an author of blood-and-thunder novels yet quite domestic in his taste")

4. domestic, domesticated - -(converted or adapted to domestic use; "domestic animals"; "domesticated plants like maize")

5. domestic - -(produced in a particular country; "domestic wine"; "domestic oil")

The noun *wiring* has two senses:

1. wiring - -(a circuit of wires for the distribution of electricity)

2. wiring - - (the work of installing the wires for an electrical system or device)

## a. Disambiguating *board* in *"selection board"*:

Let us first consider the phrase 1, to disambiguate the sense of word - *board.* Here, $q$="selection, board". To find out the correct sense of *board*, all synonyms, hyponyms and hypernyms, in each sense of *board* are compared with the clubbed sense definitions of remaining context words, i.e., $q$ - {*board*} = {*selection*}, in the query sentence $q$. The comparison for this is shown in Table 3.1.

Table 3.1: Overlap between the clubbed definitions of selection with *Synset, Hyponyms* and *Hypernyms* of *board.*

| *board* Sense No. | Number Overlap words in the Synonym sets ($t_1$) | Number of Overlap words in the Hyponym set ($t_2$) | Number of Overlap words in the Hypernym set ($t_3$) | Total Overlaps counts ($t_1+t_2+t_3$) |
|---|---|---|---|---|
| 1. | 0 | 0 | "select", 14 times | 14 |
| 2. | 0 | 0 | 0 | 0 |
| 3. | 0 | 0 | 0 | 0 |
| 4. | 0 | 0 | 0 | 0 |
| 5. | 0 | 0 | 0 | 0 |
| 6. | 0 | 0 | "organism", 1 time | 01 |
| 7. | 0 | 0 | 0 | 0 |
| 8. | 0 | 0 | 0 | 0 |
| 9. | 0 | 0 | 0 | 0 |

Thus correct sense of board in the phrase "selection board" is*:*

1.board - - (a committee having supervisory powers; "the board has seven members").

**b. Disambiguating *board* in "domestic wiring board":**

Let us consider phrase 2 above where again the word board is to be disambiguated. The context phrase is, $q$ = "domestic wiring board". As in the previous case, the correct sense of *board* is to be found out, from its total nine senses to match with context in the query sentence $q$. When synonyms sets of each sense of *board* are compared, with the clubbed sense definitions of all the remaining context words in $q$, i.e., $q-${*board*} ={*domestic wiring*}, the overlapping words are as per Table 3.2.

**Table 3.2: Overlap between the clubbed definitions of domestic and**
***Wiring* with *synset, hyponyms* and *hypernyms* of *board*.**

| *board* Sense No. | Number of Overlap words in the Synonym sets ($t_1$) | Number of Overlap words in the Hyponym set ($t_2$) | Number of Overlap words in the Hypernym set ($t_3$) | Total Overlaps counts ($t_1+t_2+t_3$) |
|---|---|---|---|---|
| 1. | 0 | 0 | 0 | 0 |
| 2. | 0 | 0 | 0 | 0 |
| 3. | 0 | 0 | 0 | 0 |
| 4. | 0 | 0 | "device" 3 times | 3 |
| 5. | 0 | 0 | 0 | 0 |
| 6. | 0 | 0 | 0 | 0 |
| 7. | "electrical" 2 times, "device" 1 time | 0 | "electrical" 2 times, "device" 5 times | 10 |
| 8. | 0 | 0 | "electrical" 10 times, "device" 5 times | 15 |
| 9. | 0 | 0 | 0 | 0 |

Thus the correct sense of *board* in query phrase "domestic wiring board" is number eight, as it has maximum overlap of 15 times. The *synsets* for sense 8 of *board are* listed below from figure 3.2.

8. Circuit board, circuit card, board, card - - (a printed circuit that can be inserted into expansion slots in a computer to increase the computer's capabilities)

## 3.6 Discussion

The illustrative example shows that the semantic network based method using WordNet results in highly satisfactory disambiguation. There are two reasons that account for this result. First, to disambiguate a word its synsets, hyponyms and hypernyms have been considered for matching with the sentential context of the ambiguous word. Second, the WordNet synsets have embedded sense tagged sentences based on the word being disambiguated. Since these sentences are sampled from the tagged texts, thus suggesting the correct sense in the given context.

It may be remarked that semantic networks provide efficient navigation, and WordNet provides pointers for the purpose of navigation among the related words. This results in an efficient disambiguation. In addition, since all the variants of the word, as well as its relations, have been used, the disambiguation is bound to be correct and efficient.

The sentence or the query, in which a word is to be disambiguated, should be sufficiently large, so that there are enough contexts available to help in resolving the ambiguity. The resolution will suffer if context words are limited.

The model of disambiguation system presented here is based on assumption that, a document relevant to a query might contain either the words in the query or their synonyms. This implies that recall can be improved by considering the synonyms as part of the DR and IE queries. However, if all the possible synonyms of the words in the query are added as part of the query, then many irrelevant documents are also likely to be retrieved. Thus, to improve both the precision and recall, only the close synonyms should be used in the query. The model suggested here, makes use of word hierarchies through Semantics Networks, which finds the words related to the word being disambiguated, based on the relation of transitivity and asymmetry.

# Chapter 4

# Fuzzy Logic based Document Retrieval

---

## 4.1 Introduction

Fuzzy sets and possibility theory provide a homogeneous framework for the representation of both imprecise/uncertain information and soft queries with a flexible interpretation. A fuzzy set F is an extension of the idea of a binary set. Uncertainty refers to the lack of available information about the state of the world for determining if a classical statement (which can only be true or false) is actually true or false [58]. Examples of such statements are: "it will rain today", "Tweety flies"(knowing that tweety is a bird and birds usually fly), etc. In such situations, what best can be done is to estimate the tendency of the statement to be true (or to be false). Several frameworks may be possible for this: i) numeral approaches such as probability theory, possibility theory, belief functions and more ad hoc certainty factor-based techniques, ii) purely symbolic deduction methods using non-classical mechanisms for producing plausible conclusions in spite of the partial lack of information.

The imprecision refers to the contents of the considered statement and depends on the granularity of the language used to describe the information. For instance, the sentence "Rajan is between 25 and 30 years old" is clearly imprecise. However, the sentence "Rajan is 25 years old" is precise only if integer values for the age are considered, and imprecise if the values should indicate the number of months. Imprecise statements stem from disjunctive information such as "Rajan is 25 or 27", or negative information, when the underlying domain contains more than two values, such as "Rajan is not between 25 and 27 years old". An extreme situation is that Rajan's age is completely unknown, which means that any value of the universe may equally be assigned. Imprecision may be represented in terms of subsets of the relevant attribute domain, which are not singletons. These subsets contain the possible values, which can be assigned to the attribute for the considered object.

A vague statement contains vague or gradual predictions. It may also include quantifiers. For instance, "Rajan is a young researcher" refers to Rajan's age using linguistic term "young". There is no universally accepted meaning of the word "young". A person of 50 years age may be considered young in a group of very elderly persons in the age group of 60-80 years.

## 4.2 Fuzzy Sets, Fuzzy Logic and Fuzzy Relations

Fuzzy sets were introduced by Zadeh, in an attempt to propose a mathematical tool to describe the approach used by people when reasoning about complex systems [89], [92]. More particularly, Zadeh focused on the presence of classes without sharp boundaries in human-originated descriptions of systems. Fuzzy sets are meant to represent these classes. The idea is to turn the class membership into a gradual notion instead of the usual all-or-nothing view of classical logic. Thus, a fuzzy set F on a referential set A is simply described by a membership function $\mu_F$ that maps each element $a$ of A to the unit interval [0,1], where 0 stands for non-membership, 1 for complete membership, and numbers in between for partial membership. Unlike probability, the fuzzy logic deals with degree of appearance and not on frequency of occurrence. It is also called *possibilistic logic*, and the corresponding relations are called *possibilistic relations* [8].

At the mathematical level, the degree of possibility and certainty are closely related to fuzzy sets, and possibilistic logic is especially adapted to automatic reasoning when available information is pervaded with vagueness [91]. A vague piece of evidence can be viewed as defining an implicit ordering on the possible worlds it refers to, and this ordering can be encoded by means of fuzzy set membership functions.

Based on the notion of membership function, it is easy to extend many mathematical definitions pertaining the sets, to fuzzy sets. Set-theoretic operations for fuzzy sets are thus defined for two fuzzy sets F and G as given below [89].

Union F ∪ G: $\mu_{F \cup G} = max(\mu_F, \mu_G)$ ... (4.1)

Intersection F ∩ G: $\mu_{F \cap G} = min(\mu_F, \mu_G)$ ... (4.2)

Complementation $\overline{F}$ : $\mu_{\overline{F}} = 1 - \mu_F$ ... (4.3)

The results in (4.1) and (4.2) require a *lattice* structure for the membership scale. Thus, the fuzzy relations are *Partial* Ordering relations. Therefore, they are also called *similarity* or gradual *preference* relations [90].

## 4.3 Relations in Crisp and Fuzzy Sets

A relation between two items in the Boolean Logic can be expressed in the form of 0 and 1, which corresponds to no relation and 100% relation, respectively. Let the set of queries be Q and set of documents in the repository to which the queries are raised, be D. A classical binary relation from Q to D can be expressed as

R: Q × D → {0, 1} ... (4.4)

For every $q \in$ Q and document $d \in$ D, the relation R between them is expressed by R($q$, $d$). If there is an index term $t \in q$ and also $t \in d$, the relation is *reflexive relation* and R($q$, $d$) =1, otherwise R($q$, $d$) =0.

In the relations over fuzzy sets, the elements of two sets have a degree of association, which ranges from 0 to 1 [46], [77]. The fuzzy technique based on fuzzy set theory and fuzzy logic can be used for Document Retrieval (DR). In fuzzy retrieval technique, the matching between document index terms and query terms is graded based on the degree or level of matching, which ranges between 0 to 1. Thus, a fuzzy relation between query set and the document set can be expressed by

$$R: Q \times D \rightarrow [0, 1] \qquad\qquad \ldots \qquad (4.5)$$

and $R(q, d)$ may be any where between 0 and 1 depending on how closely the $d$ is associated with $q$. The relation $R(q, d)$ is a fuzzy reflexive *relation*.

## 4.4 Crisp-Set based DR

Consider a set $D = \{d_1, d_2, \ldots, d_n\}$ of potential text documents to be searched for the set of queries $Q = \{q_1, q_2, \ldots, q_m\}$. Let each query be represented by a set of keywords $t_1, t_2, \ldots, t_i$ and each document by a set of term $t_1', t_2', \ldots, t_j'$. Let us assume that $n=3$, and $d_1, d_2, d_3$ are the documents, shown in *Appendix*-1, to be searched for certain queries. The queries are the questions, given below:

$q_1$: *Which king had a liberal policy towards the religion?*

$q_2$: *Who was the queen of Jahangir?*

$q_3$: *What architectures were built by Shah Jahan?*

$q_4$: *Which mughal kings had interest for arts?*

The above questions have been represented by different sets of keywords, as follows:

$q_1$= {"king", "liberal policy", "religion"}

$q_2$= {"queen", "Jahangir"}

$q_3$= {"architecture", "Shah Jahan"}

$q_4$= {"king", "art"}

Documents in the set $D = \{d_1, d_2, d_3\}$, are also represented using the terms shown below:

$d_1$ = {"Akbar", "religion"}

$d_2$ = {"Jahangir", "art and justice", "trade"}

$d_3$ = {"Shah Jahan", "Mughal architecture", "art"}

It is now required to find out the binary relation R between every query v/s every document using the crisp set-based DR. Thus, for query $q_1$ it is required to find the relations R($q_1$, $d_1$), R($q_1$, $d_2$) and R($q_1$, $d_3$). The relation R($q_1$, $d_1$) can be expressed as:

R($q_1$, $d_1$) = R({"king", "liberal policy", "religion"},

{"Akbar", "religion"})

Since, the term "religion" ∈ $q_1$ and "religion" ∈ $d_1$, thus R($q_1$, $d_1$)=1. In the similar way, the matching terms between each query versus each document is found out and represented in matrix R.

$$
R \;=\; \begin{array}{c|c|c|c}
 & d_1 & d_2 & d_3 \\
\hline
q_1 & 1 & 0 & 0 \\
\hline
q_2 & 0 & 1 & 0 \\
\hline
q_3 & 0 & 0 & 1 \\
\hline
q_4 & 0 & 0 & 1 \\
\end{array}
\qquad \dots \qquad (4.6)
$$

Thus, matrix (4.6) shows that $d_1$ is relevant to $q_1$, $d_2$ to $q_2$, $d_3$ to $q_3$ and $q_4$. If two or more documents are found with some matching terms for a given query, then the relative relevance of a document is, number of terms matched in that with respect to the document having highest number of matches. Once relevance is computed, the DR system lists the documents in the order of their relevance, from the highest to lowest.

In the classical set theory and the corresponding logic there is either a 100% match between query term ($t_i$) and the index term ($t_j$') in the document, resulting to R($t_i$, $t_j$') = 1, or there is no match at all with R($t_i$, $t_j$') = 0. However, this does not serve the purpose due to following reasons: (i) The two terms from query and document may be formed from the same basic stem word, e.g., *real* and *reality*, *order* and *orderly*, *exact* and *exactness*, etc. In such cases, the crisp logic returns zero relevance. (ii) A term in the document, which is synonymous to the term in the query, the word-by-word match returns zero relevance. (iii) The two terms may be related by some other relation, for example, school and teacher, business and finance, car and driver, etc. In such cases the fuzzy based DR, described in the next section provides an acceptable solution.

## 4.5 Fuzzy Document Retrieval

The fuzzy set theory helps in dealing with uncertainties due to the incomplete match between the question terms and the keywords in the documents present in repository. The term Fuzzy Document Retrieval (FDR) refers to methods that are based on the fuzzy set theory. For a fuzzy set, the membership value $\mathcal{R}(t_i, t_j')$ specifies for each $t_i \in q$, $t_j' \in d$, the grade of relevance of keyword $t_i$ with the document $d$. The criteria for grade of relevance are: (i) $t_i$ and $t_j'$ terms are related if they are formed from the same basic stem word, (ii) $t_i$ and $t_j'$ are synonyms to each other, then the proximity of meaning of $t_i$ and $t_j'$ decide the grade of relevance. The membership value $\mathcal{R}(t_i, t_j')$ can range from 0 to 1.

These relations are specified in thesaurus, called *fuzzy thesaurus* [77]. The fuzzy thesaurus shows the relationship between pairs of words based on their degree of relevance. The structure of a fuzzy thesaurus can be specified as,

> *<WC1> <WC2> <RD>*

where *WC1, WC2* stands for word categories and *RD* is relationship degree between the words *WC1* and *WC2*. For example,

> *attraction, force, .7*
> *studious, bookish, .8*
> *theft, steal, .9*

An entry $<x_i>, <x_j>, <1.0>$ in fuzzy thesaurus shows that $x_i$ is a perfect synonym of $x_j$. There are different approaches for construction of fuzzy thesaurus. The fuzzy thesaurus can be manually constructed where experts in the domain of text can be asked to identify in a given set of index terms the pairs of words whose meaning they consider as associated, and provide the degree of association for each pair. Alternatively, it can be automatically generated from the lexicons [59]. For this, *transitivity relationship* can be applied for computing the missing relationship degrees, using the existing ones.

The thesaurus is a *reflexive fuzzy relation*, say $\mathcal{T}$, defined over $\mathcal{X}^2$, where $\mathcal{X}$ is Universe of index terms and query $q \subseteq \mathcal{X}$. For each pair of index terms $(x_i, x_j) \in \mathcal{X}^2$, the relation $\mathcal{T}(x_i, x_j)$ expresses the degree of association of $x_j$ with $x_i$, to which the meaning of the index term $x_j$ is compatible with the meaning of the index term $x_i$. The role of this relation is to deal with the problem of synonyms among the index terms. The relation helps to identify the relevant documents which otherwise would not be identified in the absence of perfect match between the keywords in the user query and the terms in the text document.

In FDR, a query is first expressed in the form of a fuzzy set $q$ based on the index terms $\mathcal{X}$. Then, by composing $q$ with the fuzzy thesaurus $\mathcal{T}$, a new fuzzy set $\mathcal{A}$ on $\mathcal{X}$ is obtained. This set $\mathcal{A}$ represents the augmented query, given by,

$$\mathcal{A} = q \circ \mathcal{T} \qquad \qquad \dots \qquad (4.7)$$

Here $\circ$ is called *max-min* composition operator [46]. The elements of $\mathcal{A}$ are specified by,

$$\mathcal{A}(x_j) = max\text{-}min [q(x_i), \mathcal{T}(x_i, x_j)] \qquad \qquad \dots \qquad (4.8)$$

where $x_i \in \mathcal{X}$, for all $x_j \in \mathcal{X}$. The set of retrieved documents represented by a set $\mathcal{F}$ and defined over D are then obtained by composing the augmented query expressed by the fuzzy set $\mathcal{A}$ with the relevance relation $\mathcal{R}$, expressed by,

$$\mathcal{F} = \mathcal{A} \circ \mathcal{R} \qquad \qquad \dots \qquad (4.9)$$

The strategy for FDR as outlined above is shown in the figure 4.1.

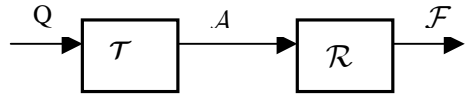

$\mathcal{T}$: Thesaurus, $\mathcal{R}$: Fuzzy Relevance Relation,

$q$: Query, $\mathcal{A}$: Augmented Query, $\mathcal{F}$: Retrieved
Documents in order of relevance.

Figure 4.1: Fuzzy Document Retrieval

The process of Information Extraction (IE) is carried out in two phases as shown in figure 4.2. In the first phase called *fetch*, documents relevant to the question under consideration are retrieved using FDR technique discussed above. The fetch phase is followed by *browse* phase, in which the information is extracted from the retrieved relevant documents and question is answered. The latter part is presented in chapter 5.

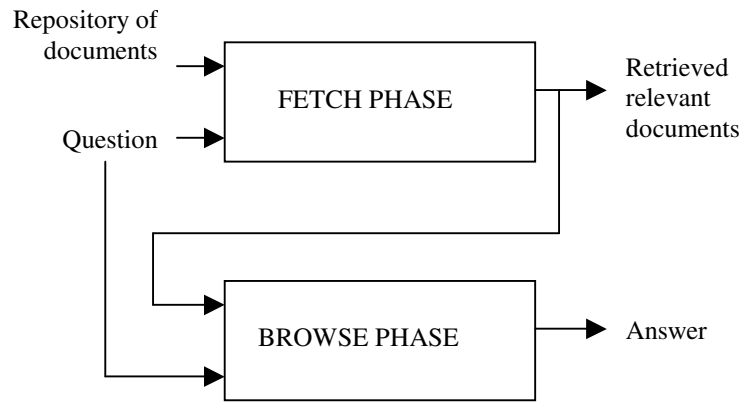

Figure 4.2: Information Extraction System.

## 4.6 Illustrative Example of FDR – Fetch Phase

The application of the technique discussed above is now illustrated. Consider the same set D = {$d_1$, $d_2$, $d_3$} of three documents. The keywords have been chosen based on the frequency of their occurrence in the documents. Also, consider the same set Q = {$q_1$, $q_2$, $q_3$, $q_4$} of four questions or queries.

These questions are first represented using keywords or index terms. In the case of question $q_1$ the keywords are: $t_1 =$ *king*, $t_2 = $ *liberal policy*, $t_3 = $ *religion*. Therefore,

$q_1 = \{t_1, t_2, t_3\}$

$\quad = [.9, \quad .6, \quad .7],$  (a vector representation of $q_1$)

In the above, the values .9, .6, .7 are called the *centralities* of $t_1, t_2, t_3$ respectively. The centralities indicate the presence of certain qualities, whose computations are modeled as computation of fuzzy membership degree. The relevant part of the fuzzy thesaurus $\mathcal{T}_1$, restricted to support $q_1$, is shown by the matrix:

$$\mathcal{T}_1 = \begin{matrix} & t_1 & t_2 & t_3 & t_4 & t_5 & t_6 \\ t_1 & \begin{bmatrix} 1 & 0 & .5 & .9 & .1 & .7 \\ 0 & 1 & .4 & .2 & .9 & .3 \\ .5 & .4 & 1 & .6 & .4 & .8 \end{bmatrix} \\ t_2 & \\ t_3 & \end{matrix}$$

where $t_4, t_5, t_6$ are the terms - *ruler, open, discipline* respectively. The composition of $q_1 \circ \mathcal{T}_1$ results in a fuzzy set $\mathcal{A}_1$, which represents the *augmented question*. Its vector form is given below.

$\mathcal{A}_1 = $ *max-min* $[q_1 \circ \mathcal{T}_1]$

$$= [.9 \quad .6 \quad .7] \circ \begin{bmatrix} 1 & 0 & .5 & .9 & .1 & .7 \\ 0 & 1 & .4 & .2 & .9 & .3 \\ .5 & .4 & 1 & .6 & .4 & .8 \end{bmatrix}$$

$$= [.9 \quad .6 \quad .7 \quad .9 \quad .6 \quad .7]$$

Assume that part of the fuzzy relevance relation which is restricted to support $\mathcal{A}_1$, is given by $\mathcal{R}_1$. Each of the index terms $t_1, t_2, t_3, t_4, t_5, t_6$ is related to documents $d_1, d_2, d_3$ by some degree of relevance, which can be expressed by following matrix.

$$
\mathcal{R}_1 = \begin{array}{c} \\ t1 \\ t2 \\ t3 \\ t4 \\ t5 \\ t6 \end{array} \overset{\begin{array}{ccc} d_1 & d_2 & d_3 \end{array}}{\begin{bmatrix} .1 & .1 & .1 \\ .5 & .2 & .1 \\ 1 & .1 & 0 \\ .3 & .4 & .2 \\ .6 & .3 & 0 \\ .8 & .3 & 0 \end{bmatrix}}
$$

Next, by composing $\mathcal{A}_1 \circ \mathcal{R}_1$, as per (4.9), results in a fuzzy set $\mathcal{F}_1$. The latter characterizes the relevance of the documents.

$$\mathcal{F}_1 = \textit{max-min} \,(\mathcal{A}_1 \circ \mathcal{R}_1)$$

$$= [.7 \quad .4 \quad .2]$$

The fuzzy retrieval matrix $\mathcal{F}_1$ shows that out of $d_1, d_2, d_3$ the document $d_1$ has highest relevance for the question $q_1$. Therefore, the answer to the question – "*Which king had liberal policy towards the religion?*", lies mostly in the document $d_1$. Therefore, the document $d_1$ should be fetched.

Similarly, the questions $q_2$, $q_3$ and $q_4$ are processed and the results are given as follows:

$q_2 = \{t_1, t_2\} = \{\textit{queen, Jahangir}\} = [.8 \;\; .9]$

$t_3 = \textit{empress}$, $t_4 = \textit{king}$, $t_5 = \textit{kingdom}$

$\mathcal{F}_2 = [.3 \;\; .9 \;\; .2]$

$\mathcal{F}_2$ shows that answer for $q_2$ lies mostly in the document $d_2$.

$q_3 = \{t_1, t_2\} = \{architecture, Shah\ Jahan\} = [.7\ \ .9]$

$t_3 = fort$, $t_4 = museum$

$t_5 = mosque$, $t_6 = temple$

$\mathcal{F}_3 = [.4\ \ .4\ \ .9]$.

$\mathcal{F}_3$ shows that the most relevant document for question $q_3$ is $d_3$ and therefore it has highest possibility of providing the answer.

$q_4 = \{t_1, t_2\} = \{king, art\} = [.8\ \ .7]$

$t_3 = painting$

$t_4 = beauty$

$\mathcal{F}_4 = [.1\ \ .7\ \ .7]$

The fuzzy retrieval matrix $\mathcal{F}_4$ shows that the relevant documents for $q_4$ are both $d_2$ and $d_3$. Therefore, the answer of question $q_4$ lies in documents $d_2$ and $d_3$. The possible solution is to find the answer from $d_2$ as well from $d_3$ and then merge them together.

For fuzzy retrieval of documents, the system lists the documents in the order of their relevance. The user may inspect all the documents supported by the fuzzy set $\mathcal{F}$ having non-zero relevance, or to inspect only some fraction of these documents. This fraction can be a number (say 5 to 10) as specified by the user in advance, or all the documents meeting the minimum threshold specified by the system [16].

Having located the most probable documents, which are expected to carry the answer for the question under consideration, the answer can be found out through IE from the fetched document(s) using the technique discussed in the next chapter.

## 4.7 Discussion

The use of fuzzy set theory for DR shows that the fuzzy relevance relation and fuzzy the-saurus are more expressive than their crisp set counter parts. In addition, since the degree of association is returned along with the retrieved documents, it helps the user to decide the order in which the documents can be viewed, particularly when the documents are in large number.

# Chapter 5

# Information Extraction using Structured Description Language

## 5.1 Introduction

Information Extraction (IE) from document repositories for answering specific questions presents formidable challenges. This is essentially due to (1) large amount of widely distributed information contents on web sites, (2) a gamut of domain specific as well as general information keeps appearing in electronic versions of news papers, magazines and journals and (3) the emergence of rapidly expanding knowledge in the form of *e-books* [21], [57].

These challenges can be dealt with by developing an IE system in two phases. In the first phase called *fetch*, documents relevant to the question under consideration are retrieved using Document Retrieval (DR) technique based on the theory of fuzzy sets, as described in chapter 4. The fetch phase is followed by *browse* phase, in which the matching is performed between the parsed results of question v/s retrieved relevant document sentences. This parsing is carried out for a Structured Description Language (SDL), which provides slot like syntactic structures for expressing sentences of English language including questions [17]. Every question as well as text sentence can easily be mapped to this structure. Thus, answer of a question can be found out by matching the question slot structure with the relevant documents' appropriate sentence slot structure, which is likely to carry the answer. This method of IE is easy to implement and provides fast and easily intelligible answer for a question [6].

## 5.2 Browse Phase – Information Extraction

In one of his famous poems the author Rudyard Kipling [44] wrote –

> "I keep six honest serving-men
> They taught me all I knew;
> Their names are *What* and *Why* and *When*
> And *How* and *Where* and *Who*.
> I send them over land and sea…"

Clearly, these interrogations are seeking answers for question cardinals in the form of *who, what, when, where,* and *why*. If taken in the present context, their computer based processing can solve major problems of IE from Natural Language (NL) texts.

The following sections describe SDL, the process of matching of text sentences from the retrieved relevant documents versus questions, and the extraction of answers.

### 5.2.1 Sentence Representation using SDL

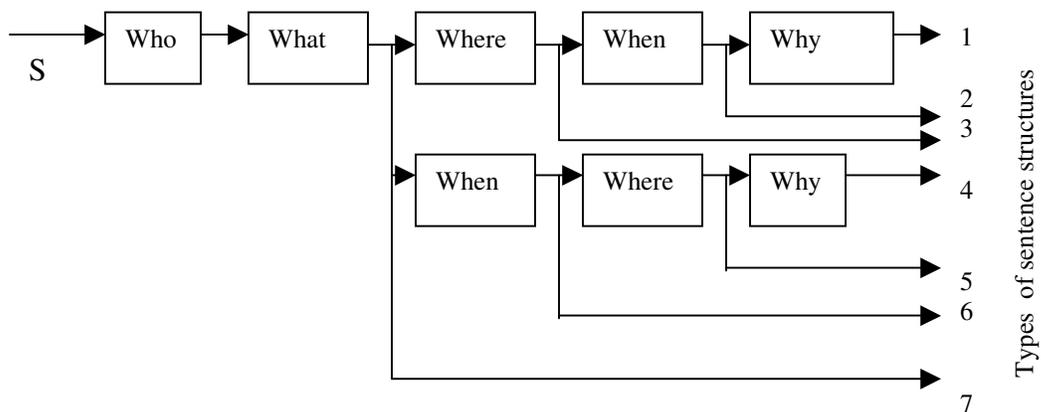

Figure 5.1: Transition Diagram of Sentence Structure in SDL.

A special language, called SDL has been developed in this section for representation of English language sentences. Using this, a text sentence S can be mapped into a transition graph or state diagram shown in figure 5.1. It shows that seven types of structurally different sentences labeled as 1, 2, …, 7, can be mapped into the SDL structure, based on the number of *wh-terms* (*who, what, when* etc.) and their positions in the sentences. Structure of each sentence type explicitly indicates the position of these interrogating wh-terms - *who, what, where, when* and *why*. The positions of these *wh-term*s are invariant.

Following examples demonstrate the mapping of English language sentences to SDL structure. It can be easily verified that the sentence numbers - a) to d) are of types 7, 7, 3, 4 respectively.

a) Akbar   | followed a liberal policy for the religion.
   *Who*    |    *What*

b) Jahangir | Married   Nur Jahan.
    *Who*    |   *What*

c) Red fort | is located    | at  Delhi.
    *Who*   |   *What*     |    *Where*

d) Shah Jahan | built  Taj Mahal  | during his rule | at Agra | in the memory of  queen
   Mumtaj.
   *Who*       |    *What*       |   *When*     | *Where* |  *Why*

It may be seen that the structures of the sentences in SDL are simpler in comparison to a language based on Context-Free Grammar (CFG) [43], [65]. SDL has following advantages over CFG: (1) a question in SDL may carry more than one sub-question, because a SDL sentence has more than one position for wh-terms. (2) The sub-answers for a question, which are distributed in a single or multiple documents, can be aggregated to fill up the slot like structures in the transition diagram to generate a single answer.

### 5.2.2 Question Answering System

The Question Answering system (QAS) shown in figure 5.2 implements the *browse* phase for IE. Once the relevant documents are retrieved for a given question through the *fetch* phase, the QAS extracts the answer from these documents [15]. The inputs to QAS are - the *question* and the most relevant *retrieved documents*. Every question is processed in two stages – SDL parsing and keywords expansion. The parser identifies the *wh-term* in the question. Next, the keywords in the question sentence are expanded using their synonyms. This improves the degree of matching between the question and the text sentence carrying the answer [3], [64].

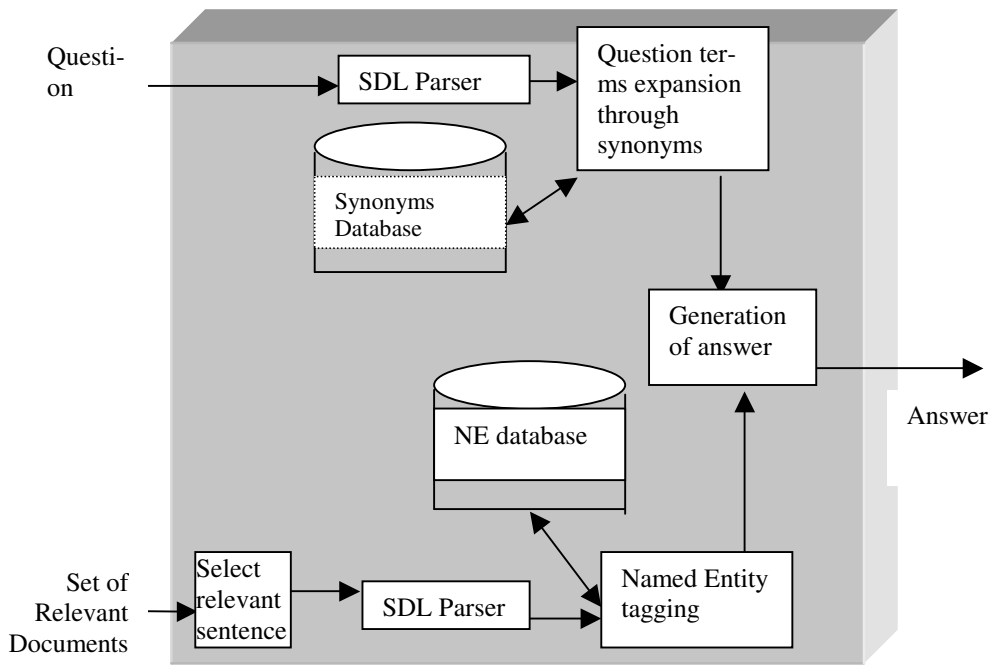

Figure 5.2: Question Answering System (QAS).

Input sentences from the relevant document are selected one by one based on their relevancy to the question. The relevancy of a text sentence depends on its degree of matching with the expanded question sentence. Each eligible sentence is then processed in three stages - SDL parsing, Named Entity (NE) tagging and generation of answer. The input sentence is parsed by the SDL parser to make it fit into one of the seven types of sentence structures, shown in figure 5.1. This helps in matching the slot values in text sentence versus the question sentence. The text sentence having highest degree of match with the question is selected as answer-carrying sentence. At the next stage, NE tagging from the selected answer sentence is carried out to find the answer for wh-term. Some examples of NEs are shown in Table 5.1.

**Table 5.1: Wh-terms and their answer types**

| Wh-terms | NE type as Answer |
|----------|-------------------|
| Who | PERSON |
| What | ACTION, STATUS |
| When | TIME |
| Where | PLACE |
| Why | REASON |
| Which | NAME |

## 5.2.3 Information Extraction for Answering Questions

Having decided the relevant document through fuzzy-set based DR in the *fetch* phase, it is required to determine the answer-carrying sentence in the relevant document. The following part describes how to locate the most probable sentences, which are likely to carry the information for answering the given question.

Consider the question – *"Which king had liberal policy towards the religion"?* Here the *wh-term* is '*which*'. To provide better matching between question and potential answer sentence, all the terms in the question sentence are supplemented by their possible synonyms [11]. This is indicated by the block 'question terms expansion through synonyms', in figure 5.2. Following is the expanded form of above question.

> *Wh-term*: {NAME}
>
> *Keywords*: {king | ruler | emperor, liberal | open, policy | rule | norms, religion}

While creating of the keywords list, the *stop-words – a, an, the, on, in, towards*, etc., are dropped from both the question and text sentence, and the matching is carried out. In the above question, the *wh-term* '*which*' is of the type NAME and stands for the name of somebody or some thing. The wh-term and keywords set are matched against the poten-tial answer sentence in the selected relevant document. The best matching answer sentence is selected based on the matching density of the keywords with the question sentence. Higher the matching density more suitable is the answer sentence.

The queries $q_1$, $q_2$, $q_3$ and $q_4$ considered in chapter 4 are reproduced below:

    $q_1$: *Which king had a liberal policy towards the religion?*

    $q_2$: *Who was the queen of Jahangir?*

    $q_3$: *What architectures were built by Shah Jahan?*

    $q_4$: *Which mughal kings had interest for arts?*

The relevant documents for these queries have already been retrieved in chapter 4. These are listed below:

| Query | Retrieved Document |
|---|---|
| $q_1$ | $d_1$ |
| $q_2$ | $d_2$ |
| $q_3$ | $d_3$ |
| $q_4$ | $d_1$ and $d_2$ |

Now, the process of information extraction during the *browse* phase is described for each query.

**$q_1$:** *Which king had a liberal policy towards the religion?*

The relevant document corresponding to this question is $d_1$. Next, through keyword matching, the potential answer-carrying sentence is selected as – "Akbar followed a liberal policy for religion". This sentence is selected based on the matching of keywords – *policy*, *liberal*, and *religion*. The phrase "which king" in the question has same implication as "who". The question and answer are reproduced below as per sentence type 7 of transition graph.

|  | *Who* | | *What* |
|---|---|---|---|
| $q_1$: | *Which king* | \| | *had a liberal policy towards the religion?* |
| Ans1: | Akbar | \| | followed a liberal policy for religion. |

Based on the above question and answer sentences, separate transition graphs for the question and answer sentence are constructed, and matched. This slot by slot matching confirms that the value of *who* slot in the question is "Akbar", which is also the answer for $q_1$.

$q_2$: *Who was the queen of Jahangir?*

The relevant document corresponding to this question is $d_2$. When query $q_2$ is parsed as per SDL and expanded by keywords, it appears as follows:

> *Wh-term*: {PERSON}
>
> *Keywords*: {queen \| wife of a king \| one married to a king \| female sovereign}

The potential answer-carrying sentence in this case is "Jahangir married Nur Jahan". This is because the keywords "Jahangir" and "married" matches in expanded query and the potential answer-carrying sentence. The answer for the *wh-term who* is person. In addition, in the answering carrying sentence *Nur Jahan* is NE, which corresponds to *who*. Finally, the sentence is re-organized and block-wise matched with the question, as shown below.

|  | *Who* | | *What* |
|---|---|---|---|
| $q_2$: | *Who* | \| | *was married to Jahangir?* |
| Ans: | Nur Jahan | \| | was married to Jahangir. |

This shows that the answer of the question $q_2$ is "Nur Jahan".

***q₃***: *What architectures were built by Shah Jahan?*

The relevant document for this question has been determined during fetch phase as *d₃*. After expansion by synonym terms the query *q₃* appears as follows:

> *Wh-term*: {ACTION, STATUS}
>
> *Keywords*: {built | construct | make, architecture | structure }

Based on the degree of keywords matching, the QAS finds answer carrying sentences in the form of sub-answers, distributed in the document. The question and corresponding answer carrying sentences are reproduced below. The subject, *Shah Jahan* in the question appears in place of object, therefore, as per the requirements of syntax of SDL in figure 5.1, the sentence is converted into active voice.

> *Who* | *What*
>
> *q₃* : *Shah Jahan* | *built What architectures ?*
>
> Ans₁: He | built Taj Mahal at Agra.
>
> Ans₂: He | also built the fort at Delhi, named as red fort, and a mosque named as Jama Masjid.

When question *q₃* is matched with Ans₁ and Ans₂ above, the answers are:

- Taj Mahal at Agra
- Fort at Delhi, named as red fort
- A mosque named as Jama Masjid

***q₄***: *Which king had interest for art*?

The question *q₄* after keywords expansion is shown below.

> *Wh-term*: {NAME}
>
> *Keywords*: {king | ruler | emperor, interest | curiosity | importance, art | creativity}

Here, the documents $d_1$ and $d_2$ are equally relevant, having a degree of relevance of 0.7 in each. The potential answer carrying sentences from these documents, selected by QAS, are given below, along with the question.

$q_4$:  *Which king  |  had interest for art*?

$d_2$: Ans$_1$:  Jahangir was highly learned man but he did not have the intellect of his father.

He  | was lover of art and justice.

$d_3$: Ans$_2$:  Under Shah Jahan,  |  the art of painting also developed significantly.

$d_3$: Ans$_3$:  Shah Jahan  |  gave liberal patronage to artists.

The wh-term *Which* here is similar to *Who*, and therefore the transition diagram part of *Who* is used here. In $d_2$:Ans$_1$, the word "Jahangir" is subject and the word "art" appears in the sentence, hence "Jahangir" is one answer. The subject "he" has been resolved to be "Jahangir" as it appears in the preceding sentence. In $d_3$:Ans$_2$, "Shah Jahan" is subject and the word "art" also appears in it, hence "Shah Jahan" is another answer. Similarly, $d_3$:Ans$_3$, results to an answer as "Shah Jahan". Merging the above three sub-answers, the complete answer is

*Jahangir,*

*Shah Jahan.*

In case, the highest relevance document does not contain the complete information required for answering the question, then the document of next lower relevance is taken for extracting the answer. The matching of keywords in the question with those in the potential answer sentence is *ranked*, and the highest ranking sentence is selected for answering. The ranking is determined by the following criteria: (1) the primary ranking is based on the percentage of question keywords found in the potential answer sentence. (2) Secondary ranking is based on the order - the keywords in answer appear in comparison to the order of those in the question. (3) Whether the keywords match exactly or their variants match in the question and answer.

## 5.3 Discussion

The chapter presents a new process of Question-Answering from a repository of NL texts. The process is carried out in two phases. In the first phase, called *fetch*, the relevant document is chosen based on theory of fuzzy sets; and the second phase, called *browse*, extracts the answer for the question from the retrieved relevant documents. It may be seen from the illustrative examples that the technique outlined in this chapter can be used for finding out the complete answer of a question from one or more relevant documents.

# Chapter 6

# Document Retrieval and Information Extraction using Bayesian Probabilistic Inference

## 6.1 Introduction

A Probabilistic Document Retrieval (DR) system ranks the documents in decreasing order of their probability of relevance to the user's *information need*. A probabilistic Information Extraction (IE) system locates the chunks of desired information based on their probability of relevance and browses them, from the documents already retrieved. The probabilistic DR systems are considered superior as compared to many other methods due to their theoretical soundness. One major difficulty in probabilistic DR model is finding a suitable method to estimate the probabilities of relevance of documents. The approach suggested here makes use of networks for representation of dependencies. The networks are based on Bayesian inferences, which are an extension of the basic theory of probability, and take into account the conditional dependencies present in the real world.

This chapter first presents the conceptual model for representation of documents and queries, followed by Bayesian inference networks' theory and its suitability for DR. Next, the application of Bayesian inferences for retrieval of documents and IE from the retrieved relevant documents is presented. The following essential pre-conditions have been considered while presenting the DR solution through Bayesian inference network:

1. The retrieval accuracy depends on the representation of queries and documents.
2. Representation of query and documents are plagued by a variety of uncertainties, viz. –imprecision, incompleteness, vagueness, and ambiguity.

## 6.2 Conceptual Model for Probabilistic DR

A conceptual probabilistic model (figure 6.1) has an event space, represented by $Q \times D$, where $Q = \{q_1, q_2, q_3, \ldots\}$ represents the set of queries, and $D = \{d1, d_2, d_3, \ldots\}$ is set of all the documents in the universe set [22]. The queries and documents are represented by *descriptors*, each of which is a set of terms or keywords. Each descriptor is a binary valued vector, in that each element corresponds to a term or keyword. A query is an expression of information need, which is regarded as a unique event, i.e., two same queries are treated as different events. The query is submitted to the system to retrieve the relevant documents as per the information need of the user.

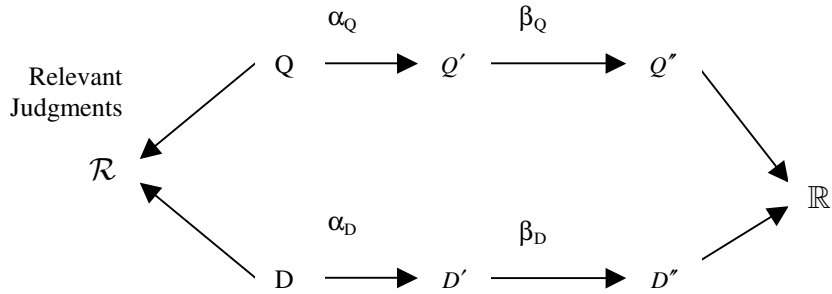

Figure 6.1: Conceptual Model for Probabilistic DR.

Let us consider $\mathcal{R}$ as the set of possible relevance judgments for the documents D and queries Q. In case of Boolean DR, $\mathcal{R} = \{ R, \bar{R} \}$ indicates that a document is either relevant or non-relevant. Thus, the relevance relationship between the query set and document set can be regarded as a mapping $r: Q \times D \to \mathcal{R}$. However, DR systems do not deal directly with documents and queries, but with their representations. For example, *index terms* are representation for a document and *Boolean expression* of terms for a query. Let $Q'$ and $D'$ be the representations of sets of queries and documents respectively, and $\alpha_Q$ be a mapping from $Q$ to $Q'$ and $\alpha_D$ be a mapping from $D$ to $D'$. Thus, two documents with the same set of index terms will be mapped onto the same representation.

To make the model more general, a further mapping is introduced from representation to *descriptions* of the objects. For example, queries and document representation in the form of index terms may be described by supplementing with weights of each term. Let these descriptions be $Q''$ and $D''$ for query sets and document sets, respectively, and corresponding mappings are $\beta_Q$ and $\beta_D$ respectively. Thus, the relevance relation between query and document sets should be based on their descriptions. The new value of the relevance function is therefore, represented by the expression r: $Q'' \times D'' \rightarrow \mathbb{R}$, which maps query-document pair to a *ranking value* or *relevancy value*, which is a real number. In response to query $q_j \in Q$, documents $d_k \in D$ are ranked according to descending order of $r(q_j'', d_k'')$. The function of a DR system that ranks the documents in the order of their relevancy for a query $q_j''$ is to calculate relevance and rank every document $d_k''$ in the collection of documents $D''$. However, for the sake of simplification, the description and representation are often treated as identical, both in the form of set of terms.

The probability that a document $d_k$ is relevant to the query $q_j$ can be expressed by $P(R|q_j, d_k)$ as per the conditional probability [79]. A precise definition of probability of relevance depends on the definition of relevance. The relevance is to some extent subjective, and depends on number of variables concerning - the document, the user and the information need of the user. A perfect retrieval is far from achievable, however, optimal retrieval can be defined for probabilistic DR. Because, it can be proved theoretically with respect to representations (or descriptions) of documents and information needs [66].

Let the queries and documents be described by the sets of respective index terms. Let T=$\{t_1, t_2, \ldots, t_n\}$ denotes the set of terms in the collection of documents. A query $q_j$ is a subset of terms belonging to T. Similarly, a document $d_k$ is a subset of terms belonging to T. For the purpose of retrieval, each document is described with the presence/absence of these index terms. Therefore, any document $d_k$ is represented with a binary vector:

$$\bar{x} \;=\; (x_1, \;\; x_2, \;\; ..., \;\; x_n) \qquad\qquad ... \qquad (6.1)$$

where $x_i = 1$ if $t_i \in d_k$, and for $t_i \notin d_k$, $x_i = 0$. The query $q_j$ is represented in the same manner. Now, the main task of a DR system based on relevance model is to evaluate the probability of a document being relevant. This is done by estimating the probability $P(R \mid q_j, d_k)$ for every document $d_k$ in the collection. Since relevancy for all the documents is evaluated for a given single query, the term $q_j$ can be dropped, and relevancy can be expressed by the Bayes theorem as follows [79]:

$$P(R \mid \bar{x}) \;\;=\;\; \frac{P(\bar{x} \mid R) . P(R)}{P(\bar{x})} \qquad\qquad ... \qquad (6.2)$$

where,

$P(R \mid \bar{x})$    is probability that the given document $\bar{x}$ is in the set of relevant documents

$P(\bar{x} \mid R)$    is probability of randomly selecting the document with description $\bar{x}$ from the set R of relevant documents,

$P(R)$    called *prior* probability of relevance, is probability that a document randomly selected from the entire collection is relevant,

$P(\bar{x})$    is probability that the selected document has description $\bar{x}$. It is determined as the joint probability distribution of the *n* terms with in the collection.

## 6.3 Bayesian Probabilistic Inference Model for DR

The basis for use of Bayesian probability for DR is Bayesian inference network [26], [35], [81]. The use of Bayesian inference networks for information retrieval represents an extension of probability-based retrieval. It is a Directed Acyclic Graph (DAG), where nodes represent propositional variables or constants and edges represents the dependence relationships between these propositions. If a proposition represented by a node *c* "causes" or implies the proposition represented by node *e* "effect", then it can be represented by a directed graph from *c* to *e*. The node *e* contains a *link* matrix that specifies P(*c*|*e*) for all possible values of two variables. When a node has multiple parents (for query node), the link matrix specifies the dependence of that node on the set of parents and characterizes the dependence relationship between that node and all nodes representing potential causes. Given a set of prior probabilities for the roots of the DAG (i.e., documents), these networks can be used to compute the probability of belief associated with all the remaining nodes. Figure 6.2 shows a typical situation in which a document $d_i$, corresponding keywords $t_1, \ldots, t_n$, and queries $q_1$ and $q_2$ are shown.

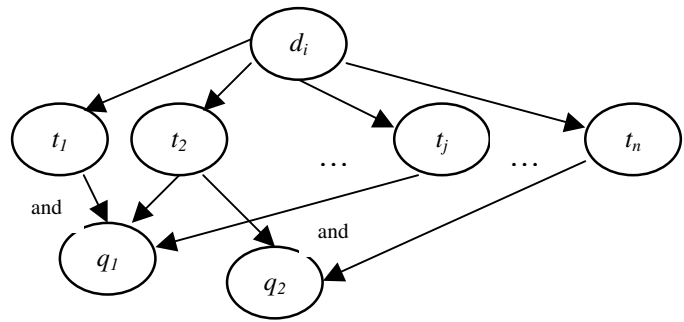

Figure 6.2 Basic Inference Network Model.

The inference network associates the random variables with the documents, index terms, and user queries. Multiple evidences of query terms in the document's representation for a given query are combined to estimate the probability that a document satisfies the user's information need. A document's variable associated with the document $d_i$ represents the event of observing the document. The index terms and documents variables are represented as nodes in a DAG. Edges are directed from document nodes to the index term nodes showing that observation of document yields the improved belief on its term nodes. The random variable associated with the user query, also shown by node, models the event that the information request specified by the query has been met. The dependence through the direction of arrows in DAG shows that the belief in the query node is function of the beliefs in the nodes associated with the query terms. In a particular case shown in figure 6.2, the document $d_i$ has $t_1, t_2, \ldots, t_j, \ldots, t_n$ as its index terms, indicated by the direction of arrows from $d_i$ to respective index term nodes. The query $q_1$ is shown to be composed of query-term $t_1$, $t_2$ and $t_j$, and $q_2$ with $t_2$ and $t_n$. Hence, $q_1 = t_1 \wedge t_2 \wedge t_j$, and $q_2 = t_2 \wedge t_n$.

A set of arcs leading to a node represents a probabilistic dependence between the node and its parents. Thus, a Bayesian network represents, through its structure the conditional dependence relations among the variables in the network. These dependence relations provide the framework for retrieving the probabilistic information.

For retrieving document, a user specifies one or more topics of interest as evidences identifying some document features. The DR task using Bayesian inference network can be specified in the form of an algorithm, shown in figure 6.3. The task requires building of an inference network for representation of query terms and document features (i.e. terms), and computation of posterior probabilities based on the prior probabilities of the documents.

---

*Algorithm 6.1: Bayes_inference*

1. *Build the network representing the query*
2. *Score each document as follows:-*
    (a) *Extract the features from the document*
    (b) *Label the features in the network*
    (c) *Calculate the posterior probabilities of relevance*
3. *Rank the document according the posterior probabilities.*
4. *end*

---

Figure 6.3: Algorithm for Bayesian Inference based DR.

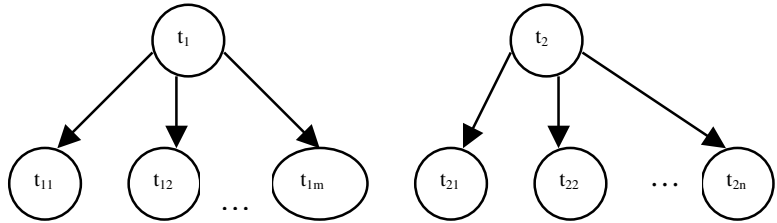

Figure 6.4 Two-Level Bayesian Network Model for DR.

The *term weighting* criteria discussed in 2.5.2 has been used for feature identification of documents in the Bayesian networks as shown in figure 6.4. The topic of interest (i.e., a query) is shown by terms $t_1$, $t_2$ (for example, $q$ = "house loan", where $t_1$=house, $t_2$=loan). There can be one or more document features to examine. A node $t_i$ represents the event "the document is related to topic $t_i$". The nodes $t_{11}$, ..., $t_{1m}$ are document features to be examined for the topic $t_1$, and $t_{22}$, ..., $t_{2n}$ are document features to be examined for topic $t_2$. Thus, nodes $t_{ij}$ represents the event that "feature $t_{ij}$ is present in the document". Here, assumption is made that $t_1$, $t_2$, ... have no dependence on each other (shown by absence of arcs between them). Similarly, $t_{11}$, $t_{12}$, ... and $t_{21}$, $t_{22}$, ... are also assumed independent of each other.

The network model shown in figure 6.4 requires two sets of probabilities to be fully computed:

1. *Prior probability* P($d_i$) that the document $d_i$ is relevant to the query topic, and

2. The conditional probability P($t_{ik} \mid d_i$) for each feature $t_{ik}$ for a given each topic $t_i$ in query, which shows that – "what is probability that feature $t_{ik}$ is present in a document, given that the document $d_i$ is relevant to query topic" ?

The task of DR system is to compute the *posterior probability* P($d_i \mid t_{i1}, t_{i2}, ..., t_{im}$), which means that – "what is probability that document $d_i$ is relevant, given that we have observed the presence or absence of all the features $t_{ij}$ for each document $d_i$. For the above inference network, the Bayes theorem can be directly applied to obtain the posterior probability, as follows:

$$P(d_i \mid t_{i1},...,t_{im}) \;=\; \frac{P(d_i).P(t_{i1},...,t_{im} \mid d_i)}{P(t_{i1},...,t_{im})} \qquad\qquad \ldots \qquad (6.3)$$

where $i = 1, \ldots,$ N are set documents in the repository.

In the experimental work presented in the following pages, the topic $t_i$ has been re-referred to as query term for a given query, and document features $t_{ij}$ have been referred to as synonyms / related words to the query term $t_i$.

## 6.4 Document Retrieval

Given:

(a) A set of queries Q=$\{q_1, q_2, \ldots, q_m\}$, where $q_1, q_2, \ldots, q_m$ are individual queries, and,

(b) A set of documents D=$\{d_1, d_2, \ldots, d_N\}$, where N is the total number of documents in the repository, with $n_i$ is size of each document, in words.

It is required to find the documents $d_i$, $i$=1,2, $\ldots,$ N, to which the query, say $q$, is related in the maximum relevance sense.

### 6.4.1 Approach

Using Bayes theorem, the probability of the overlap of keywords between the query terms (set $q$) and document terms (set $d_i$) is expressed by [79]:

$$P(d_i \cap q) \;=\; P(d_i \mid q)P(q) \;=\; P(q \mid d_i)P(d_i) \qquad\qquad \ldots \qquad (6.4)$$

where

$P(d_i \mid q)$      is probability that document $d_i$ is observed, given that query is $q$,

        (called, *posterior* probability)

$P(q)$      is probability of occurrence of query $q$,

$P(q \mid d_i)$      is probability that terms of query $q$ are observed, given that document is $d_i$,

$P(d_i)$ is probability of occurrence of the document $d_i$, called *prior* probability.

Thus, the probability that document $d_i$ is observed, given that query is $q$, can also be expressed as,

$$P(d_i \mid q) \;=\; \frac{P(q \mid d_i)P(d_i)}{P(q)} \qquad\qquad \ldots \qquad (6.5)$$

Since, P($q$) is common for the evaluation of expression for P($d_i \mid q$) for every document $d_i$, hence dropping P($q$) will not affect the ranking order of the documents $d_i$. The new value for P($d_i \mid q$) will henceforth be referred as RF($d_i \mid q$), where RF stands for *Relevance function* for ranking of documents $d_i$ for query $q$. Thus, above-mentioned expression becomes:

$$RF(d_i \mid q) \;=\; P(q \mid d_i)P(d_i) \qquad\qquad \ldots \qquad (6.6)$$

Let us first consider that there is only one term $t_1$ in $q$, and P($t_1 \mid d_i$) is probability of $t_1$ in $d_i$ given that $d_i$ has been located, and P($d_i$) is probability of occurrence of document $d_i$ in the lot. That is,

$$RF(d_i \mid t_1) \;=\; P(t_1 \mid d_i)P(d_i) \qquad\qquad \ldots \qquad (6.7)$$

Similarly, it can be computed for each query term $t_j$. Now, let us consider that $q$ comprises keywords $t_1$, $t_2$, …, $t_m$. Thus,

$$RF(d_i \mid t_1, t_2, \ldots, t_m) \;=\; P(t_1 \mid d_i)P(t_2 \mid d_i)\ldots P(t_m \mid d_i)P(d_i)$$

or

$$RF(d_i \mid t_1, t_2, \ldots, t_m) \;=\; \prod_{j=1}^{m} \; P(t_j \mid d_i)P(d_i) \qquad\qquad \ldots \qquad (6.8)$$

For the sake of simplicity, it is assumed that all documents are equally likely. Thus, $P(d_1) = P(d_2) = \ldots = P(d_N)$. With this simplification, the term $P(d_i)$ can also be dropped from the expression in (6.8), being the common multiplier in all the document's expressions. Now, the relevance function can be computed for every document $d_i$ for a given query $q = t_1, t_2, \ldots, t_m$, as follows.

$$RF(d_i | t_1, t_2, \ldots, t_m) \;=\; \prod_{j=1}^{m} \; P(t_j | d_i), \text{ for } i = 1, \ldots, \text{N} \qquad \ldots \qquad (6.9)$$

Using the fuzzy membership concept, the equation (6.9) above is further modified by introducing a fuzzy membership function $\mu_j$ for each query term $q_j$. Thus,

$$RF_\mu(d_i | t_1, t_2, \ldots, t_m) \;=\; \prod_{j=1}^{m} \; \mu_j . P(t_j | d_i), \text{ for } i = 1, \ldots, \text{N} \qquad \ldots \qquad (6.10)$$

The query $q$, comprising $m$ number of terms can be expressed by $q = t_1 \wedge \ldots \wedge t_j \wedge \ldots \wedge t_m$, called *minterm* format. When all the $k$ number of synonyms and related words $\{t_{j_1}, t_{j_2}, \ldots, t_{j_k}\}$, for each query term $t_j$, are accounted in the document $d_i$, the weight of each term $t_j$ in the query is expressed by $\mu_j = \mu_{j_1} + \mu_{j_2} + \ldots + \mu_{j_m}$. Here, $\mu_{j_1}, \mu_{j_2}, \ldots, \mu_{j_m}$ are fuzzy membership degrees of $t_{j_1} \ldots t_{j_m}$, respectively, with respect to $t_j$. Considering this, the expression for maximum relevance function for document $d_i$ (with size $n_i$ words), for query $q$ is given by,

$$RF_\mu(d_i | q) \;=\; \prod_{j=1}^{m} \; \left( \frac{\mu_{ij_1} + \mu_{ij_2} + \ldots + \mu_{ij_k}}{n_i} \right)$$

$$=\; \prod_{j=1}^{m} \; \left( \frac{\sum_{r=1}^{k} \mu_{ij_r}}{n_i} \right), \quad for \quad i = 1, \ldots, N \qquad documents$$

Finally, the relevance function is expressed as follows:

$$RF_\mu(d_i \mid t_1, t_2, ...., t_m) \;=\; \prod_{j=1}^{m} \left( \frac{\sum_{r=1}^{k} \mu_{ij_r}}{n_i} \right), \quad for \quad i = 1, ..., N \qquad ... \ (6.11)$$

In the above, $\mu_{ij_1}$ to $\mu_{ij_k}$ are the fuzzy membership values of k *number of appearances of each query terms $t_j$, including its synonyms and related terms*, in the document $d_i$. The algorithm for computation of $RF_\mu$ is given in figure 6.5.

---

*Algorithm  6.2 : Relevance_Function*

      *1.  for each  document $d_i$, i = 1, ..., N,  repeat the following*
          *a.  initialize  relevance of $d_i$,  $wtRF_i$ = 1*
          *b.  For each query term $t_j$, j =1, …, m*
              *(i) initialize relevance weight of term $t_j$, $wt_{ij}$=0*
              *(ii) for each  $t_{jk}$ appearance of synonyms, and related  terms of $t_{jk}$, in the document $d_i$, repeat the following*
                  *$wt_{ij} = wt_{ij}$ + fuzzy weight of  $t_{ij_k}$*

              *(iii) $wtRF_i = wtRF_i \wedge wt_{ij}$*
      *2.  print $d_i$, $wtRF_i / (n_i)^m$*
      *3.  end*

---

Figure 6.5: Algorithm for Relevance function computation for a document.

## 6.4.2 Illustrative Example

A query comprises keywords expressing the user's information need. The number of keywords in a query may vary depending on how precise it is. More the number of key-words in a query, the more precise it is. However, the common users' psychology is to provide only few keywords per query. As per the *Google* search inquiry, there are on the average 2.2 keywords per query [27].

Let following be the queries to the system to retrieve the relevant documents from the documents repository:

$q_1$: Which documents  describe *housing loans?*

*q₂*: Which documents narrate *education innovations?*

*q₃*: Which documents discuss *home budgets?*

*q₄*: Which documents explain *career prospects?*

*q₅*: Which documents describe *tax reforms?*

Since, only the keywords in above queries express the information need, the queries $q_1$ to $q_5$ have been represented using keywords only.

*q₁ = housing ∧ loans*

*q₂ = education ∧ innovations*

*q₃ = home ∧ budgets*

*q₄ = career ∧ prospects*

*q₅ = tax ∧ reforms*

Each query in the above is of two keywords. Before retrieval is performed for a query, the keywords in the query have been *expanded* by supplementing them with additional words, which are either synonyms of the keywords or their related words [11]. Table 6.1 shows the queries, synonyms and related words, along with their fuzzy membership weights (i.e., closeness to the query term) shown in the parenthesis.

$P_\mu(d_i \mid q_1)$ is the probability that document $d_i$ is relevant, given that there is evidence of query terms $t_1$ and $t_2$ in document $d_i$. Therefore, $P_\mu(d_i \mid q_1)$ shall be zero, if any one of $t_1$ or $t_2$ is absent in $d_i$, due to the conjunction operation between $t_1$ and $t_2$. The term $t_1$ has been expanded to $t_{11}$,..., $t_{1k}$. Here, $t_1 = t_{11}, t_{12}, t_{13}, t_{14}, t_{15}$ = *house*, *home*, *building*, *residence*, *dwelling*. Similarly, $t_2 = t_{21}, \ldots, t_{27}$ = *loan, finance, financing, mortgage, borrow, advance, credit*. Considering the document as variable identified by *i*, above terms becomes $t_{ij_k}$ . A program (*rf.c* – for relevance function, *Appendix – 3*) computes the relevance function's value for each document, as per equation (6.11) and algorithm shown in figure 6.5.

To experiment and verify the theories presented above, a repository consisting of 12 numbers of documents, shown in *Appendix–2*, has been used. These documents have been accessed for their suitability to the information need expressed through five different queries ($q_1$ to $q_5$).

**Table 6.1 : Queries, terms, their synonyms and related words.**

| Query | Query terms | Synonyms with Fuzzy membership weights |
|---|---|---|
| $q_1$ | house(1) | home(.8), building(.7), residence(.3), dwelling(.2) |
| | Loan(1) | Finance(.8), financing(.8), mortgage(.7), borrow(.5), advance(.4), credit(.3) |
| $q_2$ | education(1) | educate(.9), educational(.9), university(.8), universities(.8), instruction(.8), academic(.8), educating(.7), student(.7), schools(.7), school(.7), college(.7), faculty(.6), teacher(.6), colleges(.6), teaching(.6), institute(.6), schooling(.6), courses(.6), course(.6), campus(.5), learning(.5), learn(.5) |
| | Innovation(1) | innovations(1), modern(.8), modernization(.8), Innovative(.8), innovative(.8), advance(.7), advances(.7), advancing(.7), improvements(.6), improvement (.6), projection(.6), projections(.6), skill(.5), skills(.5), aids(.5), audio(.4), video(.4) |
| $q_3$ | home(1) | domestic(.9), household(.8), family(.7), house(.7) |
| | budget(1) | budgeting(1), account(.8), funds(.7), fund(.7), finance(.6), finances(.6), spend(.5), expense(.5), expenses(.5), purchase(.5), purchases(.5), save(.4), saving(.4), expenditure(.4), buy(.4), spending(.4), buying(.4), bills(.3), bill(.3) |
| $q_4$ | career(1) | jobs(.9), job(.9), vocation(.8), employments(.8), employment(.8), profession(.7), services(.6), service(.6), occupation(.5) |
| | prospects(1) | opportunities(.9), opportunity(.9), avenue(.8), avenues(.8), forecast(.6), options(.6), oriented(.6), oriented(.6), orientation(.6), prediction(.5), projection(.5), industry(.5), industries(.5), technology(.5), engineering(.5), public(.4), private(.4) |
| $q_5$ | Tax(1) | taxes(.9), taxation(.9), incometax(.9), taxing(.8) |
| | reforms(1) | reform(1), relief(.9), reliefs(.9), reduction(.8), reductions (.8), lower(.7), lowering(.7), benefits(.7), benefit(.7), lessen(.6), deductions(.5) |

The relevance function computing program (*rf.c*) is invoked by name *rf* and accepts two command line arguments (figure 6.6). The first argument stands for query number (1-5) and second is text file name containing names of all text documents to be searched for their relevance to the query. The *rf* command returns the text document names along with relevance function's value, shown as *ranking weight*, in decreasing order of relevance function. Only those documents' names have been considered relevant and results have been displayed where relevance function's value is greater than a threshold (i.e., greater than 20% of maximum).

```
c:\> rf   2   docfiles.txt<cr>

        Document Name    Ranking weight
        ==========================
        d03.txt          0.175067 * 1E-04
        d09.txt          0.124068 * 1E-04
        d05.txt          0.090724 * 1E-04

        ==========================
```

Figure 6.6:  Finding relevant document for query $q_2$.

In the Similar way, the relevant documents are found out for remaining queries. Table 6.2 shows the complete results. These results indicate the ranking of documents based on their relevancy to the queries. For the query $q_2$ three documents have been returned, *d03.txt* being more relevant to the query than *d09.txt* and *d05.txt*. Similarly, for other queries, the results indicate the relevancy of documents to the various queries. It has been found that documents' contents show a strong similarity to the queries, in the order of the value of their relevance functions.

This method is more robust because it returns zero or insignificant value of relevance function for those documents, which are not relevant, and hence they can be ignored. In this category are those documents, where only one term or no term from the query has found a match.

**Table 6.2: Relevance functions results for queries.**

| Document Names | Highest function values for each query for different text documents | | | | |
|---|---|---|---|---|---|
| | LF value for $q_1$ (x10$^{-4}$) | LF value for $q_2$ (x10$^{-4}$) | LF value for $q_3$ (x10$^{-4}$) | LF value for $q_4$ (x10$^{-4}$) | LF value for $q_5$ (x10$^{-4}$) |
| *d01.txt* | 0.212766 | - | 0.060284 | - | - |
| *d02.txt* | - | - | 0.051367 | - | - |
| *d03.txt* | - | 0.175067 | - | - | - |
| *d04.txt* | - | - | - | - | - |
| *d05.txt* | - | 0.090724 | - | 0.081652 | - |
| *d06.txt* | - | - | 0.026242 | - | - |
| *d07.txt* | - | - | - | - | - |
| *d08.txt* | - | - | - | - | - |
| *d09.txt* | - | 0.124068 | - | 0.372205 | - |
| *d10.txt* | - | - | - | - | - |
| *d11.txt* | - | - | - | 0.164024 | - |
| *d12.txt* | - | - | - | - | 0.164905 |

**Table 6.3: Precision and Recall results.**

| Query | Relevant documents | Documents retrieved | Recall (%) | Precision (%) |
|---|---|---|---|---|
| $q_1$ | *d01* | *d01* | 100 | 100 |
| $q_2$ | *d03, d04, d05* | *d03, d05, d09* | 66.7 | 66.7 |
| $q_3$ | *d01, d02* | *d01, d02, d06* | 100 | 66.7 |
| $q_4$ | *d05, d09, d11* | *d05, d09, d11* | 100 | 100 |
| $q_5$ | *d12* | *d12* | 100 | 100 |
| Average performance | | | 93.3 | 86.6 |

Efficiency of the retrieval is measured using parameters - *recall* and *precision* (chapter 2, section 2.7). Table 6.3 gives the values of these parameters and their averages, for queries $q_1$ to $q_5$, for a repository of twelve documents. The deviation in precision and recall from 100 percent is because some documents' relevancy is a borderline case, due to which they may be considered as relevant or non-relevant when a threshold of 20% is considered. These results are close to ideal except in $q_2$ and $q_3$. In query $q_2$ (=*education* $\wedge$ *innovations*), a non-relevant document *d09.txt* has been retrieved and a relevant document *d04.txt* has missed the retrieval. The justification for this is as follows. The document *d09.txt* is about publishing, however, its contents appear quite like from education area, and hence it is retrieved. The reason why *d04.txt* is not selected by the system as relevant is that, it discusses about the forms and agencies of distance education instead of means for education innovations, which are the requirements in the query.

A large number of terms have been kept in the expanded queries to ensure that no relevant document is missed from retrieval. This has improved the average *recall* (fraction of relevant documents retrieved from the repository) to 93.3%. However, due to the larger size of expanded queries, some non-relevant (in fact less relevant) documents are also retrieved along with maximum number of relevant documents. This has lowered the average *precision* (relevant document fraction in the retrieved documents) to 86.6%. To achieve maximum value of precision as well as recall, there is a need for optimum size of original as well as expanded query.

## 6.5 Information Extraction

Once, the relevant document is retrieved (*fetched*) through the process of DR, it is required to label the relevant text segments in it, through the process of IE. It is a three step process - (1) *Extraction* of text segments, (2) evaluation of their *ranking* and (3) *visualization* of relevant information text segments in the order of their ranking. It is assumed that segments of these texts are non-overlapping contiguous strings in the form of sentences, each represented by $s = \{s_1, s_2, ..., s_n\}$, where $s_i$ are keywords in the current sentence. The relevance ranking of $s$ for query $q = \{t_1, t_2, ..., t_m\}$ is expressed by P(s| $t_1, t_2, ..., t_m$) and computed using the *relevance function* derived in equation (6.11).

Considering that query is $q = \{t_i\}$, the probability of relevance of sentence $s$ having $n$ number of terms is given by:

$$P(s \,|\, t_i) \quad = \quad \frac{(\mu_{i_1} + \mu_{i_2} + ... + \mu_{i_n})}{n} \qquad\qquad ... \quad (6.12)$$

In the above, $\mu_{i_j}$, $j=1,n$, is fuzzy relevance relation between query term $t_i$ and the $j^{th}$ term ($s_j$) in the sentence $s$. Considering $m$ number of terms in the query, equation (6.12) is modified as

$$P(s \mid t_1, t_2, ..., t_m) \quad = \quad \sum_{i=1}^{m} \quad \frac{(\mu_{i_1} + \mu_{i_2} + ... + \mu_{i_n})}{n}$$

$$= \quad \sum_{i=1}^{m} \quad \sum_{j=1}^{n} \quad \frac{\mu_{i_j}}{n} \qquad\qquad ... \qquad (6.13)$$

where,

$\mu_{i_j}$ is the closeness or fuzziness of the $i^{th}$ term ($t_i$) of the query with the $j^{th}$ term $s_j$ in current sentence $s$,

$m$ is the total number of query terms in the expanded query $q$, and

$n$ is the total number of terms in the current sentence.

The presence of query terms in some of the sentences in the retrieved document makes them eligible candidates for relevancy to the query. Higher the occurrence of query terms or their related terms in a sentence, the more strongly it indicates relevance to the information need of the user. For computation of relevancy at the level of sentence, disjunction of the query terms is performed unlike the conjunction at the document level.

Once computed for a given retrieved text document, the value of relevance function P($s$|$t_1$, $t_2$, ..., $t_m$) is stored in an array. Finally, extracted information in the form of sentences is displayed (*browsed*) in the descending order of relevance function, and those less than threshold (20% of the maximum value) are discarded. The longer are the sentences, lesser will be the truncation and rounding off errors, and therefore, result is likely to be more close to the factual.

The algorithm for information extraction is shown in figure 6.7, and the corresponding program (*ie.c* – Information Extraction, *Appendix–4*) computes the relevance ranking at sentence level for the text document already retrieved through DR.

The program *ie.c* is executed with following command format:

*C:\> ie   query_id    text_documentfile*

where *query_id* is query number (1 to 5) and *text_documentfile* is relevant text file,  retrieved through DR.

---

*Algorithm 6.3: IE (Information Extraction)*
1.  *initialize sentences array –tsents[ ]*
2.  *initialize expanded query array[term, relevance_weight_of term]*
3.  *initialize sentence rank array[sentence_id, rank]*
4.  *set text_wordcounter =0*
5.  *while not eof(text file) do*
    *(a)  read a character from text*
    *(b)  if word boundary over then*
        *store the word into sentences array, increase text_wordcounter*
6.  *for each sentence in sentence array do*
    *(a) set sentence relevance weight  swt=0*
    *(b) for each  term in query array*
        *(i)  search query term in current  text sentence*
        *(ii)  if match  found n times then*
            *swt = swt +  relevance_weight of term * n*
    *(c) update rank array for this sentence, i.e., [sentence_id, swt]*
7.  *sort sentence rank array in decreasing order of rank*
8.  *threshold=maximum  weight of sentence * 0.20*
9.  *for all the sentences in sentence array do*
    *(a)if sentence rank is > threshold*
        *print the sentence*
10. *end.*

---

Figure 6.7: Algorithm for Information Extraction.

The text documents, which have been found relevant during DR, are processed by IE algorithm (program *ie.c*). The results are presented section 6.5.1.

## 6.5.1 Computation of Results

**Example No. 1**

*C:\> ie     1   d01.txt*

EXPANDED QUERY TERMS

====================

house home building residence dwelling loan finance financing mortgage borrow advance credit
EXTRACTED RELEVANT TEXT FROM TEXT FILE d01.txt

=================================================

[Text Segment no. 14]  [rank*1000=  0.1636]

A cover for your home loan TIMES NEWS NETWORK [ THURSDAY NOVEMBER 28 2002 12:25:40 PM ] Mr Raman a senior executive at an MNC walked straight into an insurance office after buying a property

[Text Segment no. 25]  [rank*1000=  0.0900]

The good news is that housing finance companies and banks which earlier used to lend only 75-85 per cent of the project cost are willing to finance up to 90 per cent

[Text Segment no. 24]  [rank*1000=  0.0848]

And many are willing to customize the loan to specific needs

[Text Segment no. 13]  [rank*1000=  0.0818]

The borrower can pay a lower installment amount in the initial years and increase it later along with a rise in his income

[Text Segment no. 28]  [rank*1000=  0.0625]

The result: people have ended up borrowing much more than what they would have done a few years ago

[Text Segment no. 4]  [rank*1000=  0.0571]

In fact every bank and housing finance company says that its average home loan size has gone up in the last two years

[Text Segment no. 27]  [rank*1000=  0.0562]

The figure has risen from less than Rs five lakh two years ago to nearly Rs 8 lakh now

[Text Segment no. 6]  [rank*1000=  0.0533]

Companies are even willing to give you a home loan as high as Rs 1 crore if they feel you can repay it

[Text Segment no. 0]  [rank*1000=  0.0529]

That is probably why Raman decided to take up a home loan

[Text Segment no. 7]  [rank*1000=  0.0500]

After borrowing Rs 30 lakh he decided to take up an insurance policy

[Text Segment no. 2]  [rank*1000=  0.0486]

He didn't want to leave a huge liability for his family in the event of his death

[Text Segment no. 21]  [rank*1000=  0.0476]

His family would lose their dream home should they fail to clear the outstanding loan amount

[Text Segment no. 31]  [rank*1000=  0.0417]

A safer option was to take an insurance policy

**Example No. 2**

*C:\> ie      1   d05.txt*

EXPANDED QUERY TERMS
=====================
house home building residence dwelling loan finance financing mortgage borrow advance credit

EXTRACTED RELEVANT TEXT FROM TEXT FILE d05.txt
==============================================

 THERE IS NO RELEVANT TEXT IN THIS FILE !!!

The document *d05.txt* in example-2 has been included just to show that non-relevant document provides no results. However, IE process has input only those documents which were found relevant during DR in the fetch phase. The above information extraction shows that only those sentences have been extracted from the retrieved relevant text documents, which show relevance to the query.

## 6.6 Discussion

A conceptual model for DR has been presented, and used as a basic structure for realization of the probabilistic model. It has been demonstrated through experimental results that Bayes theorem can be used for representation of logical inferences in the form of Bayesian inference networks, which in turn can be used for Document Retrieval and Information Extraction from natural language texts [18]. The necessary algorithms for determination of relevancy of texts as well for IE have been developed. The experimental results strongly support the mathematical theory, derived for Bayesian inferences.

Following are the preconditions for Bayesian probability method:

1. The query terms are independent to each other.

2. Query size should be sufficiently large to ensure that no relevant document is missed from retrieval.

3. Accuracy of retrieval depends on the accuracy of representation of query and documents.

# Chapter 7

# Document Retrieval and Information Extraction using Dempster-Shafer Evidential Reasoning

---

## 7.1 Introduction

In most existing Document Retrieval (DR) systems, a document is considered as an atomic entity and retrieved as a single unit based on its index values, in response to the user query. However, documents constitute a structure in the form of different sub-units. In the case of a book as a document, chapters are sub-units, and paragraphs and sections are sub-units in case of a chapter as a document. If a document is indexed in the form of different independent units, then it is possible to retrieve the individual components of the document, resulting to Information Extraction (IE). Indexing individual text units is helpful to aggregate different texts when the required information exists in a collection of chapters instead of single chapter, or in a collection of paragraphs instead of single a paragraph.

   This chapter presents a logical structure for representation of documents so that a document can be treated as an atomic integral unit and at the same time can be realized in the form of a structure. In such a structure, the components are related hierarchically and can be treated as clusters of collection of sub-components. Next, the theory of uncertainty - *Dempster-Shafer* (D-S) *theory of evidence*, is introduced and applied to compute the relevance of documents' components for retrieval. The D-S theory of evidence appropriately suits this form of representation of documents; it is flexible in modeling the DR-IE process, and theoretically sound. Its main difference with classical probability theory is that (1) classical probability theory is treated as a special case of D-S theory, and (2) D-S theory allows the combination of evidence and representation of ignorance. The combina-tion property of D-S theory makes it attractive for modeling of DR-IE process. The combination of evidences is computed by Dempster's *combination rule*, which allows the expression of aggregation necessary in a model for representation and retrieval of documents, and extraction of information.

## 7.2 Logical Model for Representation of Documents

Documents are represented in the form of a tree structure, where nodes correspond to components of a document called *objects* and edges represent the relationship between the various objects (figure 7.1). The internal nodes in the tree are composite in nature, as they do not contain information of their own; the information content of these nodes is an aggregation of their components nodes. The document objects may be in the form of chapters or sections or paragraphs. All the child nodes under any node represent the subcomponents of the object represented by that node. If a node represents a chapter, then its child nodes represent the various sections under that, and the further lower level child nodes corresponds to the sub-sections or paragraphs of those sections. The leaf nodes, which correspond to the lowest level objects in the hierarchy, are not composite as they represent the raw text.

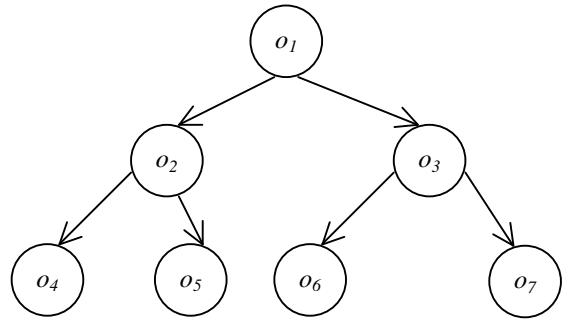

Figure 7.1: Structured Document Representation.

The structuring of documents is decided by the smallest unit of information, which needs to be extracted independently. Thus, this model turns out to be effectively information extraction system when smaller units, i.e., raw text objects or the composite objects are returned, and acts as true Document Retrieval system when whole document is returned. The root node is a special composite node, which represents the whole document.

The relation between the set of objects $O$ in a document structure is represented by a *binary relation* $\mathcal{R} \subseteq \mathrm{O} \times \mathrm{O}$. For any two objects $o_i, o_j \in \mathrm{O}$, $(o_i, o_j) \in \mathcal{R}$ means $o_j$ is direct component of $o_i$, and $\mathcal{R}(o_i)$ denotes collection of all the direct objects under $o_i$. Thus, in figure 7.1, $(o_2, o_4) \in \mathcal{R}$, $(o_3, o_6) \in \mathcal{R}$, and $\mathcal{R}(o_2) = \{o_4, o_5\}$. Objects $o_1, o_2, o_3$ are composite, $o_1$ is root, and the child nodes $o_4$ to $o_7$ represent raw information carrying objects.

Each object is represented by one or more index terms associated with their weights. The meaning associated with each object node $o_i$, called its *semantics*, is determined from its component objects using aggregation of semantics of these components. Thus, semantics of object $o_i$ is expressed as

$$\mathcal{S}(o_i) = \bigotimes_{o_j \in R(o_i)} \mathcal{S}(o_j) \qquad \qquad \dots \qquad (7.1)$$

where $\otimes$ is semantics aggregation operator, representing an operation of logical conjunction. Let $o_4, o_5$ be two objects, such that $\mathcal{S}(o_4) = $ *information retrieval* and $\mathcal{S}(o_5) = $ *information extraction*. The keywords sets *information retrieval* and *information extraction* represent the index terms of these objects. If $\mathcal{R}(o_2) = \{o_4, o_5\}$, then composite object $o_2$'s semantics should be based on the evidence of term *information retrieval* $\wedge$ *information extraction* = *information*, thus $o_2$ has to do some thing with the *information*.

## 7.3 Dempster-Shafer Theory of Evidence

Dempster-Shafer's (D-S) Theory of Evidence is a theory of uncertainty that was first introduced by *Dempster* and later on extended by *Shafer* [65], [74]. In applying D-S theory for DR-IE, each object is represented by a *body of evidence*, defined in a *frame of discernment* $\mathcal{T}$. The evidence is expressed as a set of terms that are used to index the object.

Each term $t \in \mathcal{T}$ corresponds to the proposition: "term $t$ indexes the object". A proposition $\mathcal{A} \subseteq \mathcal{T}$ expresses the fact that, "the set of terms in $\mathcal{A}$ indexes the object". Thus, the propositions are in one-to-one correspondence with the subsets of $\mathcal{T}$. A *basic probability assignment* (*bpa*) *m* represents the *uncertainty* of indexing term $t$, expressed as $m(t)$. It is the *belief* that $t$ is a good indexing term, i.e., $t$ describes the content of the object appropriately. Given a term $t$, the $m(t)$ for a document is determined from *Term Frequency*. Higher the $m(t)$ for a given object, $t$ is considered as a good indexing term for that. A term $t \in \mathcal{T}$ such that $m(t) = 0$, does not index the object, meaning that there is no evidence that the object is about $t$. The beliefs are computed using *bpa,* and represented by a *density function m*: $2^{\mathcal{T}} \rightarrow [0,1]$. The belief associated with a set of index terms $\mathcal{A} \subseteq \mathcal{T}$, can be expressed as

$$m(\mathcal{A}) = 0, \qquad \text{for } \mathcal{A} = \phi$$

and

$$\sum\nolimits_{A \subseteq T} m(\mathcal{A}) \; = \; 1 \qquad\qquad \dots \quad (7.2)$$

If there is positive evidence that $t$ has relevance to $\mathcal{A}$, then $m(\mathcal{A}) > 0$. In this case, $\mathcal{A}$ is called *focal element*, and the proposition $\mathcal{A}$ is said to be *discerned*. No belief can be assigned to a *false proposition* (represented by $\phi$). The sum all of *non-null bpa* must equate to 1. The set of focal elements constitutes a *core*, defined as

$$C = \{ \mathcal{A} \subseteq \mathcal{T} \mid m(\mathcal{A}) > 0 \} \qquad\qquad \dots \quad (7.3)$$

The core and its associated *bpa* define a *body of evidence,* using which it is possible to compute the total belief provided by the body of evidence for a proposition. This is done with a *belief function, Bel*: $2^T \rightarrow [0,1]$, defined upon *m* as

$$Bel(\mathcal{A}) = \sum_{B \subseteq A} m(B) \qquad \qquad \dots \qquad (7.4)$$

where $Bel(\mathcal{A})$ is the total belief committed to $\mathcal{A}$, that is, the total positive effect the body of evidence has on the relevance of *t* to $\mathcal{A}$. Different values of $Bel(\mathcal{A})$ can be interpreted as follows:

- *Bel($\mathcal{A}$) = 0:* no knowledge about proposition $\mathcal{A}$ is available, or the proposition $\mathcal{A}$ is false (*t* has no relevance to $\mathcal{A}$)

- *Bel($\mathcal{A}$)=1*: the proposition $\mathcal{A}$ is true (*t* has relevance to $\mathcal{A}$)

- *0 < Bel($\mathcal{A}$) < 1*: the evidence provides partial support that *t* has relevance to $\mathcal{A}$

A particular characteristic of the D-S framework (one which makes it different from probability theory) is that if *Bel($\mathcal{A}$) < 1* then the remaining evidence 1 - *Bel($\mathcal{A}$)* need not necessarily disprove $\mathcal{A}$ (i.e., supports its negation $\overline{A}$). In other words, in D-S there is no rule like $Bel(A) + Bel(\overline{A}) = 1$. Some of the remaining evidence may be assigned to propositions, which are not disjoint from $\mathcal{A}$, and thus could be plausibly transferable to $\mathcal{A}$ in light of new information. This is formally represented by a *plausibility function Pl:* $2^T \rightarrow [0,1]$, defined upon a *bpa m* as

$$Pl(A) = 1 - \sum_{A \cap B = 0} m(B) \qquad \qquad \dots \qquad (7.5)$$

where $Pl$ is the extent to which the available evidence fails to disprove $\mathcal{A}$. In *D-S*, the uncertainty of a proposition $\mathcal{A}$ is represented by an interval [*Bel($\mathcal{A$}), Pl($\mathcal{A}$)*]. The interval [1,1] represents the absolute certainty, where as [0,1] represents the absolute ignorance.

D-S theory has an operation, *Dempster's rule of combination*, for the pooling of evidences from the variety of sources. This rule *aggregates* two independent bodies of evidence, defined within the same frame of discernment, into one body of evidence. Let $E_1 = (C_1, m_1)$, and $E_2 = (C_2, m_2)$ be the two bodies of evidence defined in a frame of discernment $\mathcal{T}$, where $C_1$, $C_2$ are core of bodies of evidence, and $m_1$, $m_2$ are *bpa* respectively. The new body of evidence is defined by $E = (C, m)$, where core C is computed as follows:

$$
\begin{aligned}
C \;\; &= \;\; C_1 \;\; \otimes \;\; C_2 \\
&= \;\; \{c \subseteq T \,|\, \exists c_1 \in C_1, \;\; \exists c_2 \in C_2, \;\; c_1 \cap c_2 = c\} \qquad \qquad \ldots \qquad (7.6)
\end{aligned}
$$

For above, any non-empty intersection of any two sets in the cores $C_1$ and $C_2$ is part of the new core C, and reflects the case where two bodies of evidence $E_1$ and $E_2$ are agreed. Correspondingly, the new *bpa* is computed as [63],

$$
\begin{aligned}
m(A) \;\; &= \;\; m_1 \otimes m_2 \;\; (A) \\
&= \;\; \frac{\sum_{B \cap C = A} m_1(B) * m_2(C)}{1 \;\; - \;\; \sum_{B \cap C = 0} m_1(B) * m_2(C)} \qquad \qquad \ldots \qquad (7.7)
\end{aligned}
$$

Dempster's *combination rule* computes a measure of *agreement* between two bodies of evidence concerning various propositions discerned from a common frame of discernment. The rule focuses only on those propositions that are *supported* by both the bodies of evidence. The new *bpa* takes into account the *bpas* associated to the propositions in $C_1$ and $C_2$ that yield the propositions of C. The denominator of the above equation is a normalization factor that ensures that *m* is a *bpa* (i.e., it satisfies the equation 7.2).

The DR system based on D-S theory assigns the *bpa* values to focal elements based on the indexing terms and their frequencies in the documents, and the remainder of the belief is treated as *uncommitted belief*.

As per the Dempster's *dependency constraints*, a proposition representing the semantic content of a composite object semantically implies those of each of its direct components. Hence, as per the combination rule, for any proposition *r* discerned by *m*, for each $o_i \in \mathcal{R}(o)$ there exists proposition $t_i$ discerned by $m_i$, such that $t_1 \wedge t_2 \wedge \ldots \wedge t_n = r$. As per classical logic, $r \rightarrow t_i$, for $i = 1, n$. Therefore, (7.4) can be expressed as

$$Bel(r) \quad = \quad \sum_{r \rightarrow t} \quad m(t) \qquad \qquad \ldots \quad (7.8)$$

The above equation can be recast to show relevance of any object $o_i$ (having *bpa* $m_i$ and its representative terms $t_1, t_2, \ldots, t_n$) to the query *q* as

$$Bel_i(q) \quad = \quad \sum_{n \in q} \quad m_i(t) \qquad \qquad \ldots \quad (7.9)$$

The *Bel(q)* expresses the relevance since it is based on query terms supported by the object. Thus, *Bel(q)* can be used to rank the objects according to their relevance to the query *q*.

The plausibility expressed by equation (7.5) can be redefined in terms of query *q* for object $o_i$ as

$$Pl_i(q) \quad = \quad 1 \quad - \quad \sum_{q \cap s = 0} \quad m_i(s) \qquad \qquad \ldots \quad (7.10)$$

where *pl(q)* represents the extent to which the available evidence fails to disprove the relevance of the document to the query.

The following sections apply the D-S theory for document retrieval and information extraction.

## 7.4 Document Retrieval

This example illustrates the application of D-S theory for Document Retrieval using the query terms as evidences.

Consider the following problem:

Given (1) documents $d_1, d_2, d_3, \ldots, d_i, \ldots, d_l$,

(2) query terms $E_1, \ldots E_j, \ldots, E_m$,

(3) probability for $d_i$ as relevant document if query term is $E_j$,

$= p(d_i | E_j)$, for all values of $i$ and $j$,

(4) ignorance interval $I_j(d_1, d_2, d_3, \ldots, d_i, \ldots, d_l)$ for all query terms.

It is required to determine the order of relevance of documents for query terms $E_1 \cap E_2 \cap \ldots \cap E_m$.

The probability $p(d_i | E_j)$ can be determined as follows:

Step 1 – Counting the query term $E_j$ and its synonyms in the document $d_i$,

for $i = 1 \ldots l$. Let this be $n_{i,j}$.

Step 2 – Compute probability of relevance for $d_i$ as

$$p(d_i | E_j) \quad = \quad \frac{1 - I_j(d_1, \ldots, d_l)}{\sum_i n_{ij}} * n_{ij} \qquad \qquad \ldots \quad (7.11)$$

The value of ignorance interval $I_j(d_1, \ldots, d_l)$ accounts for possibly incomplete/ imprecise terms' frequency $n_{ij}$, of query term $E_j$ and its synonyms in the document $d_i$.

Let us consider the following typical numerical data for illustrating the computations.

$P(d_1|E_1)=0.4,$                 $P(d_1|E_2)=0.3$

$P(d_2|E_1)=0.0,$                 $P(d_2|E_2)=0.5$

$P(d_3|E_1)$=0.3,          $P(d_3|E_2)$=0.1

$P(d_4|E_1)$=0.1,          $P(d_4|E_2)$= 0.0

$I_1(d_1, d_2, d_3, d_4)$=0.2,     $I_2(d_1, d_2, d_3, d_4)$=0.1

For ease of computations, the data is rearranged as under:

|  | $d_1$ | $d_2$ | $d_3$ | $d_4$ | $I(d_1, d_2, d_3, d_4)$ |
|---|---|---|---|---|---|
| $E_1$: | 0.4 | 0.0 | 0.3 | 0.1 | 0.2 |
| $E_2$: | 0.3 | 0.5 | 0.1 | 0.0 | 0.1 |

Using the equation (7.7) following results are obtained for the combined evidences of $E_1 \cap E_2$:

| $d_1$ | $d_2$ | $d_3$ | $d_4$ | Empty set | $I_{E_1 \cap E_2}(d_1, d_2, d_3, d_4)$ |
|---|---|---|---|---|---|
| 0.22 | 0.10 | 0.08 | 0.01 | 0.57 | 0.02 |

Normalizing the above values with the help of empty set, we get:

$p(d_1|E_1 \cap E_2) = 0.512$

$p(d_2|E_1 \cap E_2) = 0.233$

$p(d_3|E_1 \cap E_2) = 0.186$

$p(d_4|E_1 \cap E_2) = 0.023$

$I(d_1, d_2, d_3, d_4|E_1 \cap E_2) = 0.046$

Hence, the document $d_1$ has the highest relevance of 51.2% for the evidence $E_1 \cap E_2$. The next document in the order is $d_2$ with a relevance value of 23.3%. It may be noted that when evidences $E_1$ and $E_2$ are combined, the ignorance level, which is 0.2 for $E_1$ and 0.1 for $E_2$, is reduced to 0.046 for $E_1 \cap E_2$. This is a logical result because the level of ignorance progressively decreases as more information in the form of evidences is obtained.

## 7.5  Information Extraction

Consider the document given in *Appendix*-5, which has been represented using a tree structure shown in figure 7.2. Objects $o_1$ to $o_3$ are composite. The raw information carrying objects $o_4$ to $o_6$ are shown along with their sets of index terms (*t=tree, w=water, a=animals, b=birds, g=tiger*).

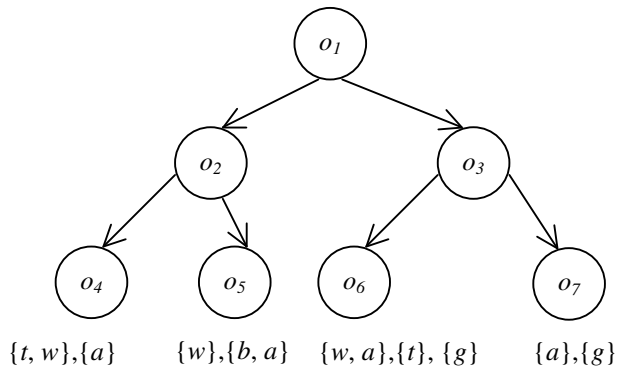

$\{t, w\}, \{a\}$   $\{w\}, \{b, a\}$   $\{w, a\}, \{t\}, \{g\}$   $\{a\}, \{g\}$

Figure 7.2: Typical Structured Document Representation.

The table 7.1 shows the index terms of raw objects (the leaf objects of the tree) and their associated *bpa*, where 'I' refers to level of *ignorance* assigned to whole of the set $\{t, w, b, a, g\}$. The sets of index terms representing objects correspond to the propositions from $p_1$ to $p_9$. For each term $t$, *bpa* $m(t)$ represents the belief that $t$ appropriately describes the contents of the object. The value of $m(t)$ for each object has been computed based on the frequency data of term $t$ in the corresponding object. $E_1$ to $E_4$ are evidences for different terms for raw objects $o_4$ to $o_7$. The aggregate belief of each raw object, for the given index terms, is calculated according to the equation (7.4).

Let us consider the query:

*Which is the maximum information covered if the documents $o_2$ and $o_3$ are browsed?*

The evidences in table 7.1 have been used to compute the combined belief in composite object $o_2$, using equation 7.7, and supported by the evidences $E_1$ and $E_2$ of composite objects $o_4$ and $o_5$ respectively. The results of this computation are shown in table 7.2.

**Table 7.1: Index terms, *bpa*s and associated uncommitted beliefs**.

| Objects $o_i$ | Evidences | Beliefs | | Total uncommitted Belief for |
|---|---|---|---|---|
| | $E_j$ | Index terms $\{t\}$ | Aggregate belief $\Sigma_{m_i}$ | I$\{t,w,b,a,g\}$ |
| $o_4$ | $E_1$ | $p_1=\{t,w\}=0.40$ $p_2=\{a\}=0.30$ | 0.70 | 0.30 |
| $o_5$ | $E_2$ | $p_3=\{w\}=0.35$ $p_4=\{b,a\}=0.50$ | 0.85 | 0.15 |
| $o_6$ | $E_3$ | $p_5=\{w,a\}=0.10$ $p_6=\{t\}=0.10$ $p_7=\{g\}=0.10$ | 0.30 | 0.70 |
| $o_7$ | $E_4$ | $p_8=\{a\}=0.45$ $P_9=\{g\}=0.30$ | 0.75 | 0.25 |

**Table 7.2: Combination of evidences $E_1$ and $E_2$ gives *bpa* for composite object $o_2$.**

| $E_1 \downarrow$   $E_2\rightarrow$ | $\{w\}=.35$ | $\{b,a\}=.5$ | I$\{t,w,b,a,g\}=.15$ |
|---|---|---|---|
| $\{t,w\}=.40$ | $\{w\}=.140$ | $\{\ \}=.20$ | $\{t,w\}=.06$ |
| $\{a\}=.30$ | $\{\}=.105$ | $\{a\}=.15$ | $\{a\}=.045$ |
| I$\{t,w,b,a,g\}=.30$ | $\{w\}=.105$ | $\{b,a\}=.15$ | I$\{t,w,b,a,g\}=.045$ |

The empty subset $\{\}$ signifies the propositions where no common semantics applies between $E_1$ and $E_2$. Under the assumption that the set of the semantics is complete, the empty set can be eliminated and non-empty set probabilities can be divided by the factor $1 - \Sigma p_0$, where $p_0$ is probability of empty set (equation 7.7). Thus, $1 - \Sigma p_0 = 1-(.2+.105)=.695$. The new values of *bpa* for object $o_2$ are given below.

$\{w\} = .352$

$\{a\} = .282$

$\{b, a\} = .216$

$\{t, w\} = .086$

$I\{t, w, b, a, g\} = .064$

The above figures represent composite object $o_2$ in terms of relevance weights of various index terms. For example, $\{w\} = .352$ indicates that object $o_2$ (combination of objects $o_4$ and $o_5$) has 35.2% information content about *water* resources. The ignorance level 6.4% indicates that remaining belief can be assigned to the universe of all the index terms.

On similar lines, the combination probabilities of object $o_3$, which are supported by evidences $E_3$ and $E_4$ for objects $o_5$ and $o_6$ respectively, are computed and shown in table 7.3.

**Table 7.3 : Combination of evidences $E_3$ and $E_4$ gives *bpa* for composite object $o_3$.**

| $E_3 \downarrow$        $E_4 \rightarrow$ | $\{a\}=.45$ | $\{g\}=.30$ | $I\{t,w,a,b,g\} = .25$ |
|---|---|---|---|
| $\{w,a\} = .10$ | $\{a\} = .045$ | $\{\}=.03$ | $\{w,a\}=.025$ |
| $\{t\} = .10$ | $\{\}=.045$ | $\{\}=.03$ | $\{t\} = .025$ |
| $\{g\}=.10$ | $\{\}=.045$ | $\{g\}=.03$ | $\{g\}=.025$ |
| $I\{t,w,a,b,g\} = .70$ | $\{a\} =.315$ | $\{g\}=.21$ | $I\{t,w,a,b,g\}=.175$ |

When the empty set is considered for normalization, the probabilities in Table 7.3 for object $o_3$ are as under:

$\{a\} = .418$

$\{g\} = .318$

$\{w, a\} = .029$

$\{t\} = .029$

$I\{t, w, a, b, g\} = .206$

The above results show that the document $o_3$ provides maximum information about animals. The information content regarding animals is 41.8%.

## 7.6 Discussion

D-S theory differs from the application of Bayes theorem in the following ways [40]:

1. Relations among the documents are represented as *Coarse documents* (class of document objects) and *fine documents* through tangled document hierarchies.

2. The computation is based on conditional query-document probabilities $P(d|q)$, instead of conditional document-query probabilities $P(q|d)$ in Bayesian probability.

Bayesian IR is best suited where (1) there is no need to represent ignorance, (2) conditioning is easy to extract information through the probabilistic representation, and (3) the prior probabilities are available. D-S theory is more suitable for situations where (1) uncertainty cannot be precisely ascertained to a proposition or its complement, (2) when conditioning effects are either impracticable to measure separately from the event itself or a simple propositional refinement, and (3) prior probabilities are not needed.

# Chapter 8

# Concluding Comments and Suggestions for Future Research Work

This thesis presents a variety of techniques for Document Retrieval (DR) and Information Extraction (IE) to meet the challenges due to information explosion. The choice of a technique is highly problem specific.

It has been shown that the semantic network based method using WordNet has resulted in highly satisfactory disambiguation. The reason is: for an ambiguous word, its *synset* (synonyms), hyponyms and hypernyms have been considered for matching with the sentential context of the ambiguous word. In addition, the WordNet synsets consists examples of sense tagged sentences based on the word being considered for disambiguation. Since these sentences are collected from the tagged texts, this technique gives reasonably correct sense in the given context. The WordNet architecture, which is based on semantic networks, provides pointers for the purpose of navigation among the related words. Thus the process of disambiguation is fast and reasonable. The sentence or the query, in which a word is to be disambiguated, should be long enough so that there is sufficient context available to help in resolving the ambiguity.

The use of fuzzy set theory for DR shows that the fuzzy relevance relation and fuzzy thesaurus are more expressive than their crisp set counter parts. The degree of association returned along with the retrieved documents helps the user to decide the order in which documents can be viewed. The Fuzzy Document Retrieval (FDR) promises higher potential for cross-language text processing and retrieval of document. Every language and its semantics have close association with the culture in which it has its roots, and therefore, semantically exact matching terms of one language cannot be found in another languages. In fact, only fuzzy relations exist between the matching words of the two or more languages.

This research work presents a new process of Question-Answering from a repository of Natural Language (NL) texts. The process is carried out in two phases. In the first phase, called *fetch* phase, the relevant documents are retrieved using theory of fuzzy sets. In the second phase, called *browse* phase, the answers of the questions are extracted from the retrieved relevant documents. It may be seen from the illustrative examples that the technique outlined in the method can be used to find the complete answer of a question from one or more relevant documents. The strength of the Question-Answering system presented lies in the use of Structured Description Language (SDL) and Synonyms, for the selection of the best possible answer.

A statistical model based on Bayesian Probability for DR and IE has been presented. It has been demonstrated that Bayes theorem can be used for representation of logical inferences in the form of Bayesian inference networks, which in turn can be used for DR and IE from natural language texts. The necessary algorithms for determination of relevancy of documents as well for Information segments have been presented. The results strongly support the mathematical theory presented in the thesis. The limitations of this approach are as under:

1. It is assumed that query terms are independent of each other, which is a basic requirement for Bayesian inference network for DR-IE. This can be easily ensured by proper selection of query terms.

2. To ensure that no relevant document is missed from retrieval a query should have reasonably large number of terms.

3. The accuracy of retrieval depends on the accuracy of representation of query and documents.

The thesis presents a new technique for DR and IE using Dempster-Shafer theory of Evidence. The belief functions are used to compute the probability of relevance of a document for a given query term as evidence. It is found that D-S theory of evidence provides more effective tool for the retrieval of documents and extraction of information from Natural Language texts, since the theory allows consideration of uncertainties and ignorance.

Efforts in future may be directed to improve the confidence level of the answers by integrating the probabilistic and possibilistic uncertainties associated with the NL texts. This integration can be done in the following alternative ways: -

Let P and μ be the probability and possibility measures, respectively, of the terms in the text.

*First alternative:*

Integrated uncertainty (Prosibility) = $P \times \mu$

*Second alternative:*

Integrated uncertainty = $\dfrac{1}{\sqrt{2}} \left( P^2 + \mu^2 \right)^{1/2}$

It should be of considerable interest for future research work to develop a feedback model for IE. The query and the answer are reviewed by critique. If the extraction of the answer is satisfactory, then the process of IE is accomplished. Should additional information be needed, the process of fetching and browsing continues till the required information is extracted. This approach closely follows the feedback control theory. It can therefore, easily be extended for developing Intelligent Information Extraction Systems (IIES) as shown in figure 8.1.

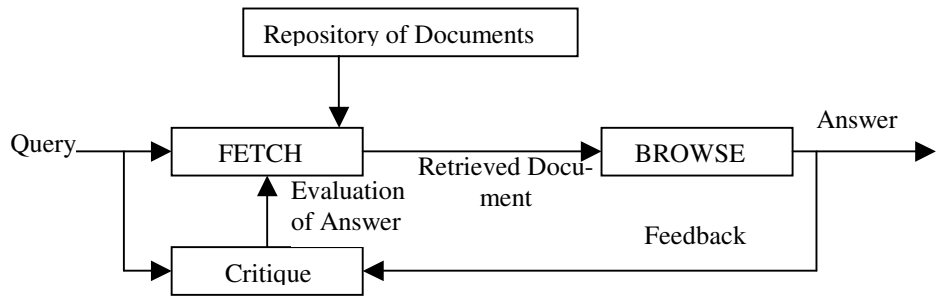

Figure 8.1: Intelligent Information Extraction System.

---

*Keywords₁*={Akbar, religion}

**Document D₁**:  Akbar was a great administrator. His administration strove for the welfare of his people irrespective of their religious beliefs. He had council of ministers and high officers to assist him. The ministers were heads of different departments. The revenue source for the Mughals was from land and trade. Akbar followed a liberal policy for religion. He believed that every religion was basically good.  He used to hold talks with the leaders of various religions. Akbar found that basically all religions taught similar things. Thus, he incorporated the principles of all the religions and found a new faith, which he called as Din-I-lahi. This was more a code of morale conduct than religion. It was his firm belief that all religions are same; therefore, he did not seek forced conversion to Islam.

Keywords₂={Jahangir, art and justice, trade}

**Document D₂**:  Jahangir was a highly learned man but he did not have the intelligence of his father. He was a lover of art and justice. A golden chain with bells attached, were tied to the wall of his palace. Any one who pulled the chain was heard and was accorded justice. Jahangir married Nur Jahan. She was not only peerless in beauty; she was also a highly intelligent and cultured lady. She was fond of music, poetry, and painting.  During the Jahangir's time, many Europeans traders started coming to India and trade started with them.

*Keywords₃*={Shah Jahan, Mughal architecture, art}

**Document D₃**: Shah Jahan was a great general of his time. Shah Jahan brought glory to the Mughal architecture. He was a man of finest tastes. He built Taj Mahal at Agra, the new city of Delhi, which he made his capital. He also built the Fort at Delhi, named as Red Fort and a mosque named as Jama Masjid. He built finest gardens. Under Shah Jahan, the art of painting also developed significantly. Shah Jahan gave liberal patronage to artists.

# Appendix –2

## Document d01.txt

A cover for your home loan[1]

TIMES NEWS NETWORK [THURSDAY, NOVEMBER 28, 2002 12:25:40 PM]

Mr Raman, a senior executive at an MNC, walked straight into an insurance office after buying a property. He wanted to buy an insurance policy worth a few lakh rupees even if it meant an additional liability. The question is why did he want another long-term liability when he had just taken on one big one with the home loan? The difference with this liability is that it takes care of your other liability. Sounds confusing? Let's get down to it.

Over the last few years, a large number of people have been taken home loans. There were good reasons for doing so, what with interest rates falling, lenders wooing customers aggressively, and real estate prices too being driven down by competition. As a result, any individual holding a decent job could think of buying his own home.

The good news is that housing finance companies and banks, which earlier used to lend only 75-85 per cent of the project cost, are willing to finance up to 90 per cent. And many are willing to customize the loan to specific needs. The borrower can pay a lower installment amount in the initial years and increase it later along with a rise in his income. The result: people have ended up borrowing much more than what they would have done a few years ago. In fact, every bank and housing finance company says that its average home loan size has gone up in the last two years. The figure has risen from less than Rs five lakh two years ago to nearly Rs 8 lakh now. Companies are even willing to give you a home loan as high as Rs 1 crore if they feel you can repay it.

That is probably why Raman decided to take up a home loan. After borrowing Rs 30 lakh, he decided to take up an insurance policy. He didn't want to leave a huge liability for his family in the event of his death. His family would lose their dream home should they fail to clear the outstanding loan amount. A safer option was to take an insurance policy.

---

[1] These documents are collected from Times of India, Electronics edition –indiatimes.com, from the editions of dates mentioned against them and have been reformatted to suit the presentation.

Unfortunately, people tend to look at an insurance policy for its tax benefits, or consider it as a long-term saving product. An insurance policy can also be used as a cover against various liabilities. Home loan being one of the most expensive and long term of liabilities, an insurance policy is a must for covering it. One way of doing this is by taking an endowment policy for an equivalent amount. The drawback is that you will end up paying a higher premium on this product even if you get survival benefits. Since the primary objective of a home loan borrower is to take care of the loan outstanding, it is better to opt for a product that provides term cover at a much lower premium.

Various insurance companies have launched term products with the specific purpose of providing cover for the home loan liability (see table). The big advantage is that your premium is much cheaper compared to other insurance policies. For instance, a 30 year old home loan borrower can opt for a product that offers a cover for 10 years by paying a one-time premium of less than Rs 1,600. Of course, the premium to be paid differs according to the age and the outstanding loan amount. This again varies from one company to another.

LIC's Jeevan Griha, for instance, offers double and triple cover. If the loan amount is Rs 30 lakh, the borrower needs to take a cover for only Rs 15 lakhs and Rs 10 lakh respectively. In the even of death, the sum assured will double or triple according to the product he chooses. If the insurer survives, he will be paid only the basic sum assured.

The private sector insurance companies have designed their home insurance products with an eye on the huge home loan segment. Here the premiums are much lower and the customer has the option of paying premiums one time or at regular intervals. The survival benefit depends on the mode of premium payment. If your worry is the huge home loan liability, you shouldn't be worried too much about the survival benefit. And also, it is time to take a policy if you don't have one even after taking a home loan.

## Document d02.txt

Budget to avoid going broke
 [TUESDAY, NOVEMBER 19, 2002 12:16:14 AM]
MITRA JHAVERI

If you think most Union finance ministers goof up in their annual budget making exercise, take this simple test. Try to budget your income and allocate sums for expenses for a whole year beginning January 2003.

At the end of the year, check whether you managed to stick to the budget. I bet that you will have completely strayed from your proposed budget. Now imagine what it is to budget for the nation. You may or may not agree with a finance minister's policies, but you have to pity anyone in that office.

However, don't give up hope on the home front. The rather old fashioned idea of budgeting your expenses still works to help you grow rich. Unless you know how much you spend, you will never know how much you can save. Moreover, unless you save, you will never become rich. So, begin to budget income and expenses right away.

Of course, merely creating an attractive-looking budget, putting it up on the living room wall, and looking at it every morning will not help. Nor will downloading programmes from the net with colourful bar charts, most of which are meaningless. You need to apply your mind while creating your budget as well as in following it.

Budget-making is a complicated exercise, though of course your home budget won't be anywhere close to the complexity of the Union government budget. You have to take into account each major expenditure in order to make it realistic and workable.

Here are a few tips:
• The first step in creating a budget is to start writing down your daily expenses for a period of a month in order to know the pattern of your expenses. This will help you create a model database of expenses to work on. Take care to select a month when you will not have too many extraordinary expenses. For example, skip November, which has its inevitable Diwali splurge - too much of deviation for your trial run. You will find expenses skewed towards gifts, garments, consumer durables, etc.
• The next step is to sit down on your PC - preferably with the Microsoft Excel programme - and create a broad category of your expenses (like children's education, medical expenses, groceries, etc).
• The third step is to further break down these categories into individual expense items (like under the head 'health', you will have individual expense items like medicines, doctor's bills, contingency expenses for unforeseen illnesses.
• The fourth step is to calculate precisely how much you would spend on each individual item of expenditure per month. Multiply this by 12 to arrive at your annual expense for that item. Of course, things like doctor's bills, medicines and medical contingencies cannot be calculated pre-cisely. However, these may be worked out based on your opinion of the health conditions of family members and an estimate of past expense levels.

• The fifth step is error correction. Look at all the expense heads once again and make sure nothing has been left out. Now come to the other side - your income. Don't consider any income, which is not sure of being received like a bonus that you are expecting if your company does well, a gift from your in-laws or parents, contest proceeds, etc. Allocate the definitely receivable income to relevant expense items. However, before you do that, keep aside 10-15 per cent for investments.

If you still have a surplus after all expense allocations, put more money in investments. On the other hand, if there is a shortfall, don't cut investment limits. Cut some flab from various expense heads, but try and not to compromise on your investment allocations.

The final and most important step is to compare your actual expenses with your budget once a fortnight and make sure you stick to the budget. If there is a consistent variation, find out why it is so. Is it because you have not accounted for some expense heads or have underestimated your expense figures? Is it because of a genuine unforeseen expense, which was not affordable in the first place? If it is the latter, then you can safely ignore it since this will not recur. If it is the former, rework your budget accordingly.

# Document d03.txt

Microsoft course

RASHMI SABLANIA

TIMES NEWS NETWORK [THURSDAY, NOVEMBER 28, 2002 01:55:11 PM]

Microsoft Corporation India Pvt Ltd has recently launched the Academic Developer Programme (ADP) in India. The programme is aimed at empowering universities, faculty members and students with the necessary skills, resources, training and real life experience, to leverage the emerging XML Web Services revolution.

According to Dilip Mistry, director .NET & Developer Evangelism, Microsoft, "Academic Developer Programme is an umbrella programme of Microsoft. India needs to produce graduates with skill-set, which can enable them to move up in the value chain. We are working with different technical institutions and this kind of programme is required to reduce the gap between the industry and the academia."

ADP has introduced eight key initiatives under this programme. Mistry said: "In faculty internship, we are offering sponsorship to PG students to stay on as a faculty member, making teaching financially more viable."

Under the Microsoft .NET Campus Challenge, students across India are invited to participate in building real-life projects identified by corporate houses on .NET. "It will further include conducting technical seminars, delivered by the representatives of the industry," he added. The MSDN Academic Alliance will cover the entire campus. "

Some of the modules of ADP are currently being implemented at Anna University, Chennai covering its 240 affiliated colleges and Visveswaraiah Technological University in Karnataka, covering 102 college affiliated to the university.

# Document d04.txt

Distance education made easy

PRASOON PANT

TIMES NEWS NETWORK [THURSDAY, NOVEMBER 28, 2002 02:20:30 PM]

The Indira Gandhi National Open University (IGNOU), an apex 'open learning and distance education' institution in the country has in recent times taken a giant step in the area of distance education.

They are 'Gyan Darshan'-- the educational TV channel and the 'Gyan Vani'-- the FM radio network for education and development in the country. In India, where the challenge is to bring backward communities and the rural populace under the ambit of education, these channels are like a boon as they can be accessed with the help of an ordinary television set or an FM radio.

The brainchild behind IGNOU's effort to set up these audio--visual channels is R Sreedhar, director, Electronic Media Production Centre (EMPC), IGNOU. Sreedhar has an experience of more than 22 years in the All India Radio (AIR), and is a recipient of several 'Akashvani Awards'. He has been at the forefront of setting up these channels from the conceptualization to the implementation stage. Sreedhar says, "The 'Gyan Vani' is supposed to be a network of 40 FM educational channels, across the country, of which seven are already operational. The 'Gyan Darshan' on the other hand is a single TV channel that broadcasts various educational programmes, like the course contents of many educational institutions."

"The aim is to optimize the reach and availability to the student community. At present, the course material of IGNOU and various other institutions is disseminated through these channels. The main advantage is that students just have to tune in to their FM or television sets and the course modules are within their reach -- free of cost," he adds. Besides IGNOU, institutions like IITs, IIMs, UGC, NCERT are using these channels for their course instruction.

According to Sreedhar, 'Gyan Vani' is becoming popular due to the mixed content of programmes aired on it. So, it's not just a drab list of boring lectures but a judicious mix of educational, cultural and musical programmes to lift up the mood of a listener as well as fulfilling his or her educational needs. "Besides, one of the unique selling propositions of the programmes on 'Gyan Vani' is the interactive session where a student sitting anywhere in the country can ask questions to experts, who participate in the programmes," says Sreedhar.

"The future of FM radio seems bright in India and hence the fast acceptability of education through this mode is apparent. We believe that more cable operators would broadcast 'Gyan Darshan' as well. Educational institutions across the country could approach us if they desire to put their programmes 'on air' at virtually no cost, for the first year of broadcast. These institutions can send in taped versions of their course modules to IGNOU for telecast on 'Gyan Vani' and 'Gyan Darshan'," remarks Sreedhar.

# Document d05.txt

Innovations in education

ARUN SHARMA

TIMES NEWS NETWORK [THURSDAY, NOVEMBER 28, 2002 01:51:02 PM]

The 'Innovative Education Initiative' (IEI) of Intel focuses on preparing today's teachers and students for tomorrow's demands. In 2001, it contributed about US $ 120 million to primary, secondary and higher education across the world. Excerpts from an interview with Aruna Ramanathan, head of education programme, Intel-India.

Why has Intel chosen education as its primary external focus?

Education is a critical focus area for Intel. We believe the economy of the future depends on the quality of education in our classrooms. Our initiative is focused on improving science, mathematics, engineering and technology education. It is important that we have a science savvy population who can design the next generation of products.

What is the objective of the company's IEI?

To prepare today's teachers and students for tomorrows demand. Under it, Intel develops and supports education programmes, which aim at improving effective use of technology in classroom teaching, improving science, mathematics, engineering and technology education and broaden access to technology. Our initiative is based on providing the benefit of technology to students in the under-served areas, and implementing a professional development programme for teachers.

Intel believes the same spirit of innovation that drives the global economy can also achieve dramatic results in the classroom. To put this vision into practice, we are actively working with governments, educators, and students to harness the technology.

Which programmes come under the IEI in India?

The key plans are: 'Intel International Science and Engineering Fair', which aims at infusing a spirit of discovery in school children and increase their interest in science and technology. 'Teach to the Future' aims at enabling teachers to integrate the use of computers into subject teaching.

The 'Community Development Programme' aims at training students and unemployed youth from the lower income segments for their future careers. It moreover, aims to provide free and easy access to quality job-oriented training on diverse new technologies.

## Document d06.txt

Plastic money for your family

TIMES NEWS NETWORK [THURSDAY, NOVEMBER 28, 2002 12:18:54 PM]

Severe competition has prompted credit card outfits in India to dole out attractive incentives. It is also a normal practice to offer add-on or supplementary cards. That leaves you with a question: Is it better to opt for a supplementary card or go in for a new card, especially if both spouses are eligible for a credit card?

The basic advantage of an add-on card is that even if your spouse or any other family member is not eligible to have his or her own credit card, you can gift the person an add-on card and he/she will be able to enjoy the advantages of plastic money. Eligibility criteria to for obtaining a new credit card vary across banks. Typically, there is the minimum income criterion. A credit card owner can apply for a supplementary card for his or her spouse, brother, sister, parents or child (above 18 years).

However, don't go overboard if you are going in for multiple add-on cards lured by the 'free' add-on. Here are some tips for you in the selection process.

• Look closely at the annual charges. In all probability, they will be payable for the add-on cards-at least after the first year.

Also, don't forget that you are responsible for clearing the outstanding amount on the add-on cards as the responsibility for payment is yours (as far as the credit card company is concerned). Annual charges for supplementary cards differ across issuing organizations from Rs 150 to Rs 1,250.

• To lure members to use cards more often, credit card companies have promoted the concept of reward points. For example, for every Rs 100 worth of purchases made through the credit card, a Citibank cardholder gains one point. These points could be redeemed for gifts or products and even for paying the annual fee. Standard Chartered rewards one point for Rs 125 worth of purchases made through the card. If you are planning to gift an add-on card, check the fine print closely. Are the add-on members eligible to earn reward points? If your spouse is likely to use the supplementary credit card regularly to pay monthly bills, purchase household items, etc., it may make sense to go in for a new card if your credit card company does not give reward or bonus points to supplementary card holders.

• A supplementary card does not change your credit limit. Your credit limit is divided between the main card and the add-on cards. In contrast, a new card increases your credit limit.

• Features like personal accident insurance for air, road or other accidents are packaged along with the subscription. The kind of credit card you hold will determine the type of insurance cover and amount that you will be offered. In general, gold cardholders are covered for higher amounts as against the silver or executive cardholders. Standard Chartered offers a Rs 15 lakh air and Rs 5 lakh general insurance to its Gold cardholders, plus an add-on accident cover worth Rs 5 lakh to its members. ICICI Bank also provides personal accident cover to its Gold credit card members (air accident cover of Rs 15 lakh and general accident cover of Rs 5 lakh). Besides, ICICI Bank also provides insurance cover for the supplementary card members (air accident Rs 6 lakh and general Rs 2 lakh). Some of the cards cover insurance for spouse, irrespective of whether they are supplementary cardholders.

In addition, baggage cover, purchase protection cover and credit shield are issued free of cost to card members. The baggage cover feature is useful for frequent flyers. Purchase protection feature automatically insures all items bought on the credit card from damage or loss due to fire or theft, for a defined sum of money, which depends on the type of card you have.

In the unfortunate event of death of the cardholder, the onus of payment of the outstanding credit card bills falls on his or her spouse or parents. The credit shield feature provides for a waiver of payment of such outstanding amount on the card up to a certain limit. Check the fine print: A supplementary cardholder may not be eligible for all these covers.

• Go only for features that make sense to you. If your spouse is not going to be traveling by air regularly, air accident insurance or baggage cover for an add-on member is hardly relevant.

• Check for important clauses like lost or misused card liability.

• The annual fee for a new card is obviously more than that for an add-on card. The annual fee for the HSBC Gold card is Rs 2,000 (50 per cent off for the first year) and for the add-on is Rs 1,000 (waived for the first year). But check the eligibility criteria, other benefits and the freebies offered. Citibank gives a Pond's gift hamper free on subscription to its Citibank Women card. Decide whether your partner or family member is better off with his own card or will an add-on card serve the same purpose.

# Document d07.txt

Insurance for the insurer

TIMES NEWS NETWORK [TUESDAY, NOVEMBER 19, 2002 12:06:38 AM]

MADHU T

Barring a marriage, nothing else beats an insurance contract, particularly life insurance, in terms of length. It runs for 15, 20 or even 30 years. This raises the obvious question: what if your insurance company was not around to honour the claim?

After all, insurance companies also go bust all over the world. Fear not. If your policy is 'reinsured', then even if your insurance company were to go under, you money is safe.

What on earth is reinsurance? It is a kind of risk cover that insurance company takes. When an insurance company fears that large claims from a particular type of policy can wipe out its corpus, it turns to reinsurance companies.

Reinsurance companies specialize in covering the risks of insurance companies - both life and non-life. They charge a premium for that. Just like the premium you pay to the insurance company on your policy, the insurance company pays a premium to the reinsurance firm for the risk cover.

There are global reinsurance giants such as Swiss Re, Munich Re, Cologne Re, etc. Our own General Insurance Company (GIC) is also busy turning into a reinsurance company. General in-surance companies always used to reinsure their "high risk" products. However, before the opening up of the insurance sector one rarely heard the word associated with life insurance policies of Life Insurance Corporation of India (LIC).

"It is not true that we never reinsured our products. It is just that we didn't have to advertise it to convince our customers to buy our product as the question of LIC not paying was never there," points out an LIC official adding, "LIC always used to reinsure high-value policies."

"It is an international practice and also a good business practice," says Saugata Gupta, chief marketing officer, ICICI-Prudential Life Insurance. "It makes business sense to have your high-value policies reinsured as large claims can affect your financial strength," agrees Pankaj Seith, head of marketing, HDFC Standard Life Insurance. Both companies reinsure their policies above a "certain cut-off" with Swiss Re, a reinsurance global giant.

"It is vital for these (private) companies to reinsure their large policies because any big claim can wipe out their small corpus and capital," agrees the LIC official. Which are the products that drive insurance companies running for reinsurance cover?

"Any high-value policy which is many times above the average size of our policy," replies a private life insurer who prefers anonymity. For example, a life insurance company would surely reinsure the Rs 20 crore covers bought by our Bollywood heartthrobs. Covers offered on credit cards are another segment which companies are eager to reinsure.

How do insurance companies arrive at the cut-off figure to reinsure? "It depends on the capacity of the insurance company to absorb risk," says Seith. "We have different cut-offs for different products depending on our risk perception." Insurers say it varies from company to company and changes from time to time.

"A lot depends on the experience of the company and the norm followed by its foreign partner," says the private insurer. OM Kotak, for example, is said to reinsure policies above Rs 5 lakh in general while LIC is said to generally reinsure covers above Rs 25 lakh.

What does it mean to you? One, your money is safe because the risk of your insurer going bust stands eliminated. Two, your premium will be slightly loaded, but that is not something you should really mind, right?

Moreover, here is a tip: large risk covers on term policy, the cheapest form of insurance because it covers pure risk, are usually reinsured.

"We charge very low premium on term plans, taking a huge risk. So it is in my interest to derisk," says a private insurer. This means you get a large cover for a cheap rate and it is also re-insured from the risk of your insurance company going under. Did someone say you can't have the cake and eat it too?

# Document d08.txt

JDMC book week

TIMES NEWS NETWORK [THURSDAY, NOVEMBER 28, 2002 02:10:16 PM]

JDMC book week

For the fourth consecutive year Janki Devi Memorial College (JDMC) organized the annual book week from November 11-15, 2002, an unique event in Delhi University. The Book Week was inaugurated by the Chairperson of the Governing Body of JDMC, Kusum Krishna and Usha Mujumunshi, Head of INSA. A series of activities were organized in the course of the week, which included Book Quiz, Book Jacket Design Contest and a Book Hunt.

A discussion on copyright law was attended by a distinguished panel of speakers: S Ansari, S B Ghosh and C P Vashisht. A book reading session was held on the second day of the week where Gyanpeeth Award winning writer Indira Goswami and a noted Bengali poet and novelist Nabanita Deb Sen read out from their works.

Management convention

'Managing Business Tomorrow' was the theme of seventh Annual Management Convention organised by Noida Management Association on November 22 at the Power Management Institute, Noida. The convention was organised under the chairmanship of R V Shahi, secretary, Ministry of Power, Government of India.

Different aspects of 'Managing Business Tomorrow' were discussed in three engrossing sessions. The speakers in these sessions included S K Narula, Chairman, Airport Authority of India; P K Gupta, CEO, Modi Industries Ltd; Mahendra Swarup, CEO, Indiatimes.com; Sanjeev Bhikchandani, CEO, Infoedge and Sudershan Banerjee, CEO, Hutchison Essar.

GGSIPU colloquium

Guru Gobind Singh Indraprastha University (GGSIPU) had recently organised a colloquium on the 'Challenges facing higher education in todays changing scenario'. Speaking on the occasion, University Grants Commission, chairman, Arun Nigavekar said that higher education is a basic and essential tool for building social and economic infrastructure in an emerging nation like India. He stressed on the need to make higher education accessible in the rural and semi urban areas, besides making degrees in arts, humanities and commerce relevant and utility oriented.

In his address vice-chancellor, GGSIPU, K K Aggarwal urged for a system, which will allow students to enroll in multiple courses so that they are able to enhance their skills.

ICEM workshop

The International Centre for Event Marketing and Management (ICEM), an NGO, had recently organized a two-day workshop for the police officers on meditation and stress management techniques. Inaugurating the workshop, DCP, North West, Sanjay Singh said that these programme are beneficial for the officers, who after serving long working hours, get stressed and tired. He moreover, appreciated the efforts put by ICEM in putting this on police agenda and said similar functions will be carried in future as well. According to Prince Singhal, president, ICEM, the workshop covered topics such as stress relaxation methods, spiritual discourse, meditation techniques and question answer session.

# Document d09.txt

Publishing industry calling

TIMES NEWS NETWORK [THURSDAY, NOVEMBER 28, 2002 12:53:36 PM]

Publishing in India is a vibrant Rs 7,000 crore industry growing at the rate of 15 per cent to 20 per cent every year. This is one industry, which provides immense career opportunities for youth today looking for an option away from the world of engineering, medicine, banking, civil services and many more. However, lack of knowledge about career options in the publishing industry has failed to attract the best people in a sector where excellence is the hallmark in a world driven by information technology.

As of now, there are very few courses offered on book publishing in the country. Sensing this need, the Federation of Publishers' and Booksellers' Associations in India (FPBAI) has decided to start a three-month 'Certificate Course in Publishing' at 'The FPBAI School of Publishing Sciences' in the Capital from January 2003. Sukamar Das, President, FPBAI said: "The objective of this programme is to train young and fresh graduates with an aptitude for publishing and shape a career in publishing and book-selling. Also, a specialised course in publishing will help to make the functioning of Indian publishing industry more professional."

Moreover, this programme offers a well structure and organised institution and curriculum for those already working in the publishing and book-selling professions but have not had an exposure to a course in publishing. "We have devised this programme in such a manner that it offers an opportunity to the human asset in the publishing industry and book trade to develop new skill and refine existing ones on a continued basis consistent with the advances in technology and changes in the business environment," added Das.

The three-month programme focuses on different aspects of publishing and fee has been fixed at Rs 18,000.

Visit: fpbaindia.org

## Document d10.txt

Need-Based Selling Leads to High Persistence

TIMES NEWS NETWORK [THURSDAY, NOVEMBER 28, 2002 12:31:05 PM]

Private insurance companies have had a good 18 months so far. In the markets they operate in, they have wrested about 10 per cent market share, which is better than what private banks or mutual funds managed to achieve when these sectors were opened.

But this is only the first hundred meters of what promises to be a long marathon. Gary Comerford, VP, International Marketing at Sun Life Financial and Director, Birla Sun Life Insurance discussed with Sanjay Kr Singh the implications for the Indian customer of changes in the insurance market.

How well have unit-linked products been accepted in the Indian market?

At a press conference I had done five years ago in India, I had made a commitment to bring the most innovative products to India. And unit-linked products are without doubt the most innovative. Because of its inherent flexibility, a unit-linked product allows you to customize the product to suit customer needs. A lot of Birla Sun Life's advertising also focuses on the flexibility element.

Do you feel there is a need to educate the market about unit-linked products? Because in India unit-linked products are widely perceived as synonymous with the stock market. When you are innovative, you also have to be a missionary. Let me also add that we offer three investment options -- where the equity components are 10, 20 and 35 per cent -- allowing individuals to opt for the appropriate level of risk they wish to take. These to my mind are very prudent limits.

Another point I would like to make is that Indian stock markets have without doubt been hammered hard over the last eighteen months. But the Indian stock market is also under-valued at present. So you also don't want the Indian consumer to miss out on the growth that is inevitably going to occur.

The Indian economy is growing at five per cent. Only recently when you had the drought there was talk that it would lead to recession. But the recession has been averted because the other sectors of the economy are robust. So there are good reasons for investing in the Indian economy. All we do through unit-linked products is to give consumers the choice regarding how they want to do this.

In the Birla Sun Life's scheme of selling insurance, how much attention do you pay to needs analysis?

Only today morning I had a meeting with our top financial advisors in our company, and I asked them that what makes you successful. And the very first one answered that it is the needs analysis we do.

The capital-needs analysis we train our advisors to do looks in a sophisticated way at what a customer needs by way of life insurance. You could simply say that the customer should buy insurance cover worth so many times his salary, or worth Rs 1,00,000, or any such figure. But we want to sell you life insurance looking at your assets, your expenses and your financial goals.

Early indications show that Birla Sun Life is enjoying a good level of persistency. Good persistency is caused by a product being properly sold, based on the customer's needs, so that they make the second, third, fourth and further years of premium payment. People don't keep a product that they don't like.

Do you plan to enter the pension market?

Just as we believe that life insurance is a core component of a person's financial portfolio, we believe the same about a good pension plan. We plan to be a substantial player in the private pension market and will introduce innovative products there too.

Since you are into both mutual fund and life insurance business, do you have any plans to offer comprehensive wealth management services?

We offer many components of a comprehensive wealth management service. Look at Birla Sun Life Mutual Fund, which offers cash, equity and debt products. And we have Birla Sun Life Securities to deal with the portfolio needs of more sophisticated and high net worth individuals.

What would you count amongst Birla Sun Life Insurance's achievements, and what are the immediate challenges?

We have at present the highest average premium per policy sale. Also, we were the first to introduce the 15-day free look-in period. Now I believe it is being made mandatory through legislation.

We are also very satisfied with the quality of training we have offered to our sales advisors and to our corporate agency force. And we have met most of our projections.

As we grow rapidly, the greatest challenge would be to maintain quality control -- to have the training and management in place for maintaining very high levels of service and expectations.

# Document d11.txt

Revving up

SURESH CHANDRA

TIMES NEWS NETWORK [THURSDAY, NOVEMBER 28, 2002 02:22:57 PM]

An increase in the number of vehicles and freight traffic has led to work openings for automobile engineers in both salaried and self-employment categories.

Automobile Industry is the barometer of the economic and social transformation in the country. In fact, being linked to road transport, automobile industry is playing a vital role in the national face-lift.

The industry is one of the major sources of employment and promises to generate employment for thousands of new hands directly and indirectly. It has been influencing various sectors of the economy and services such as agriculture, aviation, fire services and many more spheres of our social and national activity. The automobile industry provides a medium of moving men and material across the geographical spread of the country.

An automobile engineer is a key personnel of the automobile industry and is responsible for repair, maintenance, design and manufacture of vehicles, such as cars, trucks and buses, vans, tempos, mopeds, motor cycles, scooters and tractors. He or she has to look after proper, efficient and economical running of the vehicles. For this, the engineer tests the performance and efficiency of the engine chassis and the transmission system apart from other components. He or she ensures proper lubrication of vital parts and makes adjustments in the mechanism to ensure optimum performance. At times, he or she may overhaul the vehicles by dismantling them and after cleaning and removing defects and repairing/ replacing worn-out parts, reassembles them. There-after, testing and adjusting them for smooth running is carried on.

An automobile maintenance engineer exercises general supervision over mechanics and other workers in the workshop, factory or garage and is responsible for maintenance and accounting of stores. He may moreover design new models, keeping in view their performance, capacity, dura-bility and cost of manufacture and maintenance. For this, he or she studies the data of the working of previous models, introduces fresh ideas, makes new computations, and prepares working draw-ings of the new model. Subsequently, the working model is tested to eliminate defects and short-comings.

Openings

Numerous private motor repair and maintenance service stations, automobile garage and work-shops offer job opportunities to automobile engineers. Private transport companies, state road transport undertakings and defense services offer a large variety of maintenance jobs. With the establishment of many automobile and ancillary manufacturing and assembling plants, both in the private and public sectors, employment avenues are expanding in the field of design and manu-facturing of automobiles. Insurance companies, covering risk on motor vehicles, offer jobs as 'claim inspectors'. Opportunities exist on the sales promotion side and in the teaching departments of training institutions.

As for self-employment opportunities, an automobile engineer may start a repair-workshop or garage in a town or a service centre for the repair, maintenance and hiring of tractors and other agricultural machinery in rural areas.

Training

The study of automobile engineering involves understanding the mechanics of vehicle chassis, internal combustion engine, electrical system of vehicle, motor transport problems, workshop technology, research and design.

The department of training and technical education, Government of Delhi, offers three years' diploma course in automobile engineering at -- G B Pant Polytechnic, Okhla, New Delhi-20 and Pusa Polytechnic, Pusa, New Delhi-12 and at one privately-managed polytechnic -- Guru Teg Bahadur Polytechnic, Vasant Vihar, New Delhi. Admissions are open to matriculates with an ag-gregate of 50 per cent marks in science and maths through a written test conducted by Guru Go-bind Singh Indraprastha University, New Delhi. A post-diploma course in automobile engineering is one to one-and-a-half years' duration.

A degree programme in engineering is of four years' duration and is open to 10+2 with PCM. The admission is through a written test. In various engineering colleges, students taking up mechanical engineering courses can take automobile engineering as an optional subject. A diploma holder may appear in the examination conducted by the Institution of Engineers (India). Those who pass section A and section B of the Associate Membership Examination of the institution are eligible for higher jobs on a par with degree holders in engineering.

## Document d12.txt

Will Kelkar's recommendations lead to lower saving?

[THURSDAY, NOVEMBER 28, 2002 12:15:04 PM]

The much-awaited report by the task force on direct taxes under the chairmanship of Vijay Kelkar has failed to meet the expectations of the business community, the salaried class and also agriculturists as far as the tax reform part is concerned.

On second thoughts, this report does not discuss tax reforms in detail but concentrates on tax administration and governance, where undoubtedly it has suggested seminal changes.

The Kelkar committee report needs to be commended on the suggestions it has made on tax administration, like computerization, networking, and most importantly, outsourcing. The Income Tax department has a full-fledged computer department, which has been working for the past several years to computerize assesses' database and issue PAN numbers. However, as the environment in the department is not very conducive for such measures, even PAN numbers have not yet been allotted to all. India has emerged as a major outsourcing hub in the international market. There is no reason why domestic resources and expertise can not be tapped for the purpose of tax administration.

The suggestion for establishment of Tax Information Network (TIN) is a welcome step and will prove to be of immense value to both the government and citizens. The utilization of TIN for filing returns will change the shape of things. The main benefit will go to salaried employees and to senior citizens who will be saved the tedium of visiting the Income Tax department office for getting refunds. In a computerized environment, the refunds will be automatically downloaded by the intermediary and paid through bank draft or the electronic clearance mechanism.

On tax reforms, the committee has given recommendations without understanding the basis on which the present Income Tax Act provides various reliefs and deductions from the total income of the assessee. The report has followed the taxation laws of countries like the US, where the socio-economic scenario is different from that prevailing in our country.

The suggestion calling for withdrawal of standard deduction, relief u/s 88 for contributions made to provident fund/LIC, etc., and deduction on home loans, are going to hit salaried employees severely. In the light of recommendations for withdrawal of these deductions, the increase in taxable income limit from Rs 50,000 at present to Rs 1,00,000 will actually result in more burden on the salaried assessee in most cases.

Similarly withdrawal of deductions allowed for investment in infrastructure projects, for setting up industrial units in backward areas, units established in Software Technology Parks, Special Economic Zones and 100 per cent Export Oriented Units is likely to be reflected in poor economic growth in the country.

Another administrative reform suggested by the committee is to allow autonomy to the Central Board of Direct Taxes (CBDT). It suggests that CBDT should sign an MOU with the government for collection of revenue. This would amount to privatization of the revenue department, and may result in unfortunate situations where CBDT's at times ambitious targets may not match the growth projections of the country.

One suggestion is to allow search and seizure operations and not allow assessees to escape prosecution even if they pay taxes on hitherto concealed income. This suggestion is brutal and is likely to adversely affect the business environment. Search operations are an extreme option. They invade the privacy of an individual. And moreover, the improvement in tax collection from such actions has not been very encouraging.

The committee has also acknowledged this by stating that the collection of taxes due to search operations is less than 1 per cent of total revenue collection. The fear of such action is actually a greater deterrent than the action itself. More taxes can be collected if open inquiries are conducted in case of cases of tax evasion. The main benefit of such open inquiries is that the assessee can be told to pay taxes immediately in the current year, which will end the non-stop litigation in the assessment of search cases.

Business in the country needs an encouraging environment. Such extreme actions as search operations should be limited to unearthing the unaccounted monies of terrorist outfits, gangsters and underworld dons. Prosecution in case of concealed income of business will only lengthen the queue in various courts where innumerable people have been struggling to get justice for years. Further it could become a tool for personal vendetta in the hands of income tax authorities.

The author is a Delhi-based Chartered Accountant.

---

## /* rf.c Source Listing for Relevance Function Computation */

```
/* rf.c – Relevance Function computation for  number of text files,
     and listing them in the order of relevance function:-

The  program  grades  each  text  document  for  Document  Retrieval  by
comparing its the words with the words in the expanded query keywords,
computes  the  weight  of  each  text  document,  and  displays  each  text
document  name  along  with  its  relevance  weight,  in  the  order  of
relevance function, from high to low order by keeping the threshold of
20% of the highest value.

The command format is:

C:|> rf  i  textdocuments<cr>

where i stands for query no., is 1 or 2 ... or 5, and textdocuments is
textfile  containing  names  of  document  files  to  be  searched  in  the
document repository.

Note: All the  arguments are in lowercase
 */
#include <stdio.h>
#include <string.h>
#include <conio.h>
#include <ctype.h>
struct extd_qtype {
    char keyw[25];
    float fuzzywt;
    };
```

```
struct extd_qtype this_query1[50]; // for temp storage of current
                                   //expanded query for keyword1
struct extd_qtype this_query2[50]; // for temp storage of current
                                   // expanded query for keyword2

//database of expanded query terms with weights, for each of two terms,
//for five queries

static   struct   extd_qtype   e_query11[]   =   {{"house",1},{"home",.8},
{"building", .7},{"residence",.3},{"dwelling",.2}};
static struct extd_qtype e_query12[] = {{"loan",1}, {"finance",.8},
{"financing",.8},{"mortgage",.7},{"borrow",.5},{"advance",.4},{"credit"
,.3}};

static struct extd_qtype  e_query21[]={{"education",1},{"educate",.9},
{"educational",.9},{"university",.8},{"universities",.8},{"instruction"
,.8},{"academic",.8}, {"educating",.7},{"student",.7},{"school",.7},
{"schools",.7},{"college",.7},{"faculty",.6},{"teacher",.6},{"colleges"
,}, {"teaching",.6},{"institute",.6},{"schooling",.6},{"courses",.6},
{"course",.6},{"campus",.5},{"learning",.5}};

static struct extd_qtype e_query22[]={{"innovation",1},{"innovations",
1}, {"modern",.8}, {"modernization",.8}, {"innovative",.8},
{"innovatives",.8},{"advance",.7},{"advances",.7}, {"advancing",.7},
{"improveents",.6}, {"improvement",.6},{"projection",.6},
{"projections",.6},{"skill",.5},{"skills",.5},{"aids",.5},{"audio",.4},
{"video",.4}};

static struct extd_qtype  e_query31[]={ {"home",1},{"domestic",.9},
{"household",.8},{"family",.7}, {"house",.7}};
static struct extd_qtype  e_query32[] = {{"budget",1}, {"budgeting",
1},{"account",.8},{"funds",.7},{"fund",.7},{"finance",.6},{"finances",.
6},{"spend",.5},{"expense",.5},{"expenses",.5},{"purchase",.5},{"purcha
ses",.5},{"save",.4}, {"saving",.4}, {"expenditure",.4}, {"buy",.4},
{"spending",.4}, {"buying",.4},{"bills",.3}, {"bill",.3}};
```

```
static struct extd_qtype  e_query41[] ={{"career",1}, {"job",.9},
{"vocation",.8},{"employment",.8}, {"profession",.7},{"services",.6},
{"service",.6},{"occupation",.6}};

static struct extd_qtype  e_query42[]={{"prospects",1},
{"opportunities",.9}, {"opportunity",.9},{"avenue",.8}, {"avenues",.8},
{"forecast",.6}, {"options",.6},{"orient",.6},{"oriented",.6},
{"orientation",.6},{"prediction",.5},{"projection",.5},{"inductry",.5},
{"industries",.5},{"technology",.5},{"engineering", .5}, {"public"
,.4}, {"private", .4} };

static struct extd_qtype e_query51[]={{"tax",1},{"taxes",.9},
{"taxation",.9}, {"incometax",.9},{"taxing",.8}};

static struct extd_qtype  e_query52[] = { {"reforms",1}, {"reform", 1},
{"reliefs",.9}, {"relief",.9}, {"reduction",.8}, {"reductions",.8},
{"lower",.7},{"lowering",.7},{"benefits",.7},{"benefit",.7},{"lessen",.
6}, {"deductions", .5}};

char words[10000][25];   // can hold many  sentences of text

struct docu_rank_stru {
    char docu_name[15];
    int size_in_words;
    float rank;
} docu_rank[100]; // stores the ranking of each document
                //id is sentence number, rank is sentence rank

int wc;          // set word count to zero
static  int qs1[]={5, 22, 5, 8, 5 }; //number of query terms in each
                                //expanded query for the term1
static  int qs2[]={7, 18, 20, 18, 12 }; // number of query terms in
                                //each expanded query for the term2
int query_no, dc=0;    //query number, documents' count to be searched
```

```
//here is main
void main( int argc, char *argv[]){

                    // argv[1] query id number
                  // argv[2]  document names file name
    void read_text(char *w, int);
                  // argument are - ftext file name, file count no.
    void save_q(void);   // save current expanded query to this_query
    void  docu_score(int);      // score each  text document
    void print_docu_score(void);
                        // relevance func calculate print for each docu.
    FILE  *fp;
    char  temp[15], prech=' ',ch;
    int i,j, wi=0;   //open the argv[2] file and store all the
                      // document names in docu_names array
    if((fp=fopen(argv[2],"r"))==0){
        printf("Text file %s cannot be opened\n",argv[2]);
        exit(0);
    }
//collect all documents names and store in an array
    while((ch=getc(fp))!=EOF) {
      if((prech==' ' || prech=='\n' || prech=='\t') &&
      (ch==' ' || ch=='\n' || ch=='\t')) //continuous white space, skip
          ;
      else
          if(ch!=' ' && ch!='\n' && ch!='\t' )
            temp[wi++]=ch;    // valid character
          else
            if ((prech!=' ' && prech !='\n' && prech != '\t' ) &&
               (ch==' ' || ch=='\n' || ch=='\t' ) )
              {                       // current word over
                temp[wi++]='\0';  // terminate curent word
                strcpy(docu_rank[dc++].docu_name,temp);
                      // new word found, store in array,
                      // incr total document count
                wi=0;
```

```
              } //if prech
         prech=ch;
    }     // while over && file over
   fclose(fp);

  //  to adjust for the last word after the  eof was encountered
  temp[wi++]='\0';
  i=0;
  while(isspace(temp[i])) i++;
  j=0;
  while((temp[j]=temp[i])!='\0')
  { i++;
    j++;
  }
  if (strlen(temp)>0)
      strcpy(docu_rank[dc++].docu_name,temp);
       // new word found at end, store in the array

   query_no=atoi(argv[1]);
   save_q();      // save expanded query terms in query array
                  // for current query id

   // read the text documents one by one, and process each
   for(i=0; i < dc; i++) {
      read_text(docu_rank[i].docu_name,i);
       // read text_docu no. i, and store into array
      docu_score(i);            // score the document no. i
   }  // for, all documents score
   print_docu_score();
      // print all documents in order of Relevance Function
}//main

// function for reading the textfile and  storing into the array,
// sentence by sentence
```

```
void read_text( char *fname, int n){ // read from document file fname,
file sequence no
    char ch,prech=' ',  temp[25];
    int i,j, wi=0;
  //  global wc total word count, global sentc=sentence counter
  // wi current word index
    FILE  *fp;
    wc=0;
    if((fp=fopen(fname,"r"))==0){
       printf("Text file %s cannot be opened\n",fname);
       exit(0);
     }
    while( (ch=getc(fp))!=EOF){  // continue while not enf of file
       if( (prech==' ' || prech=='\n' || prech=='\t') &&
         (ch==' ' || ch=='\n' || ch=='\t'))
           // continuous white space, skip
         ;
       else
         if(ch!=' ' && ch!='\n' && ch!='\t' && ch!='.' &&
             ch!=',' && ch!=';' && ch!='?' && ch!='-' && ch!='!')
            temp[wi++]=ch;         // valid character
         else
            if (((prech!=' ' && prech !='\n' && prech != '\t' &&
prech!='.' && prech!='-' && prech!=',' && prech!=';' && prech!='?' &&
prech!='!') && (ch==' ' || ch=='\n' || ch=='\t' || ch=='.'|| ch==';'||
ch==',' || ch=='?' || ch=='-'|| ch=='!')) || wi>22)
      {                              // word over
        temp[wi++]='\0';            // terminate curent word
        strcpy(words[wc++],temp);
      // new word found, store in array, incr total wc
        wi=0;
      } //if prech
        prech=ch;
    }    // while over && file over
    fclose(fp);
```

```
 //  to adjust for the last text word after the  eof was encountered
//
 temp[wi++]='\0';
 i=0;
 while(isspace(temp[i])) i++;
 j=0;
 while((temp[j]=temp[i])!='\0')
   {i++;
   j++;
   }
 if (strlen(temp)>0)
  strcpy(words[wc++],temp);
 // new word found at end, store in the array
 docu_rank[n].size_in_words=wc;
//printf("%s%d\n",docu_rank[n].docu_name, docu_rank[n].
//size_in_words);
} // read_text
  // move the corresponding expanded query into the this_query array
 void save_q() {
   int i;
   if (query_no==1) { //query is no. 1
      for (i=0; i< qs1[0]; i++)
        this_query1[i]= e_query11[i];
      for (i=0; i< qs2[0]; i++)
        this_query2[i]= e_query12[i];
    }
   else
      if (query_no==2) { //query is no. 2
         for (i=0; i< qs1[1]; i++)
            this_query1[i]= e_query21[i];
         for (i=0; i< qs2[1]; i++)
            this_query2[i]= e_query22[i];
      }
        else
           if (query_no==3) { //query is no. 3
              for (i=0; i< qs1[2]; i++)
```

```
                    this_query1[i]= e_query31[i];
          for (i=0; i< qs2[1]; i++)
                    this_query2[i]= e_query32[i];
        }
        else
          if (query_no==4) { //query is no. 4
            for (i=0; i< qs1[3]; i++)
               this_query1[i]= e_query41[i];
             for (i=0; i< qs2[3]; i++)
                 this_query2[i]= e_query42[i];
            }
          else
            if (query_no==5) { //query is no. 5
              for (i=0; i< qs1[4]; i++)
                  this_query1[i]= e_query51[i];
              for (i=0; i< qs2[4]; i++)
                  this_query2[i]= e_query52[i];
            }
            else
             {
             printf("Wrong query number, it should be 1-5\n");
             exit(0);
             }
    } //save_q()

// score each document and store its ranking into the docu_rank array
// along with document name

void docu_score(int m){      // m is document no.
int j,k, qrysize1, qrysize2; // number of synonyms in current query
                         // in term 1 and 2
float wt1=0, wt2=0;          // init weights for two queryterms

 // do this for the first query term (for debug only)
 qrysize1= qs1[query_no -1];
 qrysize2= qs2[query_no -1];
```

```
 //printf("qsize  %d\n",  qrysize1);
 //printf("qsize  %d\n",  qrysize2);
 for(j=0; j< wc; j++) {  // for all the words in text
 // try to find match for this jth text word in the expanded-
 // query this-query1[]
      for ( k=0; k< qrysize1; k++) {
         if (strcmp(this_query1[k].keyw,words[j])==0){ // matches
          wt1=wt1+this_query1[k].fuzzywt;
// printf("%s   %f\n",this_query1[k].keyw,this_query1[k].fuzzywt);
        }
   } // for all k keywords

   for ( k=0; k< qrysize2; k++) {
      if (strcmp(this_query2[k].keyw,words[j])==0){ // matches
              wt2=wt2+this_query2[k].fuzzywt;
    //printf("%s   %f\n",this_query2[k].keyw,this_query2[k].fuzzywt);
                    }
    } // for all k keywords
 } // for all j text words

// assign fuzzy value of lower as EF to document, i.e., wt1 and wt2
        if(wt1<wt2)
          docu_rank[m].rank=wt1;
        else
          docu_rank[m].rank=wt2;
   //printf(" wt %f\n", wt);
   // rank = sum of the weight of terms/number of term in sentene
} // docu_score()

// list those sentences from the current document whose relevancy
// ranking is greater than threshold
void print_docu_score(){
      int i,j;
      float  thold;      // thold = threshold
      struct docu_rank_stru trank;
```

```
    //compute exact rank
      for(i=0; i< dc;i++) {
         docu_rank[i].rank = docu_rank[i].rank*1e+4
//(docu_rank[i].size_in_words * docu_rank[i].size_in_words );
      } // for i

//order the text document sentences in highest to lowest order of
//ranking
    for(i=0; i< dc-1;i++) {
       for(j=i+1; j<dc; j++){
          if (docu_rank[i].rank < docu_rank[j].rank) {
             trank= docu_rank[i];
             docu_rank[i]=docu_rank[j];
             docu_rank[j]=trank;
           } // if
        } // for j
      } // for i

    thold=docu_rank[0].rank * 0.20;

      // list the document names with their ranking
      printf("\nDocument Name    Ranking weigt");
      printf("\n==================================\n\n");
      for(i=0; i< dc; i++)
   if(docu_rank[i].rank > thold)
      printf("%s%f*1E-04\n",docu_rank[i].docu_name, docu_rank[i].rank);
      printf("===============================\n\n");
 } // print_docu_score()
```

---

## /* ie.c  Source Listing for Information Extraction Algorithm */

```
/* ie.c Information Extraction does the following jobs :-

Grades each sentence for IE in the text by comparing its words with
keywords in the expanded query, computes the weight of each sentence,
and displays each sentence (i.e., extracted information) in the order
of its ranking, from high to low, by keeping the threshold of 20% of
the highest value.

     The command is of the format:

C:|> ie  i  reltextdoc<CR>

where i is 1 or 2 ... or 5 standing for query number, and reltextdoc is
already retrieved relevant text document to be searched for information
extraction

Note : All the  arguments are in lowercase
 */
#include <stdio.h>
#include <string.h>
#include <conio.h>
#include <ctype.h>
struct extd_qtype {
    char keyw[25];
    float fuzzywt;
    };
struct extd_qtype this_query[50];  // for temp storage of current
//expanded query
```

```
struct extd_qtype *e_query1, *e_query2, *e_query3, *e_query4,
*e_query5, *this_query;
/* query 1 expansion */
static struct extd_qtype  e_query1[] = {  {"house", 1}, {"home", .8},
{"building", .7},{"residence", .3}, {"dwelling", .2}, {"loan", 1},
{"finance",.8},{"financing",.8},{"mortgage",.7}, {"borrow",.5},
{"advance",.4},{"credit",.3}};

/* query 2 expansion */
static struct extd_qtype  e_query2[] = {
{"education",1},{"educate",.9},{"educational",.9}, {"university",.8},
{"universities",.8},{"instruction",.8},{"academic",.8},{"educating",.7}
,{"student",.7},{"school",.7}, {"schools",.7},{"college",.7},
{"faculty",.6},{"teacher",.6},{"colleges",}, {"teaching",.6},
{"institute",.6},{"schooling",.6},{"courses",.6},{"course",.6},
{"campus",.5},{"learning",.5},{"innovation",1},{"innovations",1},
{"modern",.8},{"modernization",.8},{"innovative",.8},
{"innovatives",.8},{"advance",.7},{"advances",.7},{"advancing",.7},{"im
proveents",.6},{"improvement",.6},{"projection",.6},{"projections",.6},
{"skill",.5},{"skills",.5},{"aids",.5},{"audio",.4},{"video",.4}};

/* query 3 expansion */
 static struct extd_qtype  e_query3[] = {
{"home",1},{"domestic",.9},{"household",.8},{"family",.7},{"house",.7},
{"budget", 1}, {"budgeting", 1}, {"account", .8}, {"funds",.7},{"fund",
.7}, {"finance", .6},  {"finances",.6},{"spend",.5},{"expense",.5},
{"expenses",.5},{"purchase",.5},"purchases",.5},{"save",.4},{"saving",.
4},{"expenditure",.4},{"buy",.4},{"spending",.4},{"buying",.4},{"bills"
,.3},{"bill",.3}};

/* query 4 expansion */
 static struct extd_qtype  e_query4[] =  {
{"career",1},{"job",.9},{"vocation",.8},{"employment",.8},
{"profession",.7}, {"services",.6},{"service",.6},{"occupation",.6},
{"prospects",1},{"opportunities",.9},{"opportunity",.9},{"avenue",.8},
```

```
{"avenues",.8},{"forecast",.6},{"options",.6},{"orient",.6},{"oriented"
,.6}, {"orientation",.6},{"prediction",.5},{"projection",.5},
{"inductry",.5},{"industries",.5}, {"technology",.5}, {"engineering",
.5}, {"public" ,.4},{"private", .4} };

/* query 5 expansion */
static struct extd_qtype  e_query5[] = {{"tax",1},{"taxes",.9},
{"taxation",.9},{"incometax",.9},{"taxing",.8},{"reforms",1},
{"reform", 1}, {"reliefs", .9}, {"relief",.9},{"reduction", .8},
{"reductions", .8},{"lower", .7}, {"lowering",.7},{"benefits", .7},
{"benefit",.7}, {"lessen",.6}, {"deductions", .5}};

char tsents[10000][25];  // storage for relevant document sentences
int sent_bound[1000][2]; // sentence boundaries array
                        // main index is sentence number,
                        // 1st column is sentence start index in tsents
              // 2nd column is sentence end(last word) index in tsents

 struct sno_rank{
    int id;
    float rank;
    } sent_rank[1000];  // stores the ranking of each sentence
                        // id is sentence number, rank is sentence rank

int wc=0;           // set word count to zero
int sentc=0;        // sentence count, used as index in sent_bound array
static  int qs[]={12, 40, 25, 26, 17 }; // number of query terms in
                                        //each expanded query
int query_no;

//here is main
void main( int argc, char *argv[]){
    void read_text( char *w);
    void save_q(void);  // save current expanded query to this_query
    void  sent_score(void);// score each sentence of this text document
    void iextract(char *w);    // extract the information
```

```
            query_no=atoi(argv[1]);
            read_text(argv[2]);
            save_q();
            sent_score();
            iextract(argv[2]);
    }//main

// function for reading the textfile and storing into the array,
//sentence by sentence

void read_text( char *fname){
    char ch,prech=' ',  temp[25];
    int i,j, wi=0;//global wc total word count, global sentc=sentence
//counter
 // wi current word index
    FILE  *fp;
    if((fp=fopen(fname,"r"))==0){
       printf("Text file %s cannot be opened\n",fname);
       exit(0);
     }
    sent_bound[sentc][0]=0;  // save sentence start index
    while( (ch=getc(fp))!=EOF){  // continue while not end of file
        if( (prech==' ' || prech=='\n' || prech=='\t') &&
          (ch==' ' || ch=='\n' || ch=='\t'))
     // continuous white space, skip
          ;
        else
          if(ch!=' ' && ch!='\n' && ch!='\t' && ch!='.' &&
             ch!=',' && ch!=';' && ch!='?' && ch!='-' && ch!='!')
            temp[wi++]=ch;               // valid character
          else
            if (((prech!=' ' && prech !='\n' && prech != '\t' &&
                prech!='.' && prech!='-' && prech!=',' && prech!=';'
                & prech!='?' && prech!='!') && (ch==' ' || ch=='\n' ||
                ch=='\t' || ch=='.'|| ch==';'|| ch==',' || ch=='?' ||
                ch=='-'|| ch=='!')) || wi>22)
```

```
            {                          // word over
             temp[wi++]='\0';          // terminate current word
             strcpy(tsents[wc++],temp);
              // new word found, store in array
              // incr total wc
             if(ch=='.') {
             sent_bound[sentc][1]=wc-1; // save sentence end index
              sentc++;
          // increment sentence counter
              sent_bound[sentc][0]=wc;
          // save next sentence start index
              }  //if ch
             wi=0;
            } //if prech
     prech=ch;
 }   // while over && file over
 fclose(fp);

 //  to adjust for the last text word after the  eof was encountered
//
 temp[wi++]='\0';
 i=0;
 while(isspace(temp[i])) i++;
 j=0;
 while((temp[j]=temp[i])!='\0')
   {i++;
   j++;
   }
 if (strlen(temp)>0)
   strcpy(tsents[wc++],temp);
    // new word found at end, store in the array
 sent_bound[sentc][1]=wc-1;
   // save sentence end index for last sentence
   //printf("No of sents %d\n", sentc);
   //test for words
   //for(i=0; i<wc; i++) printf("%s\n",tsents[i]);
```

```
} // read_text

// move the corresponding expanded query into the this_query array
void save_q() {
 int i;
      if (query_no==1) { //query is no. 1
         for (i=0; i< qs[0]; i++)
            this_query[i]= e_query1[i];
          }
      else
         if (query_no==2) { //query is no. 2
             for (i=0; i< qs[1]; i++)
                this_query[i]= e_query2[i];
          }
          else
             if (query_no==3) { //query is no. 3
                for (i=0; i< qs[2]; i++)
                   this_query[i]= e_query3[i];
              }
             else
                if (query_no==4) { //query is no. 4
                   for (i=0; i< qs[3]; i++)
                      this_query[i]= e_query4[i];
                }
                else
                   if (query_no==5) { //query is no. 5
                       for (i=0; i< qs[4]; i++)
                          this_query[i]= e_query5[i];
                   }
                   else
                    {
                    printf("Wrong query number, it should be 1-5\n");
                    exit(0);
                    }
 //  for(i=0; i<qs[query_no -1];i++) printf("%s  %f \n",
 //            this_query[i].keyw, this_query[i].fuzzywt);
```

```
    } //save_q()

// score each sentence and store its ranking into the sent_bound array
// at the position of third column in this array

void sent_score(){
            int i,j,k, qrysize;
         //wt=current sentence weight, qurysize is
         // number of words in current query
          float wt;
    //go sentence by sentence and score each, sentc is total sentence
    // count in this document
          qrysize= qs[query_no -1];
    //printf("qsize  %d\n",  qrysize);
          for(i=0; i< sentc; i++) {
                wt=0;
      // pickup the start and end of current sentence, in tsents[]
               for (j=sent_bound[i][0]; j<=sent_bound[i][1]; j++){
      // try to find match of this text word in the expanded-
      // query this-query[]
              for ( k=0; k< qrysize; k++) {
                if (strcmp(this_query[k].keyw,tsents[j])==0)
                   // matches
                        wt=wt+this_query[k].fuzzywt;
                  } // for k
                } // for j
                sent_rank[i].id=i;
                 //printf(" wt %f\n", wt);
          //rank = sum of the weight of terms/number of term in sentene
            sent_rank[i].rank= wt/(sent_bound[i][1]- sent_bound[i][0]);
          //printf("start %d end %d\n",sent_bound[i][0],
          // sent_bound[i][1]);
          } // for i, next sentence
    } // sent_score()
```

```
// list those sentences from the current document whose relevancy
ranking is greater than threshold
void iextract(char *fname){
      int i,j,flag;
      float  thold;      // thold = threshold
      struct sno_rank  trank;
//order the text document sentences in highest to lowest order of
//ranking
      for(i=0; i< sentc-1;i++) {
        for(j=i+1; j<sentc; j++){
          if (sent_rank[i].rank < sent_rank[j].rank) {
              trank= sent_rank[i];
              sent_rank[i]=sent_rank[j];
              sent_rank[j]=trank;
          } // if
        } // for j
      } // for i
     thold=sent_rank[0].rank * 0.20;

      // list the expanded query terms
      printf("\nEXPANDED QUERY TERMS");
      printf("\n====================\n\n");
      for(i=0; i< qs[query_no -1]; i++)
          printf("%s ", this_query[i].keyw);
// list the sentences from this document whose ranking is above  the
// threshold
     printf("\n\nEXTRACTED RELEVANT TEXT FROM TEXT FILE %s\n",fname);
     printf("==============================================\n\n");
     flag=0;
     for(i=0; i < sentc; i++){
        if (sent_rank[i].rank > thold) {   // sentence > 20%
                                  // list this sentence
        printf("[Text Segment no. %d]  [rank*1000=%8.4f]\n",
            sent_rank[i].id,sent_rank[i].rank);
        flag=1;
```

```
        for(j=sent_bound[i][0]; j<=sent_bound[i][1]; j++)
            printf("%s ",tsents[j]);
        printf("\n\n\n");
      } // if
    } // for i
    if(!flag)printf("\n\n THERE IS NO RELEVANT TEXT IN THIS FILE
      !!!\n");
} // iextract()
```

---

**Unit 1:**

*Keywords $o_4$: {trees, water}=.4, {animals}=.3*

Forests are very useful for us. We get number of things from forests trees for our daily use and industries. Some of them are timber, fuel wood, pulp, resin, gum, cane, grasses, and medical herbs. Timber is used for building houses and making of furniture, tools, and agriculture products. Pulp and soft wood are used for making paper. There are many other uses of forests. They cause rainfall. They check soil erosion, and help in conserving soil. They control floods and reduce damage by winds, etc. Rainfall water collected in the forests is used for accumulation in dams, and supplied in rivers and canals for irrigation purpose. Trees in forests also give fresh oxygen.

Forests are home for many wonderful animals; these animals include giraffe, deer, zebra, elephants, and rabbits. Staying away from human, these animals feel free and protected.

The trees from the forest are used in number of industries.

*Keywords $o_5$: {water}=.35, {birds, animals}=.5*

People in villages get water from rivers, ponds and wells. This water is not germfree. The villagers also bath and wash their clothes in the ponds and rivers. Even they give bath to their animals in the same ponds and rivers. The water from well is good but they are open. Due to this water gets dirty. This causes germs and diseases. Hand pumps are also used to get the water from ground.

Villagers regular income depends on animals and food grains produced in the fields. The animals, like - cow, buffalo, goats provide milk for own consumption as well as for sale. The sheep are for wool production. Sale of animals becomes the one the source of income. The various birds are not good for villagers, as they eat the crops at the early stage of growth, as well as the grains are directly picked from the ripe crops.

**Unit 2:**

*Keywords $o_6$: {water, animals}=.1, {trees}=.1, {tiger}=.1*

At number of places where there are plains and water is in plenty, the animals and birds are in plenty. The places are formed as bird sanctuaries. The water is accumulated in large size lakes which do not dry for long times.

The forest department has planted number of trees in the areas where water is in plenty. However, when the water is scarce these trees often dry.

The wildlife in our country is our national heritage and asset. Various kinds of wild animals are found in the Indian forests. There are elephants, rhinoceroses, tiger, lions, etc. A large variety of deer and antelope, monkeys and langurs, wolves and jackals are found in the hills.

The wild life protection has increased the number of tourist in the country, and has becomes one of the major source of foreign exchange. Every year lacs of tourists from around the world visit India.

*Keywords $o_7$: {animals}=.45, {tiger}=.3*

In order to save wildlife our government has taken several steps. The government has set up about 83 National Parks. In these parks, wildlife and natural vegetation are preserved. Tigers are preserved in the Corbett Park. There is a park for rhinoceroses in Assam. There also wild life and birds sanctuaries (447) where animals and rare birds live and roam around without any fear of being hunted. The board of Wildlife advises the government as to what steps it should take to preserve the wildlife. A wildlife week is observed in a year to lay stress on the need to preserve the wild life.

Tiger and rhinoceroses are some endangered species of wildlife in India, so for them special projects have been prepared. Tiger project have proved very successful. About 16 tiger reserves have been set up in different parts of the country where special care is being taken for all around protection and betterment of tigers.